\documentclass[journal]{IEEEtran}
\usepackage{amsmath}
\usepackage{amsfonts}       % blackboard math symbols
\usepackage{multirow}
\usepackage{makecell}
\usepackage[capitalize]{cleveref}

\usepackage{ifpdf}
\usepackage{cite}
\ifCLASSINFOpdf
   \usepackage[pdftex]{graphicx}
\else
   \usepackage[dvips]{graphicx}
\fi
\usepackage{amsmath}
\usepackage{array}
\usepackage{adjustbox}
\usepackage{tikz}
\usepackage{algorithmic,algorithm}
\usepackage{color}
\ifCLASSOPTIONcompsoc
  \usepackage[caption=false,font=normalsize,labelfont=sf,textfont=sf]{subfig}
\else
  \usepackage[caption=false,font=footnotesize]{subfig}
\fi

\usepackage{url}
\usepackage{xspace}
\makeatletter
\newcommand\onedot{\futurelet\@let@token\@onedot}
\def\@onedot{\ifx\@let@token.\else.\null\fi\xspace}
\def\eg{\emph{e.g}\onedot} 
\def\ie{\emph{i.e}\onedot} 
 
\def\etal{\emph{et al}\onedot}

\makeatother

\begin{document}
\title{Physically-Constrained Transfer Learning through Shared Abundance Space for Hyperspectral Image Classification}

%\author{Michael~Shell,~\IEEEmembership{Member,~IEEE,}
%        John~Doe,~\IEEEmembership{Fellow,~OSA,}
%        and~Jane~Doe,~\IEEEmembership{Life~Fellow,~IEEE}% <-this % stops a space
%\thanks{M. Shell was with the Department
%of Electrical and Computer Engineering, Georgia Institute of Technology, Atlanta,

\author{Ying~Qu, \IEEEmembership{Member,~IEEE,}
	Razieh Kaviani Baghbaderani, \IEEEmembership{Student Member,~IEEE}, Wei Li, \IEEEmembership{Senior Member,~IEEE}, Lianru Gao, \IEEEmembership{Senior Member,~IEEE}, Yuxiang Zhang and
	Hairong~Qi, \IEEEmembership{Fellow,~IEEE}
%	and Chiman~Kwan, \IEEEmembership{Member,~IEEE}
\thanks{Ying~Qu, Razieh Kaviani Baghbaderani and Hairong~Qi are with the Advanced Imaging and Collaborative Information Processing Group, Department of Electrical Engineering and Computer Science, University of Tennessee, Knoxville, TN 37996 USA (e-mail: yqu3@vols.utk.edu; rkavian1@vols.utk.edu; hqi@utk.edu).}
\thanks{Wei. Li and Yuxiang Zhang are with the School of Information and Electronics, Beijing Institute of Technology, Beijing 100081 China (e-mail: liwei089@ieee.org).}
\thanks{Lianru Gao is with the Key Laboratory of Digital Earth Science, Aerospace Information Research Institute, Chinese Academy of Sciences, Beijing 100094, China (e-mail: gaolr@aircas.ac.cn).}}

%\thanks{Chiman~Kwan is with Applied Research LLC, Rockville, MD,20850 USA (chiman.kwan@arllc.net)}
%\thanks{Steven Vance is with NASA, Jet Propulsion Lab, Pasadena, CA. USA (email: steven.d.vance@jpl.nasa.gov)}}


\markboth{Submitted to IEEE TRANSACTIONS ON GEOSCIENCE AND REMOTE SENSING}%
{Shell \MakeLowercase{\textit{et al.}}: Bare Demo of IEEEtran.cls for IEEE Journals}



\maketitle
%\onecolumn

\begin{abstract}
Hyperspectral image (HSI) classification is one of the most active research topics and has achieved promising results boosted by the recent development of deep learning. However, most state-of-the-art approaches tend to perform poorly when the training and testing images are on different domains, ~\eg, source domain and target domain, respectively, due to the spectral variability caused by different acquisition conditions. Transfer learning-based methods address this problem by pre-training in the source domain and fine-tuning on the target domain. Nonetheless, a considerable amount of data on the target domain has to be labeled and non-negligible computational resources are required to retrain the whole network. In this paper, we propose a new transfer learning scheme to bridge the gap between the source and target domains by projecting the HSI data from the source and target domains into a shared abundance space based on their own physical characteristics. In this way, the domain discrepancy would be largely reduced such that the model trained on the source domain could be applied on the target domain without extra efforts for data labeling or network retraining. The proposed method is referred to as physically-constrained transfer learning through shared abundance space (PCTL-SAS). Extensive experimental results demonstrate the superiority of the proposed method as compared to the state-of-the-art. The success of this endeavor would largely facilitate the deployment of HSI classification for real-world sensing scenarios.
\end{abstract}


% \fi
\IEEEpeerreviewmaketitle

\section{Introduction}
\label{sec:intro}
Hyperspectral images (HSI) collect rich spectral information from hundreds of narrow and contiguous spectral bands covering the electromagnetic spectrum from visible to near-infrared wavelengths. Because of the improved image acquisition technologies, HSI analysis has been playing a very important role % matured into one of the most powerful technologies 
in the field of remote sensing.  
%attracted an increasing attention from remote sensing community 
%and has been deeply involved in various earth resource, environmental mapping and monitoring tasks~\cite{paoletti2019deep,li2019deep}. 
Among the numerous advanced techniques for HSI analysis, HSI classification  
becomes one of the most active research topics and contributes significantly in a wide range of applications, including ecological science, ecology management, precision agriculture, military applications, etc.~\cite{brown2010hydrothermal, he2017recent,liang2016sampling,paoletti2019deep,li2019deep}.

HSI classification aims to categorize each individual pixel of HSI into one of the given classes based on either the raw input data or features extracted from the raw data. Hence the classification performance relies heavily on the effectiveness of the feature extraction procedure. %At the early stage of development, 
Traditional HSI classification approaches mainly extract hand-crafted (or engineered) shallow  features~\cite{du2001linear,ghamisi2017advanced,paoletti2019deep}. One major limitation of these early approaches is the limited generalization capacity when applied to complex scenarios~\cite{li2019deep}. Recently, deep learning-based approaches have been developed that automatically extract deep features from the input data. These deep features %, as the emerging frontier of machine learning, has been introduced to overcome the above limitations and 
have demonstrated unique strength in HSI classification. The deep learning-based approaches can be generally categorized into three groups according to the types of spaces they explore,~\ie, the spectral-based networks~\cite{hu2015deep,SongYang2019LCCf}, %that extract features from the \textit{spectral space} of HSI, 
the spatial-based networks~\cite{sharma2016hyperspectral,cao2018hyperspectral,zhu2018deformable}, % that extract features from the \textit{spatial space} of HSI, 
and the spectral-spatial-based networks~\cite{zhong2018spectral,hamida20183,luo2018hsi,roy2019hybridsn}.  %that explore both spaces. 
By either increasing the \textit{representative power}~\cite{chen2014deep,chen2015spectral,peng2017active} or \textit{discriminative power}~\cite{luo2018shorten,mou2017deep,li2019deep} of the features, these methods could achieve very promising state-of-the-art classification performance. 

Nonetheless, deep learning-based classification suffers from deficiencies inherent with any supervised deep learning practices and HSI classification in particular, %there remains bottlenecks and impediments that hard to breakthrough,
~\ie, 1) a considerable amount of samples have to be labeled before training to prevent overfitting, and more importantly, 2) even if we take the time and effort to label a large amount of samples to train a classifier, such classifier may perform poorly if the test images are from a ~\textit{different} domain,~\ie, images taken under different acquisition conditions. This is mainly because there exists large ``spectral distribution variability'' among the same class of samples in different domains, due to the changes in illumination, environmental, atmospheric and temporal conditions~\cite{paoletti2019deep,li2019deep}.

This, naturally, raises a question: \textit{can we take advantage of the existing labeled data and learn prior knowledge from them to classify images from a different domain?}

%how can we train a more general classifier, whose weights can extract flexible and effective features from both the existing labeled samples and newly collected data, such that the classifier can apply on the new data directly without extra efforts?

In the following, we define the domain of the existing labeled data as the \textit{source domain} and the domain of the test data taken under different conditions as the \textit{target domain}. There have been a few transfer learning~\cite{deng2018transfer,zhang2019hyperspectral,paoletti2019deep} or active deep learning~\cite{deng2018active,lin2018active,li2019deep} based methods specifically designed to address the above problem. They transfer the prior knowledge learned from the source domain to the target domain  by \textit{pre-training} in the source domain and \textit{fine-tuning} in the target domain. These methods could achieve competitive performance and largely reduce the number of required labeled data in the target domain. %Nonetheless, the trained classifier still has a strong dependency towards the training dataset from the source domain, and the learned features cannot be easily transferred to the target domain without fine-tuning. 
However, the learning strategy based on fine-tuning is far from perfect. In order to fine-tune the network, a considerable amount of data in the target domain has to be labeled and non-negligible computational resources for additional optimization steps are required to retrain the whole network~\cite{li2016revisiting,paoletti2019deep,li2019deep}. 
%Such additional non-trivial and time-consuming burden could further complicate the network training procedure which is already intimidating/prohibitive enough for most people.

%Such additional computational burden could greatly complicate the training of a DNN which is already intimidating enough for most people. 

This paper aims to design a new transfer learning scheme to bridge the gap between the source and target domains without extra efforts for data labeling and network retraining in the target domain. This is done by minimizing the domain discrepancy in a potential shared space,~\ie, the  abundance space, among different domains. 

%which is achieved based on the physical characteristics of the spectra from both domains. 

%To break the above mentioned hurdles, the key is to learn a domain invariant model by minimizing the domain distribution discrepancy. Based on this principle, we propose a physically constrained transfer learning approach to address the challenges of HSI classification problem in shared space, referred to as PCTL-SS. The essential contribution of this work is the realization of the transfer learning scheme that can project the data from different domains into a shared space based on the physical characteristics of the spectra, as shown in Fig.~\ref{fig:domain}. With this scheme, the network is able to extract \textit{share intrinsic representations} from the same type of objects in different domains taken under different acquisition conditions; such that the classifier trained in the source domain can be directly utilized to categorize pixels in the target domain without additional efforts for data labeling or fine-tunning.

%such that given a few training samples in the source domain, we are able to train a classifier that can be directly applied to categorize pixels of the target data possessing the same object labels, without having to collect more ground truth labels in the target domain. 

Due to the resolution issue, each pixel in an HSI usually covers a large geographical area, resulting in the so-called ``mixed pixel'',~\ie, each pixel tends to cover more than one constituent material.  These mixtures are generally assumed to be a linear combination of a few spectral bases with the corresponding coefficients (or abundance) satisfying two physical constraints, \ie, the sum-to-one constraint and the non-negative constraint. %The proposed  method  is  designed  based  on  the  assumptions that both the pixels in the source and target domain can be decomposed into spectral bases and abundance. 
For the pixels belonging to the same class in different domains, although they may possess different spectral characteristics, their abundance,  indicating how much contribution each spectral basis has in constructing a certain pixel, should remain similar. Based on this hypothesis, we propose a physically-constrained transfer learning method to address the challenges of HSI classification, referred to as PCTL-SAS. 

\begin{figure}[htp]
	\begin{minipage}{1\linewidth}
	\centering
		\includegraphics[width=0.6\linewidth]{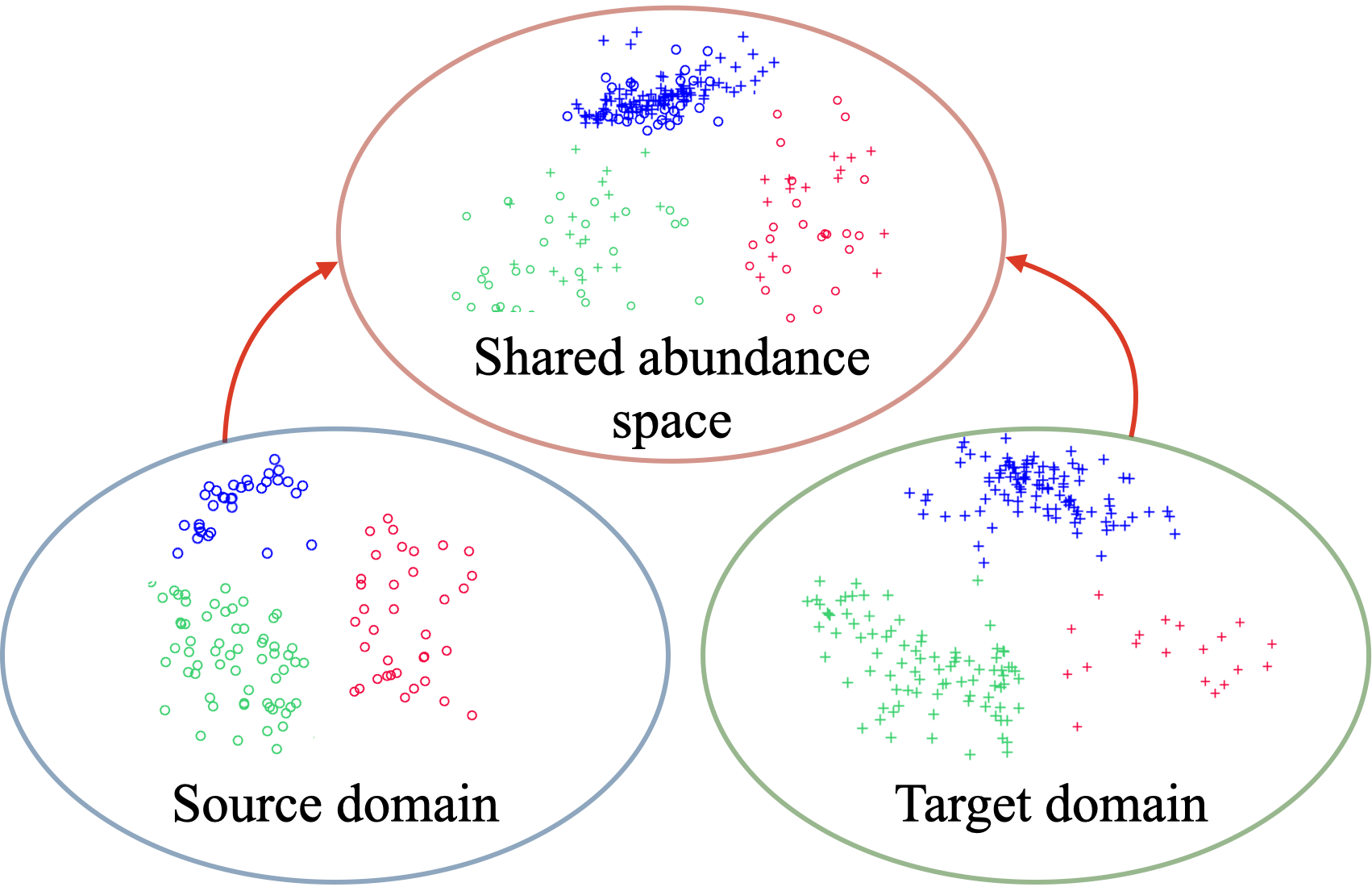}
		\caption{Transfer learning of domain-invariant shared intrinsic representations through the shared abundance space.}
		\label{fig:domain}
	\end{minipage}
\end{figure}
The essential contribution of this work is the realization of the transfer learning scheme that can project the HSI data from the source and target domains into a shared abundance space based on their physical characteristics, as shown in Fig.~\ref{fig:domain}. With this scheme, the network is able to extract domain invariant \textit{shared intrinsic representations}, such that the classifier trained in the source domain can be directly utilized to categorize pixels in the target domain without additional efforts for data labeling or retraining. 
The novelty of this work is three-fold. 
\begin{itemize}
	\item First, to bridge the gap between the source and target domains, the proposed method projects the images from both domains into a shared abundance space with a shared Dirichlet-encoder  that naturally meets the non-negative and sum-to-one physical constraints. Based on the fact that each pixel usually consists of only a few bases, the abundance space is further constrained with the sparse entropy function to extract more effective representations.
	\item Second, we propose an affine-transfer decoder that %which allows the network to 
	learns the potential transfer function between different domains. The transfer function is further regularized by the proposed mutual discriminative network to match the correspondence between the pixels belonging to the same class in different domains. In this way, the matched shared intrinsic representations can be extracted from different domains.
	%To train a general classifier that works in the target domain without any labels, a   %And a mutual discriminative network to further enforce such correspondence by maximizing the mutual information (MI) between the representations and their own HSI input. an affine-transfer decoder is proposed to learn the spectral bases as well as their corresponding transfer information among different domains. 
	\item Third, to utilize the information from the different dimensional spaces of a given pixel, a densely-connected 3D-CNN classifier is proposed and concatenated to the shared encoder. In this way, both the representative and discriminative power of the network is enhanced. What's more important is that the classifier is built on the domain-invariant shared abundance space, such that it can be trained with the labeled data in the source domain that still works in the target domain without additional data labeling or retraining. 
\end{itemize}

The rest of the paper is organized as follows. Sec.~\ref{sec:related} provides an overview of the state-of-the-art HSI classification approaches. Sec.~\ref{sec:formulate} provides the general formulation of the transfer learning problem and discusses the rationale of the proposed approach. Sec.~\ref{sec:proposed} elaborates the proposed PCTL-SAS method. Sec.~\ref{sec:exp} performs comprehensive evaluations of the proposed method. Conclusions are drawn in Sec.~\ref{sec:conclusion}.

\section{Related Work}
\label{sec:related}
\subsection{Hyperspectral Image Classification}

Hyperspectral image (HSI) classification is one of the most vibrant fields in remote sensing community and has been developed for decades~\cite{li2019deep}. At the early stage, traditional methods are mainly trained based on the pixel-wise strategy, which explores %to address the problem of HSI classification according to
the spectral information of individual pixels. Numerous pixel-wise classifiers have been developed, among which, support vector machine (SVM) \cite{melgani2004classification,2017Multiscale}, k-nearest-neighbor \cite{samaniego2008supervised}, maximum likelihood criterion \cite{ediriwickrema1997hierarchical}, and logistic regression \cite{li2010semisupervised} are more popular. %, which played a crucial role in a broad range of applications. 
To improve classification accuracy, an instrumental way is to extract effective features or reduce the dimension of the HSI with different techniques before classification,~\eg, principal component analysis (PCA) \cite{licciardi2011linear,jenson1979principal}, linear discriminative analysis (LDA) \cite{bandos2009classification}, independent component analysis (ICA) \cite{villa2011hyperspectral}, and minimum noise fraction (MNF) \cite{green1988transformation}, to name a few~\cite{2018yu,yu2019global,hong2020graph}. Nonetheless, these methods rely heavily on the hand-crafted features designed for a specific task, which limited their ability to handle complex scenarios. 

The performance of HSI classification has been boosted with the emergence of deep learning~\cite{li2019deep,paoletti2019deep}. The deep-learning-based approaches are able to extract hierarchical features automatically in an end-to-end manner, which demonstrated unique strength in HSI classification. Generally, existing deep-learning-based approaches can be categorized into three groups according to the types of HSI spaces they explore,~\ie, the  spectral-based methods, the  spatial-based  methods, and the spectral-spatial-based methods~\cite{li2019deep}. Spectral-based methods were developed based on 1D convolutional neural network (CNN) architecture to perform pixel-wise classification according to the spectral characteristics of HSI~\cite{hu2015deep,SongYang2019LCCf}. 
Although 1D-CNN-based methods could achieve better results compared to those of the traditional methods, the correlation information among spatially-adjacent pixels were not fully explored. Spatial-based methods further improved the classification accuracy by %adopting a patch-wise training strategy which 
applying 2D-CNN on the patch surrounding the given pixel to utilize its spatial adjacency information~\cite{sharma2016hyperspectral,cao2018hyperspectral,zhu2018deformable}. In order to fully utilize the spectral-spatial correlation information, spectral-spatial-based methods were developed by adopting the 3D convolutional kernels and demonstrated promising  classification results in recent literature~\cite{hamida20183, zhong2018spectral, luo2018hsi, li2017spectral, roy2019hybridsn, yue2015spectral}. For example, Spec-Spat~\cite{li2017spectral}, 3D-CNN~\cite{hamida20183}, and SSRN~\cite{zhong2018spectral} were designed with different network structures but all based on 3D-CNN blocks. HSI-CNN~\cite{luo2018hsi} and HybridSN~\cite{roy2019hybridsn} were constructed with both the 3D-CNN and 2D-CNN basic blocks. 

All these state-of-the-art approaches could achieve promising results when the training and testing are preformed in the same domain. However, they may perform poorly on a different domain due to spectral variability, even if the labeled classes are shared among different domains.

\subsection{Hyperspectral Image Transfer Learning}
Recently, a few transfer-learning-based methods were proposed attempting to address the challenges mentioned above. Generally, given sufficient amount of labeled data from both the source and target domains, these methods solve the HSI classification problem by adopting a transfer learning strategy, which trains the model on the source domain and fine-tune such model on the target domain. \cite{liu2019unsupervised} constructed multiple geodesic flows using pairs of spatial and spectral features obtained from both the source and target domains that are further fed to an SVM to complete the classification task. \cite{zhang2019hyperspectral} utilized a 3-D lightweight CNN structure and performed two transfer learning strategies, including cross-sensor and cross-modal strategies. HT-CNN~\cite{he2019heterogeneous} reused the priors learned with low-level and mid-level of VGGNet~\cite{simonyan2014very} and adopted an attention mechanism to adjust the feature maps due to the difference between the heterogeneous data sets.

% Recently, transfer learning-based studies have attempted to address the classification problem in a target domain using a trained model on the source domain. A group of methods exploit the unlabeled data in the target domain, in addition to the labeled data in the source domain, to fine-tune the algorithm. \cite{liu2019unsupervised} constructed multiple geodesic flows using pairs of spatial and spectral features obtained from both the source and target domains that are further fed to an SVM to complete the classification task. \cite{zhang2019hyperspectral} utilized a 3-D lightweight CNN structure and performed two transfer learning strategies, including cross-sensor and cross-modal strategies.

Another set of works employ the active learning strategy to manually annotate a small number of samples in the target domain to boost the classification performance. \cite{deng2018transfer} sequentially retrained a multi-kernel classifier with a collection of the available data from the source domain and the added, most informative samples acquired with active queries from the target domain. The work of \cite{lin2018active} extracted the salient samples from both the source and target domains, explored the correlation of these domains through a deep network layer-by-layer analysis, and further fine-tuned the network. \cite{deng2018active} proposed a  hierarchical stacked sparse auto-encoder to learn joint spectral–spatial features which are later transferred to the target domain by fine-tuning the network using a few labeled samples from both the source and target domains.

Although these methods could classify images in the target domain successfully, the learning strategy based on fine-tuning is far from  perfect.  In  order  to  fine-tune the models,  sufficient amount of data samples from both the source and target domains have to be labeled and the whole network has to be retrained with non-negligible computational resource~\cite{li2019deep,paoletti2019deep}.

\section{Problem Formulation and Motivations}
\label{sec:formulate}
Given $n$ samples, $X_S = \{\mathbf{x}_S\vert {\mathbf{x}_S}_i\in \mathcal{X_S}, i = 1,\cdots, n_S \}$, in the source domain, $\mathcal{X}_S$, and the corresponding label set $Y_S=\{y_S|{y_S}_i\in\mathcal{Y}_S, i = 1,\cdots, k_S\}$ in the source label domain, $\mathcal{Y}_S$, we could build a training set and learn a classifier $f_S:\mathcal{X}_S\rightarrow\mathcal{Y}_S$. However, the classifier $f_S$ trained in the source domain may fail to categorize pixels, $X_T= \{\mathbf{x}_T\vert {\mathbf{x}_T}_i\in \mathcal{X_T}, i = 1,\cdots, n_T \}$, in the target domain, $\mathcal{X}_T$, due to the distribution discrepancy, ~\ie, $p(X_S)\neq p(X_T)$. In order to address this problem, previous works usually retrain the whole network in the target domain $\mathcal{X}_T$ with weights initialized by $f_S$ in the source domain~\cite{paoletti2019deep,li2019deep}. However, to retrain the network in the target domain, $\mathcal{X}_T$, we have to label a sufficient amount of data in $\mathcal{X}_T$, which would introduce additional non-trivial and time-consuming burden.

In this paper, we seek to bridge the gap between the source and target domains by realizing a new transfer learning scheme through shared space. By projecting the data from both domains into a proper shared space $\mathcal{A}$, we aim to minimize the domain discrepancy and train a general classifier $f_{ST}$ that works in both domains. Hence the first challenge is how to find such shared space given only the labels in the source domain. 

%such that a general classifier $f_{ST}$ could be learnt and utilized in both domains?
%a general classifier $f_{ST}$ could be learnt on intrinsic representations shared by pixels among different domains. 
We approach this problem from a unique angle,~\ie, by exploiting the physical characteristics inherent in data in both the source and target domains. As discussed in Sec.~\ref{sec:intro}, each pixel in an HSI tends to cover more than one constituent materials. Thus, these mixtures can be modeled as a linear combination of a few spectral bases with the corresponding mixing coefficients (or abundance). Mathematically, they can be expressed as,
\begin{align}
\begin{split}\label{equ:content}
&\mathbf{x}_S =\mathbf{a}_S B_S
\end{split}\\
\begin{split}\label{equ:style}
&\mathbf{x}_T = \mathbf{a}_T B_T
\end{split}
\end{align}
where $\mathbf{a}_S\in\mathbb{R}^{1\times c}$ and $B_S\in\mathbb{R}^{c\times l}$ denote the coefficient vector (or abundance vector) and the spectral bases of the given pixel $\mathbf{x}_S\in\mathbb{R}^{1\times l}$ in the source domain, respectively. $\mathbf{a}_T\in\mathbb{R}^{1\times c}$ and $B_T\in\mathbb{R}^{c\times l}$ denote the abundance vector and the spectral bases of the given pixel $\mathbf{x}_T\in\mathbb{R}^{1\times l}$ in the target domain, respectively. $l$ and $c$ denote the number of spectral bands of the HSI and the number of the spectral bases, respectively. Assume that $\mathbf{x}_S$ and $\mathbf{x}_T$ belong to the same class, they may behave distinctively due to different illumination conditions, so do their spectral bases $\mathbf{a}_S$ and $\mathbf{a}_T$. However, for the same class, the abundance vectors $\mathbf{a}_S$ and $\mathbf{a}_T$, indicating the proportions of the bases, should be similar in both domains, hence serving as the potential shared representations and forming the corresponding shared space to facilitate the training of a general classifier $f_{ST}$. We refer to this space as the shared abundance space. See Sec.~\ref{sec:evaluate_tf} for an illustration of the effectiveness of the abundance space representation. Given such a potential shared space, the second challenge is then how to project the data from both domains into the shared abundance space, such that the shared representations could be extracted.

We tackle this problem with three physical properties inferred from the characteristics of the data,~\ie, 1) the non-negative and sum-to-one constraints on the abundance $\mathbf{a}_S$ and $\mathbf{a}_T$; 2) the affine correlations between $B_S$ and $B_T$; and 3) the uncertainty residing in the usage of $\mathbf{a}_S$ ($\mathbf{a}_T$) to represent $\mathbf{x}_S$ ($\mathbf{x}_T$). Since there is no ground truth label in the target domain, we need to exploit these intrinsic properties of data in order to regulate the solution space. The first item is inferred from the properties of the abundance. Since the abundance vector indicates the proportions the spectral bases in making up the mixed pixels, the shared space should be constrained with the non-negative and sum-to-one physical constraints.  %the second and the third items, to find the correspondence between the pixels belonging to the same classes in different domains. 
The second item is inferred from the properties of the spectral bases.  Although the spectral bases of the mixed pixels belonging to the same classes may be distinctive, they tend to hold an affine relationship~\cite{Borsoi2020HSIReview},~\ie, 
\begin{equation}
	B_S = \mathbf{c}B_T+\mathbf{d}
	\label{equ:affine}
\end{equation} 
In addition, since the shared abundance vector can represent both $\mathbf{x}_S$ and $\mathbf{x}_T$, it has the potential to be the optimal representation that maximizes the reduction of the uncertainty of both $\mathbf{x}_S$ and $\mathbf{x}_T$. %To further increase the classification accuracy, the information of different dimensions of the pixels should also be considered during training procedure. 

\section{Proposed Approach}
\label{sec:proposed}
\begin{figure}[htp]
	\begin{minipage}{1\linewidth}
		\includegraphics[width=1\linewidth]{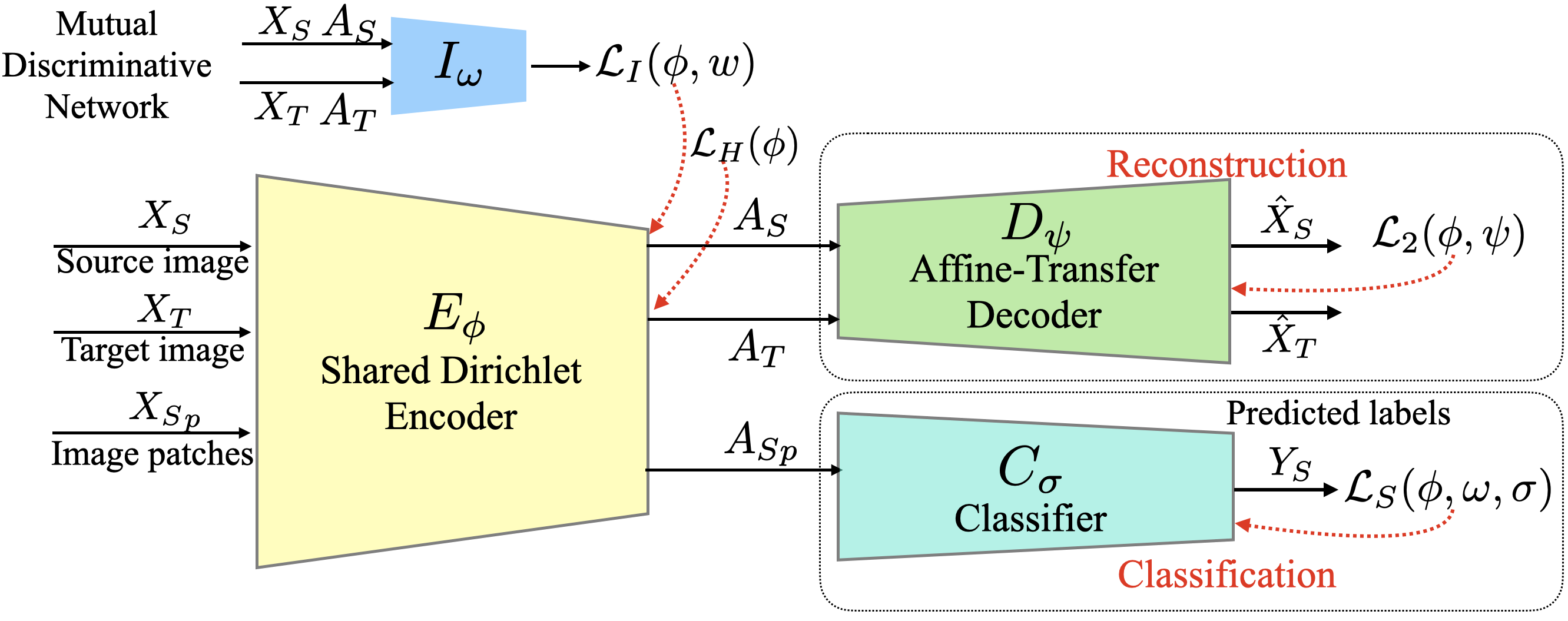}
		\caption{Flowchart of the proposed physically-constrained transfer learning approach for HSI classification.}
		\label{fig:flow}
	\end{minipage}
\end{figure}

%TODO abundance domain to abundance space?
We propose a physically-constrained transfer learning approach for HSI classification across different domains through the extraction of the shared abundance space, as shown in Fig.~\ref{fig:flow}. The network architecture mainly consists of four unique modules,~\ie, 1) a shared sparse Dirichlet-encoder $E_{\phi}$ which projects the data from different domains into a shared abundance space with the non-negative and sum-to-one physical constraints, 2) an affine-transfer decoder $D_{\psi}$ to embed the potential affine relationship between the source and target domains, 3) a mutual discriminative network $I_{\omega}$ to enforce the correspondence between the pixels in the source and target domains such that the shared intrinsic representations could be extracted from pixels belonging to the same class, and 4) a densely-connected 3D-CNN-based classifier $C_{\sigma}$ concatenated with the shared encoder to increase the generalization, representative, and discriminative power of the network.

\subsection{Architecture}
\label{sec:proposed:structure}
%As shown in Fig.~\ref{fig:flow}, the network structure includes four major components, 1) a shared encoder $E_{\phi}$, 2) an affine-transfer decoder $D_{\psi}$, 3) a mutual discriminative network $I_{\omega}$ and 4) a classifier $C_{\sigma}$.
%The architecture of proposed method consists of three subtask branches constructed by the four components, $E_{\phi}$, $D_{\psi}$, $I_{\omega}$ and $C_{\sigma}$. The first branch is constructed by  
To realize the transfer learning scheme for generalization purpose, both the representation and discrimination power of the network need to be boosted. %two subtasks are implemented,~\ie, 1) the reconstruction subtask and 2) the classification subtask. 
% TODO Do I need to explain why the network is set in this way?

The effective representation is realized with the proposed encoder-decoder structure consisting of $E_{\phi}$ and $D_{\psi}$. This structure is designed according to the linear mixing model shown in Eqs.~\eqref{equ:content} and \eqref{equ:style}, where the representations extracted by the encoder $E_{\phi}$ correspond to the abundance of the images and the decoder $D_{\psi}$ embeds the spectral bases of the images in Eqs.~\eqref{equ:content} and \eqref{equ:style}. Let's define the input image domain of the encoder-decoder network  as $\mathcal{G} =\{X_S, X_T\}$, where  $X_S\in\mathcal{X}_S$, $X_T\in\mathcal{X}_T$ denote the source image and the target image, respectively. Similarly, the output domain is denoted with $\hat{\mathcal{G}}=\{\hat{X}_S, \hat{X}_T\}$, where $\hat{X}_S$ and $\hat{X}_T$ denote the reconstructed source and target images, respectively. The shared Dirichlet encoder $E_{\phi}:\mathcal{G}\rightarrow\mathcal{A}$, projects the input data from the image domain to the so called ``abundance space'' $\mathcal{A} = \{A_S, A_T\}$, where the representations $A_S$ and $A_T$ are extracted (latent variables on the hidden layer) from the source and target images, respectively. The affine-transfer decoder $D_{\psi}:\mathcal{A}\rightarrow\hat{\mathcal{G}}$ reconstructs the images back from the representations to the output image domain $\hat{\mathcal{G}}$ with the spectral bases of both the source and the target images. 

The affine-transfer decoder of the network not only learns the spectral bases of the images in the source and target domains, but also finds their potential affine relationship as demonstrated in Eq.~\eqref{equ:affine}. To further regularize the representations influenced by the affine relationship, we propose a mutual discriminative network $I_{\omega}$ and concatenate it with the encoder. $I_{\omega}$ maximizes the mutual information between the representations $A_S$ ($A_T$), and their own corresponding input $X_S$ ($X_T$). In this way, the shared intrinsic representations are able to be extracted between the pixels belonging to the same classes in different domains. 

Besides the effective representation scheme described above, the network architecture also realizes  effective discrimination by the proposed densely-connected 3D convolutional neural network (CNN), $C_{\sigma}$, which is concatenated to the shared Dirichlet-encoder. The classifier is trained with the patches of the source images ${X_S}_p$ to utilize both the spatial and spectral information of the data. It can be defined as $f_S:{X_S}_p\rightarrow {A_S}_p\rightarrow\mathcal{Y}_S$, which is built as a mapping function between the inputs of the source image ${X_S}_p$ and their labels $Y_S$. By sharing the same encoder, the classifier $C_{\sigma}$ further increases the discriminative power of the network. More important, since the weights of the classifier are updated on the domain-invariant abundance space $\mathcal{A}$, the generalization capacity of the classifier is largely enhanced such that it can be directly utilized to categorize target images without data labeling or network retraining.

The detailed network structure is further elaborated in Sec.~\ref{sec:proposed:dir}, Sec.~\ref{sec:proposed:decoder} and Sec.~\ref{sec:proposed:classifier}. %The fully-connected layer is the building block of the encoder-decoder network, and 3D-CNN is the building block of the classification network. Please refer to 
Sec.~\ref{sec:detail} also includes implementation details. 

%With this structure, the network is able to project the data from image domains to the shared abundance domain and learn the potential physical transfer relationship between the source and target domains. 

%Mutual information is one of many quantities that measures how much one random variables tells us about another. It is a dimensionless quantity with (generally) units of bits, and can be thought of as the reduction in uncertainty about one random variable given the knowledge of another

%2) The feature correspondence matching subtask relies on both the affine-transfer decoder $D_{\psi}$ and the mutual discriminative network $I_{\omega}$. The affine-transfer decoder not only learns the spectral bases of the images in the source and target domains, but also find their potential affine relationship demonstrated in Eq.~\eqref{equ:affine}. However, to extract proper shared intrinsic representations, the affine relationship should be built between the pixels belonging to the same classes in different domains. Thus we propose a mutual discriminative network 

%%This correspondence is further elaborated below. 
%The classification structure shares the same encoder with the reconstruction branch, such that the extracted shared inherent representations from different domains contribute more to the classification result. 
\subsection{Shared Sparse Dirichlet Encoder}
\label{sec:proposed:dir}
\begin{figure}[htp]
	\centering
	\includegraphics[width=0.7\linewidth]{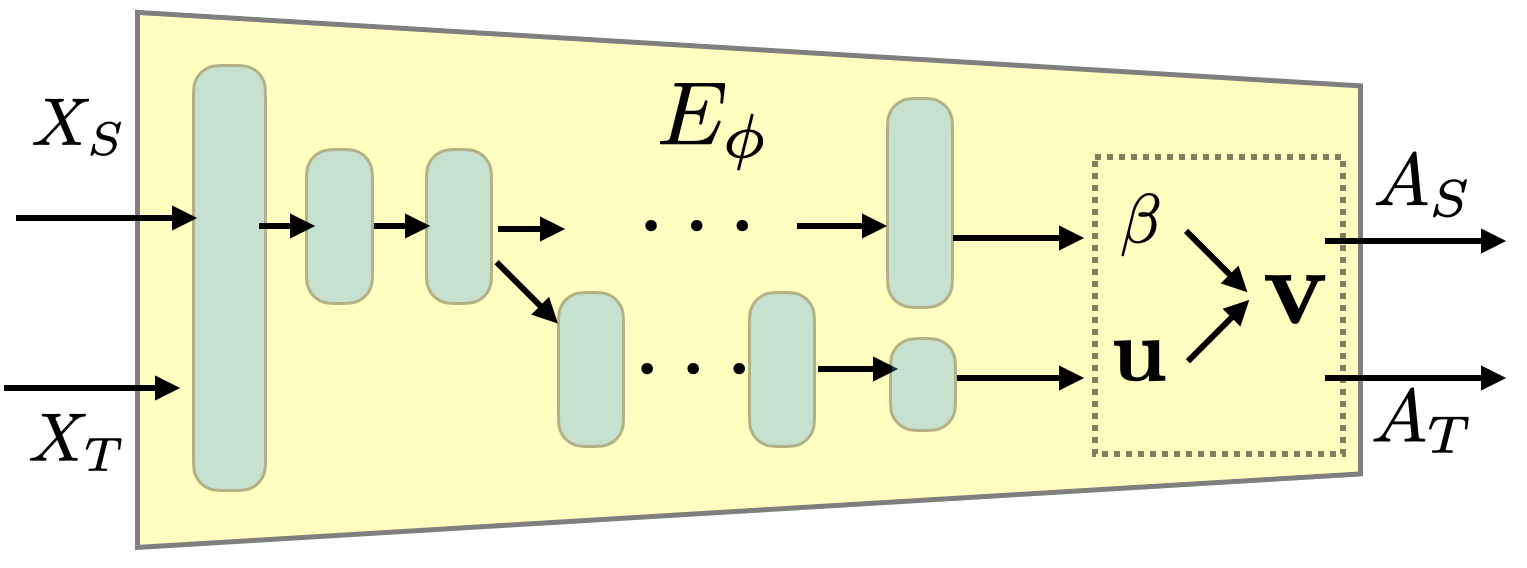}
	\caption{Structure of the proposed shared sparse Dirichlet encoder.}
	\label{fig:flow_encoder}
\end{figure}
As discussed in Secs.~\ref{sec:intro} and ~\ref{sec:formulate}, both the source image $X_S$ and the target image $X_T$ can be represented by a linear combination of the spectral bases and their corresponding abundance vectors as shown in Eqs.~\eqref{equ:content} and~\eqref{equ:style}. Although the spectral bases may vary due to spectral variability~\cite{Borsoi2020HSIReview}, their abundance vectors, which indicate how much contribution the spectral bases have in constructing a given pixel should be the similar for the same class. Thus, the abundance vectors indicate the domain-invariant representations we aim to find. Please refer to Sec.~\ref{sec:evaluate_tf} for an illustration of the effectiveness of abundance-based representation. Due to the physical properties of the abundance, the representations should meet two physical constraints,~\ie, non-negativity and sum-to-one. We extract such representations by projecting the data from both the source and target domains to the shared abundance space, with sparse Dirichlet-encoder shown in Fig.~\ref{fig:flow_encoder}. 

The Dirichlet-encoder is constructed with the stick-breaking process ~\cite{nalisnick2016deep,qu2018unsupervised}, which can be illustrated as breaking a unit-length stick into $c$ pieces. The pieces would follow a Dirichlet distribution, and thus the two physical constraints, non-negativity and sum-to-one are naturally met. A single piece, $a_j$, can be expressed as
\begin{equation}
a_j =\left\{
\begin{array}{ll}
v_1 \quad & \text{for} \quad j = 1\\
v_j\prod_{o<j}(1-v_o) \quad &\text{for} \quad j>1 , 
\end{array}\right.
\label{equ:stick}
\end{equation}
where $v_j$ is drawn from a Kumaraswamy distribution, \ie, $v_j\sim \text{Kuma}(u, 1,\beta)$ as shown in Eq.~\eqref{equ:draw}
\begin{equation}
%v\sim (1-u^\frac{1}{\beta})^{\frac{1}{a}}
v_j\sim (1-(1-u^\frac{1}{\beta})).
\label{equ:draw}
\end{equation}
The stick-breaking process in Eq.~\eqref{equ:stick} can be explained as sequentially breaking $v_j$ from the remaining stick $\prod_{o<j}(1-v_o)$. The constructed representation, $0\leq a_j\leq 1$, would naturally meet the non-negativity constraint. As the number of pieces $c$ increases, the summation of $a_j$ is close to one,~\ie, $\sum_{j=1}^{j=c}a_j\approx 1$, which would enforce the representations to meet the sum-to-one physical constraint.

Based on the fact that each pixel usually consists of a few spectral bases, we further encourage the representations to be sparse with normalized entropy function~\cite{huang2017sparse}, defined in Eq.~\eqref{equ:entropyfun} with $p=1$. 
\begin{equation}
H_{p}(\mathbf{a}) = -\sum_{i=1}^{N}\frac{{\vert a_i}_\rightarrow\vert^p}{\Vert {a_i}_\rightarrow \Vert_p^p}
\log\frac{{\vert a_i}_\rightarrow \vert^p}{\Vert {{a_i}_\rightarrow} \Vert_p^p}. 
\label{equ:entropyfun}
\end{equation}
Even a vector is sum-to-one, its normalized entropy function could decrease monotonically when it becomes sparse. This could not be satisfied with traditional $l_1$ regularization or Kullback-Leibler divergence.

%Since the Dirichlet-encoder has the sum-to-one property, the traditional widely used $l_1$ regularization or Kullback-Leibler divergence could not be used to enhance the sparsity. Instead, we adopt the normalized entropy function~\cite{huang2017sparse}, defined in Eq.~\eqref{equ:entropyfun}, which decreases monotonically when the data become sparse.
% \begin{equation}
% \mathcal{H}_{p}(\mathbf{s}) = -\sum_{i=1}^{N}\frac{{\vert s_i}_\rightarrow\vert^p}{\Vert {s_i}_\rightarrow \Vert_p^p}
% \log\frac{{\vert s_i}_\rightarrow \vert^p}{\Vert {{s_i}_\rightarrow} \Vert_p^p}. 
% \label{equ:entropyfun}
% \end{equation}
% We choose $p=1$ for efficiency.  
The objective function for sparse loss can then be defined as 
\begin{equation}
\label{equ:sparse}
\mathcal{L}_H(\phi) = H_{1}(E_\phi(X_S)) + H_{1}(E_\phi(X_T)).
\end{equation}

\subsection{Affine-Transfer Decoder and Mutual Discriminative Network}
\label{sec:proposed:decoder}
With the shared sparse Dirichlet-encoder, we are able to project the data from the image domain to the abundance space. However, since data distributions of the source and target domains are different and there is no label information for the images in the target domain, we still need to match data belonging to the same classes but in different domains to extract their effective shared intrinsic representations. For this purpose, more physical constraints are exploited to further regularize the solution space. These include the affine correlations between different domains and the uncertainty reduction of the representations. In our network design, we enforce these two physical constraints with the proposed affine-transfer decoder and the mutual discriminative network. 

%In the defined problem, the pixels in the target domain does not have ground truth labels, thus the inherent representations of the same type of objects in different domains should be well-matched for better classification. In our network design, we enforce such correspondence with affine-transfer decoder and mutual discriminative network based on mutual information.  

\begin{figure}[htp]
	\centering
	\subfloat[]{\includegraphics[width=0.35\linewidth]{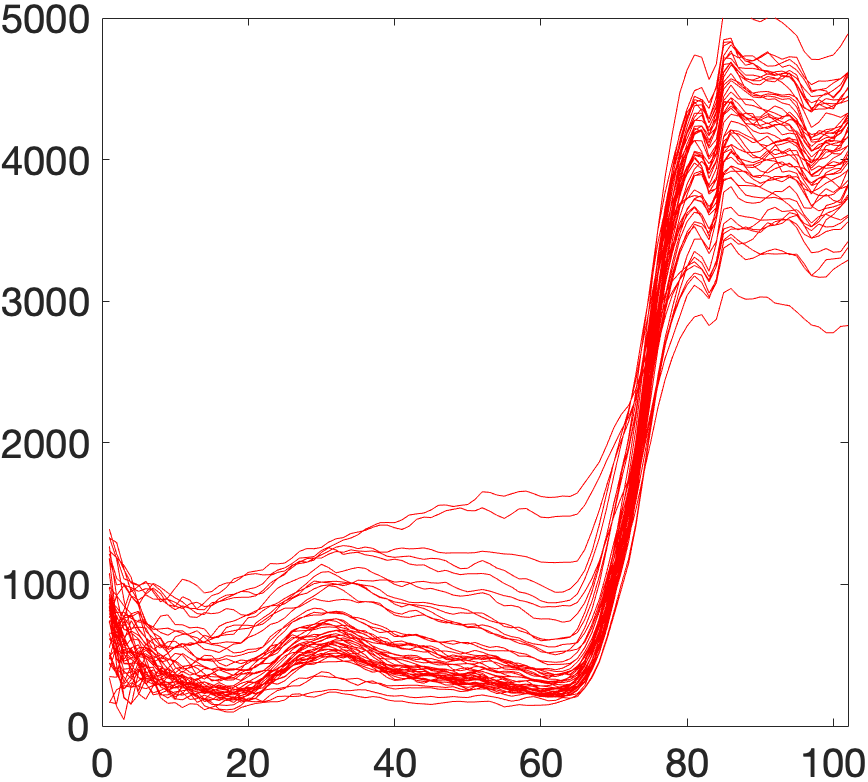}}
	\subfloat[]{\includegraphics[width=0.35\linewidth]{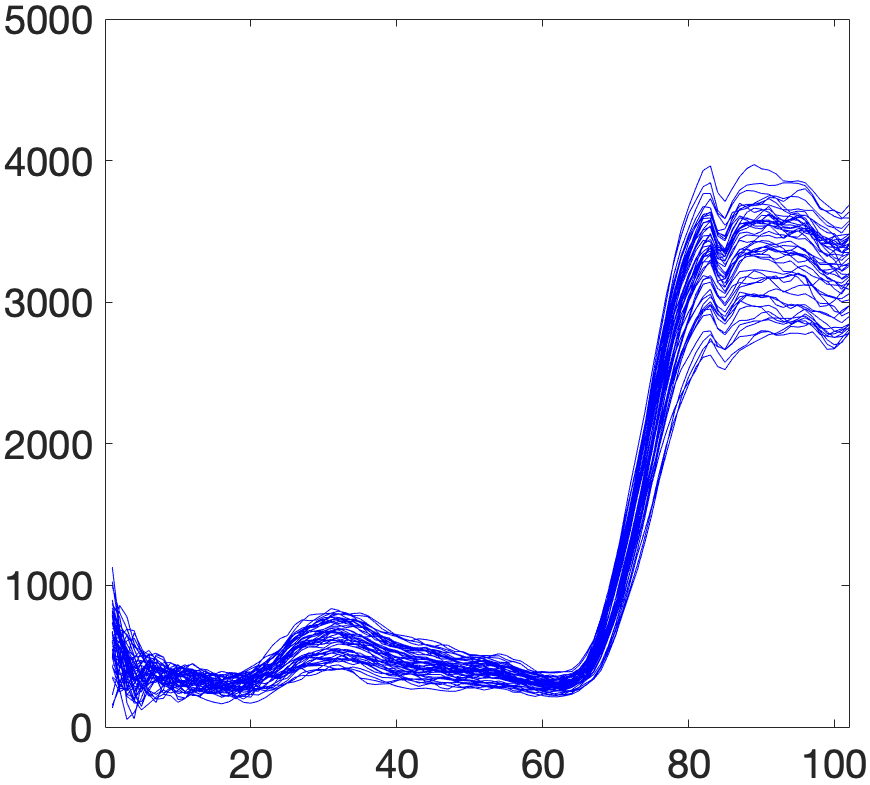}}
	\caption{The spectral reflections of data from the same class in (a) the source domain and (b) the target domain.}
	\label{fig:spectral}
\end{figure}

\subsubsection{Affine-Transfer Decoder}
\begin{figure}[htp]
	\centering
	\includegraphics[width=0.6\linewidth]{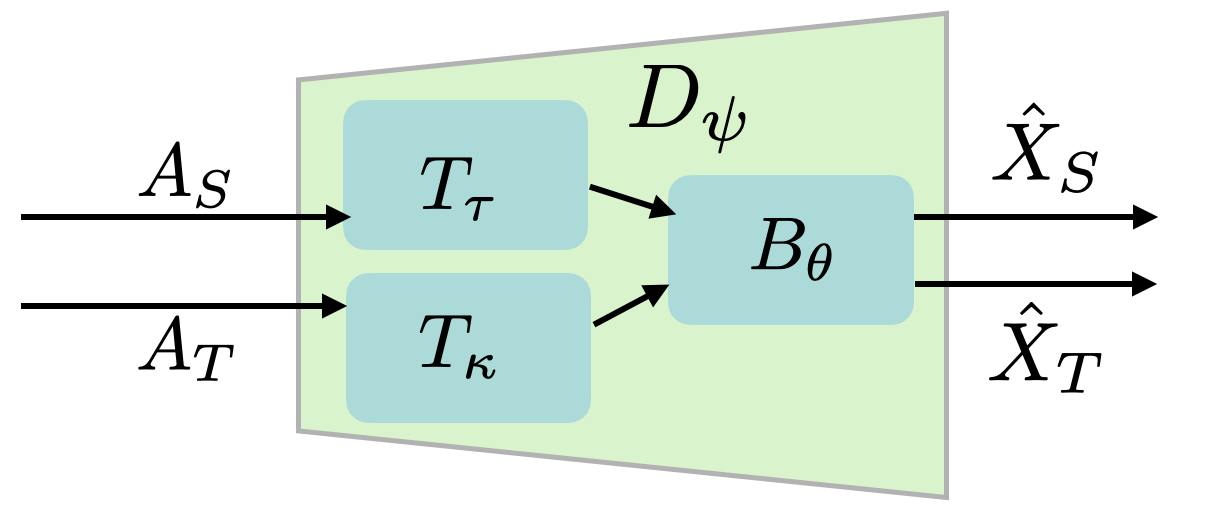}
	\caption{Structure of the proposed affine-transfer decoder.}
	\label{fig:flow_affine}
\end{figure}

Due to different environmental and illumination conditions during data acquisition, the HSI pixels belonging to the same class but different domains may possess different spectral characteristics as shown in Fig.~\ref{fig:spectral}. We can observe that although the reflectances in the source and target domains are different, their relationship can be approximated with an affine transfer,~\ie, the spectral reflectance in one domain is able to be represented as the affine transformation of that in another domain. Similarly, their spectral bases can be modeled with affine relationship as shown in Eq.~\eqref{equ:affine}. Since the pixels belonging to the same classes in different domains have different distributions, if we use a general decoder, the difference between the distributions may be reflected on the representations, making it difficult to find the shared representations. In this paper, we propose an affine-transfer decoder to address this problem. The affine-transfer decoder is built to learn the spectral bases of both domains and simulate their physical affine relationship. In this way, the network has the potential to learn the shared intrinsic representations. The network structure is shown in Fig.~\ref{fig:flow_affine}. 

%Due to different environmental and illumination conditions, the reflectance of the same type of objects may have different reflectance although they possess the same spectral basis~\cite{drumetz2019spectral}. Since the same type of objects may have drastic spectral changes in different domains, we assume that the bases of the source HSI $B_S$ and target HSI $B_T$ have an affine relationship.  To capture the correlated representations, 
To improve the flexibility of the network, instead of directly adopting the affine-transfer model, we introduce a shared bases $B_\theta$ to build the relationship between the spectral bases $B_S$ and $B_T$ of the source and target domains. Then the affine-transfer decoder is defined as  
\begin{align}
\begin{split}\label{equ:bc}
&T_\tau(B_\theta) = \mathbf{c}_{\tau}B_\theta+ \mathbf{d}_\tau
\end{split}\\
\begin{split}\label{equ:bs}
&T_\kappa(B_\theta) = \mathbf{c}_{\kappa}B_\theta + \mathbf{d}_\kappa,
\end{split}
\end{align}
%TODO on the domain instead of in the domain
where $\mathbf{c}_{\tau}$, $\mathbf{d}_\tau$ and $\mathbf{c}_{\kappa}$, $\mathbf{d}_\kappa$ are the network weights. $T_\tau(B_\theta)$ and $T_\kappa(B_\theta)$ share the same basis weights $B_\theta$ and they correspond to $B_S$ and $B_T$ in Eqs.~\eqref{equ:content} and ~\eqref{equ:style}, respectively. The affine-transfer decoder can not only extract the spectral bases from images on both domains but also learn their potential affine relationship. Compared to the encoder-decoder network with general decoder, the proposed model with the affine-transfer decoder allows the network to extract shared intrinsic representations even they hold different statistic distributions. 
\subsubsection{Mutual Discriminative Network}
\begin{figure}[htp]
	\centering
	\includegraphics[width=0.55\linewidth]{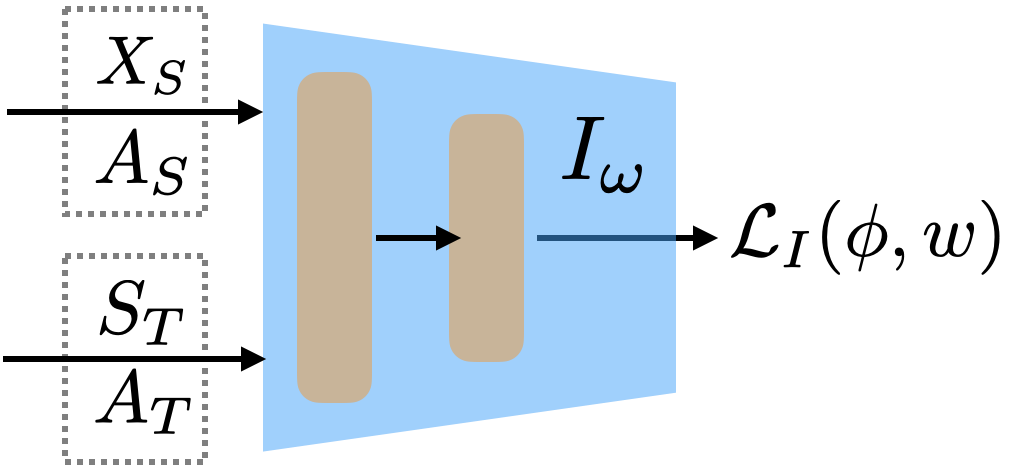}
	\caption{Structure of the proposed mutual discriminative network.}
	\label{fig:flow_mi}
\end{figure}

With the affine-transfer decoder, the network has the potential ability to extract shared intrinsic representations. However, to avoid negative transfer, the affine-transfer relationship should be built between the same classes in different domains. We enforce such correspondence by regularizing the solution space %with the uncertainty reduced by the representations. This is realized 
by maximizing the mutual information (MI) between the representations and their raw input. This is realized by the proposed mutual discriminative network. The network structure is shown in Fig.~\ref{fig:flow_mi}.

%By maximizing the mutual information (MI) between the representations and their own input, the network tends to find the shared optimal representations of the pixels belonging to the same class, which would automatically match the correspondence between different domains.

In our network design, the classification accuracy would rely heavily on the extracted representations in the abundance space. The optimal representations should not only reconstruct the image well but also reduce the uncertainty of the image to the maximum extend. By finding the optimal representations for each pixel, the shared  representations could be automatically matched between the pixels belonging to the same class in different domains. We introduce mutual information (MI) to maximize the reduced uncertainty of the image. MI has been widely used for multi-modality registrations. It measures how much uncertainty of one variable is reduced given the other variable with Shannon-entropy. The shared intrinsic representation should reduce the uncertainty of both the source and target domains. Based on the work of Belghazi~\etal~\cite{belghazi2018mine}, the MI can be estimated through a neural network according to the lower bound of KL-divergence based on the Donsker-Varadhan (DV) representation. Thus, we also define a discriminative network $I_w$ to maximize the average MI between each representation and its own input, \ie, $I_\omega(X_S,E_{\phi}(X_S))$ and $I_\omega(X_T, E_{\phi}(X_T))$. However, in our case, instead of estimating the exact MI, we only use MI to regularize the solution space. Therefore, an alternative lower bound based on Jensen-Shannon is introduced, which is more stable than the DV-based objective function~\cite{Hjelm2018Learning}.
% The MI is usually calculated with the KL-divergence’s lower bound based on the Donsker-Varadhan (DV) representation. 
% , which allows estimating MI through a neural network.
%Thus  we maximize the average MI between the representations and their own inputs, i.e., $\mathcal{I}(X_S,E_{\phi}(X_S))$ and $\mathcal{I}(X_T, E_{\phi}(X_T))$ through the same discriminative network $\mathcal{T}_w$ simultaneously. The MI is usually calculated with the KL-divergence’s lower bound based on the Donsker-Varadhan (DV) representation. However, since in our case, the MI is only used to regularize the solution space, we introduce an alternative lower bound based on Jensen-Shannon, which could not estimate the exact MI, but works more stable than the DV-based objective function~\cite{Hjelm2018Learning}.

Taking the image $X_S$ in the source domain as an example, the objective function for the mutual discriminative network $I_{\omega}:\mathcal{G}\times\mathcal{A}\rightarrow\mathbb{R}$ is defined as 
\begin{equation}
\label{equ:dvmi}
\begin{array}{ll}
\mathcal{I}_{\phi,\omega}(X_S, E_{\phi}(X_S)) &=
E_{\mathbb{P}}[-sp(-I_{\omega}(X_S,E_{\phi}(X_S))]\\
&- E_{\mathbb{P}\times\tilde{\mathbb{P}}}[sp(I_{\omega}(X_S',E_{\phi}(X_S))],\\
\end{array}
\end{equation}
where $\omega$ denotes the network weights, and $sp(x) =\log(1+e^x)$. Note that $X_S'$ is an input with samples drawn from the distribution $\tilde{\mathbb{P}}=\mathbb{P}$, which is a negative sample generated by randomly shuffling the input data. 
%  by randomly shuffling the input data. The term carrying the shuffling data is called the negative sample. 
The objective MI function for both the source and target image is defined as
\begin{equation}
\label{equ:mi}
\begin{array}{ll}
\mathcal{L}_{I}(\phi,\omega) = \mathcal{I}_{\phi,\omega}(X_S, E_{\phi}(X_S)) + \mathcal{I}_{\phi,\omega}(X_T, E_{\phi}(X_T))
\end{array}
\end{equation}
%And the reconstruction error is defined as $\mathcal{L}_{2}$ for both source and target samples. 

\subsection{Densely-Connected 3D-CNN based Classifier}
\label{sec:proposed:classifier}
\begin{figure}[htbp]
	\centering
	\includegraphics[width=0.8\linewidth]{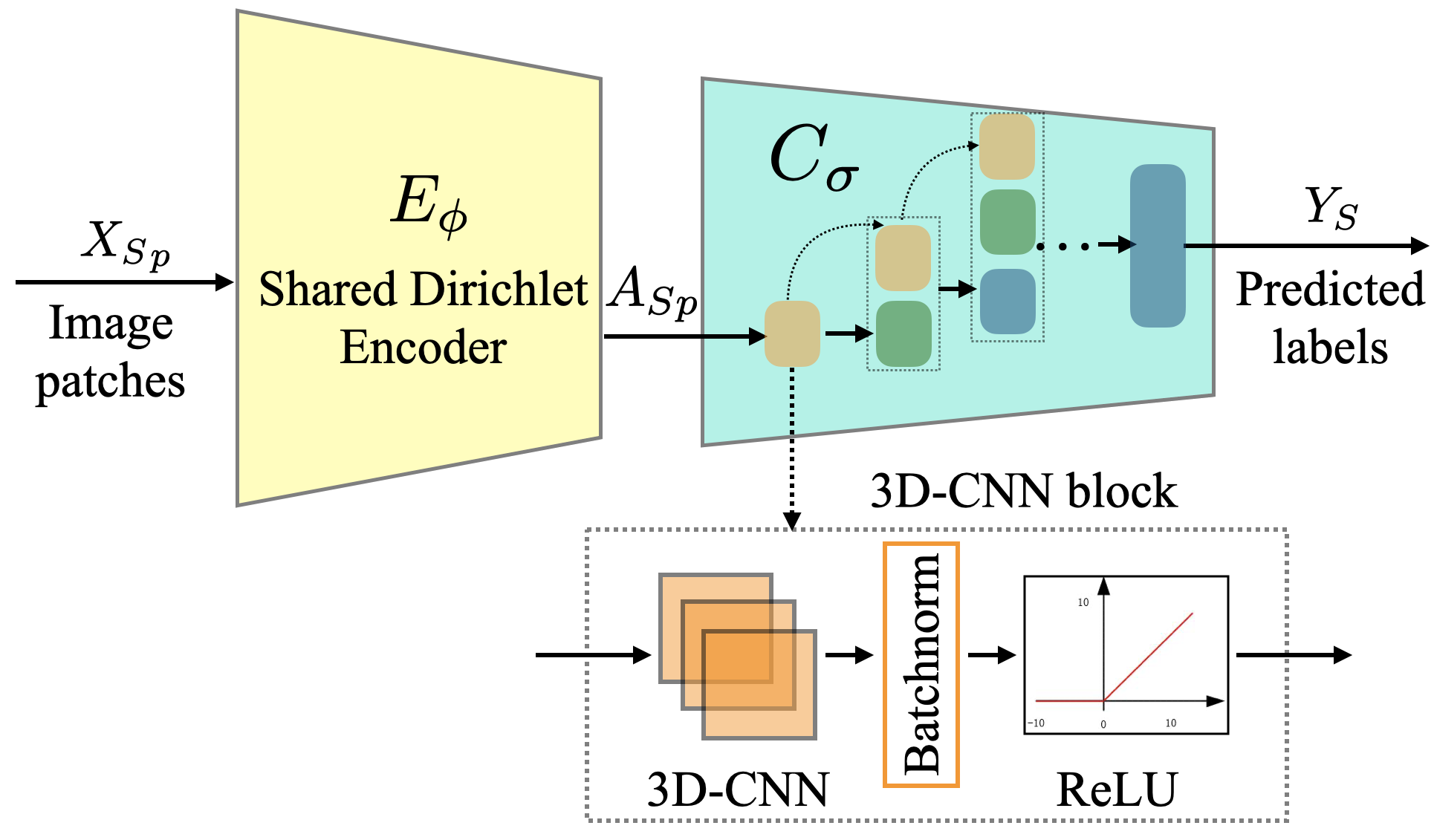}
	\caption{Structure of the proposed densely connected 3D-CNN classifier.}
	\label{fig:flow_class}
\end{figure}

To train a classifier that works in both domains with better classification accuracy, a densely-connected 3D convolutional neural network (3D-CNN) based classifier $C_\sigma$ is proposed and concatenated to the shared Dirichlet-encoder. In this way, the classifier is mainly trained on the domain invariant abundance space, which would largely increase the generalization of the network. The network structure of the classifier is shown in Fig.~\ref{fig:flow_class}.

The basic block of the classifier consists of a 3D-CNN, a batchnorm layer~\cite{ioffe2015batch} and a relu activation function~\cite{nair2010rectified}, as shown at the lower part of Fig.~\ref{fig:flow_class}. With 3D-CNN, the network is able to categorize a pixel not only considering its own information but also its adjacent pixels. The batchnorm and the relu are used to accelerate the training speed. To increase the discriminative power of the network, the 3D-CNN blocks are densely connected, \ie, each layer is fully connected with all its subsequent layers. 

The last layer of the classifier is a fully-connected layer following a dropout operation~\cite{srivastava2014dropout} to prevent overfitting. A softmax activation function and cross entropy objective function are introduced to predict the class label. Given the ground truth label, $Y_S$, from the source domain and the predicted label $Y_{PS}=C_{\sigma}(E_{\phi}(X_{Sp}))$ from the network, the objective function is defined as 
\begin{equation}
\label{equ:class}
\begin{array}{ll}
\mathcal{L}_{S}(\phi,\sigma) = -\sum Y_S \log(Y_{PS})
\end{array}
\end{equation}

\subsection{Implementation Details}
\label{sec:detail}
The objective function of the proposed architecture includes four components, a reconstruction loss $\mathcal{L}_2(\phi,\psi)$, a sparse constraint $\mathcal{L}_H(\phi)$, the mutual discriminative loss $\mathcal{L}_{I}(\phi,\omega)$, and a classification loss $\mathcal{L}_{S}(\phi,\sigma)$. The reconstruction loss is defined as 
\begin{equation}
\begin{array}{ll}
\mathcal{L}_2(\phi,\psi)  &=  \Vert D_{\psi}(E_\phi(X_S))- X_S\Vert_2\\
&+\Vert D_{\psi}(E_\phi(X_T))- X_T\Vert_2,
\end{array}
\label{equ:l2}
\end{equation}
where $\Vert\cdot\Vert$ denotes the $l_2$ norm. The definition of the other losses are defined in Eqs.~\eqref{equ:sparse},~\eqref{equ:mi}, and~\eqref{equ:class}, respectively. The objective function for the proposed network can then be expressed as
\begin{equation}
\label{equ:opt}
\mathcal{L}(\phi,\psi,\omega,\sigma) = \mathcal{L}_2(\phi,\psi) + \alpha\mathcal{L}_H(\phi)+\lambda\mathcal{L}_{I}(\phi,\omega)+\mathcal{L}_{S}(\phi,\sigma),
\end{equation}
where $\alpha$ and $\lambda$ are the parameters that balance the trade-off among different losses. Since the reconstruction and classification loss are equally important, their parameters are fixed to 1.

The reconstruction and the classification network are optimized together by back-propagation illustrated in Fig.~\ref{fig:flow} with red dashed lines. In this way, it not only enhances the representative power of the network but also the discriminative power of the network. Note that, the inputs for the reconstruction network are the source and target images, while the input to the classification network are the patches of the source images. 

During the training procedure, we are able to extract the shared intrinsic representations and train a general classifier on them. In the testing procedure, we can directly get the predicted labels by feeding the target image patches into the network. That is, no data labeling or network retraining are required to classify the image in the target domain.

The reconstruction network is constructed with a few fully-connected layers, and the classifier consists of a few 3D-CNN layers with kernel size $7\times 7\times 3$. The number of layers and nodes in the proposed network are shown in Table~\ref{tab:layers}. 
\begin{table}[htb]
	\caption{Layers and nodes in the proposed network.}
	\label{tab:layers}
	\begin{center}
		\begin{tabular}{c|cccccc}
			%		  \begin{tabular}{p{1cm}|p{1cm} p{0.5cm} p{0.6cm} p{0.55cm}p{0.5cm}p{2.4cm}}
			\hline
			&$E_\phi$&$I_w$&$T_\tau/$$T_\kappa$&$B_\theta$&3D-CNN\\
			\hline
			layers &6&2&1/1&2&5\\
			nodes &[3]&[13,1]&[11]&[11]&[12,32,12,12,30]\\
			\hline
		\end{tabular}
	\end{center}
\end{table}

\section{Experiments and Results} 
\label{sec:exp}
\subsection{Datasets}
To evaluate the performance of the proposed PCTL-SAS, we apply the method on three sets of source and target image pairs with different spectral characteristics that bring different levels of challenges to the classification problem. For each dataset pair, we choose the image with less number of samples as the source image and the other one as the target image. And the common classes shared by both the source and target images are selected for performance evaluation. To visualize the different levels of challenges brought by the three sets of data, we project the three dataset pairs into a two-dimensional space using SVD method, as demonstrated in Figs.~\ref{fig:visual_pavia:a},~\ref{fig:visual_houston:a}, and~\ref{fig:visual_datacube:a}, respectively. We can observe that there are some overlaps between the spectra of the same class on the source and target domains on the first dataset, while on the second and third dataset pairs, there exists a large discrepancy between the same class on the source and target domains. This domain discrepancy indicates that the second and the third datasets are more challenging than the first one. The details of the dataset pairs are described below.

\subsubsection{PaviaU-PaviaC} 
\begin{figure}
	\centering
	\subfloat[Pavia University (PU)]{\includegraphics[width=0.2\linewidth]{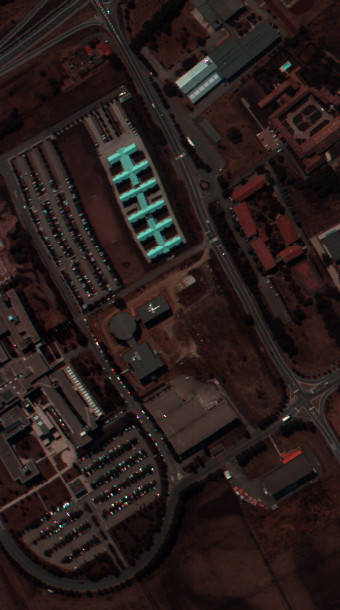}}\hfill
	\subfloat[GT of PU]{\includegraphics[width=0.2\linewidth]{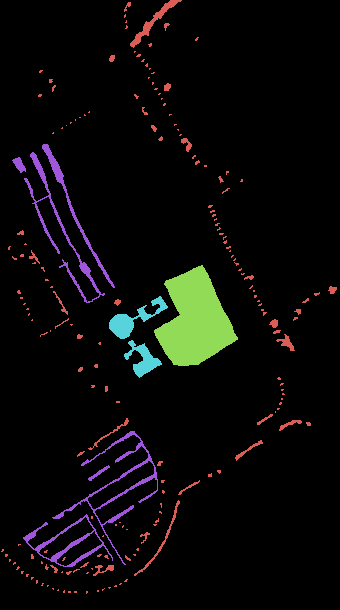}}\hfill
	\subfloat[Pavia Center (PC)]{\includegraphics[width=0.28\linewidth]{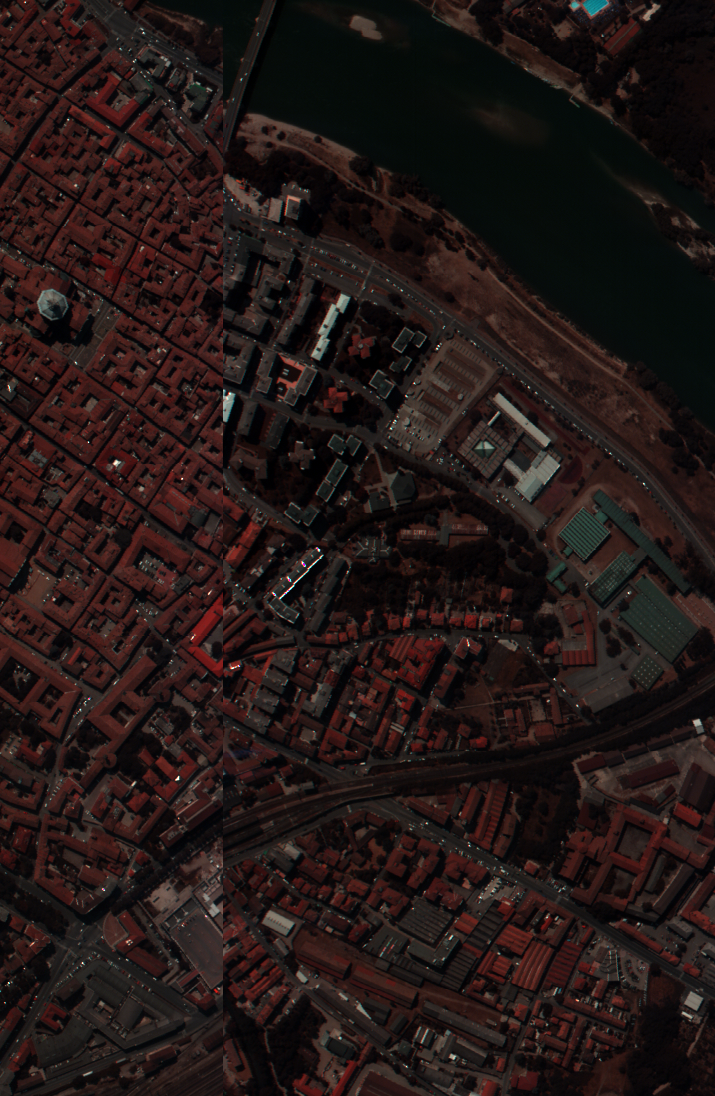}}\hfill
	\subfloat[GT of PC]{\includegraphics[width=0.28\linewidth]{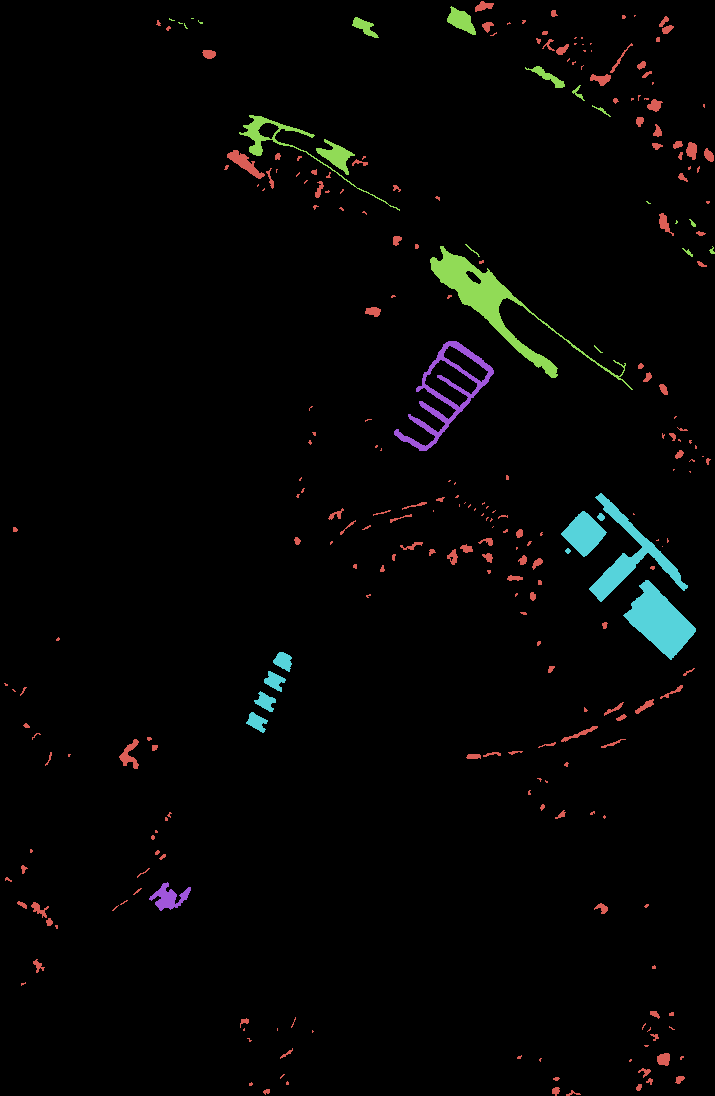}}
	\caption{(a) The source image Pavia University (PU). (b) The ground truth of PU. (c) The target image Pavia Center (PC). (d) The ground truth of PC.}
	\label{fig:pavia_input}
\end{figure}

The PaviaU-PaviaC data pair consists of the source image, Pavia University (PU), and the target image, Pavia Center (PC). The PU and PC images were acquired by the Reflective Optics System Imaging Spectrometer (ROSIS) sensor over Pavia, Italy~\cite{paviadata}. The spatial resolution of both images are 1.3m. The PU image has the dimension of $610 \times 340$, with 102 bands, and the last band removed. The PC image has the dimension of $1096 \times 715$ with 102 bands. Four shared classes,  tree, bare soil, bitumen, and bricks are chosen among the source and target images to evaluate the proposed method. The number of samples corresponding to each class is listed in Table \ref{tab:datasetPavia} and a sample pair of images are shown in Fig.~\ref{fig:pavia_input}.

\begin{table}[htb]
	\caption{Dataset description for the source image, Pavia University (PU) and the target image, Pavia Center (PC).}
	\label{tab:datasetPavia}
	\centering
		\begin{tabular}{c|c|c|c}
			\hline
			No. & Class & Source: PU & Target: PC\\
			\hline
			1 & Trees &3064& 7598\\
			2 & Bare soil & 5029 & 6584\\
			3 & Bitumen & 1330 & 7287\\
			4 & Bricks & 3682 & 2685\\
			\hline
			& Total & 13105 & 24154\\
			\hline
			& 1\%&131&0\\
			& 3\%&393&0\\
			& 5\%&655&0\\
			\hline
		\end{tabular}%
\end{table}

\subsubsection{Houston} 
\begin{figure}
	\centering
	\subfloat[Houston-2013 (HS13)]{\includegraphics[width=1\linewidth]{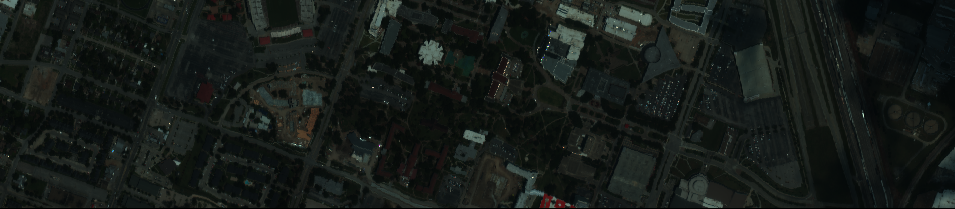}}\hfill
	\subfloat[GT of HS13]{\includegraphics[width=1\linewidth]{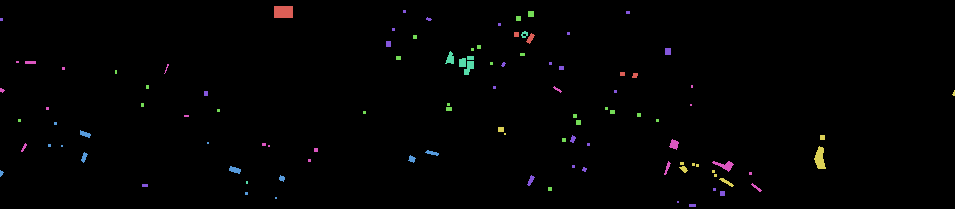}}\\
	\subfloat[Houston-2018 (HS18)]{\includegraphics[width=1\linewidth]{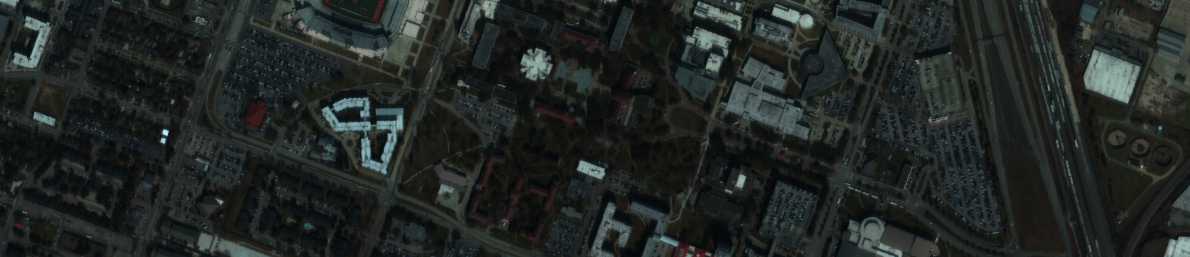}}\hfill
	\subfloat[GT of HS18]{\includegraphics[width=1\linewidth]{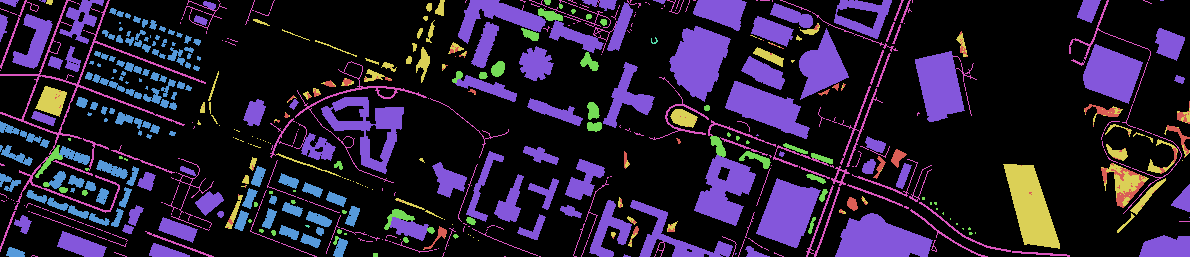}}
	\caption{(a) The source image Houston-2013 (HS13). (b) The ground truth of HS13. (c) The target image Houston-2018 (HS18). (d) The ground truth of HS18.}
	\label{fig:houston_input}
\end{figure}

The Houston pair is composed of the source image, Houston-2013 (H13)~\cite{debes2014hyperspectral}, and the target image Houston-2018 (H18)~\cite{le20182018}, which were collected over the University of Houston campus at different times with spatial resolution of 2.5m. The H13 and H18 have 144 and 48 bands, respectively. In H13, the overlapped 48 bands corresponding to that of the H18 are kept. The overlap area of two images are resized to the same dimension, $209 \times 955$. Seven shared classes are chosen. The images and the number of samples are shown in Fig.~\ref{fig:houston_input} and Table \ref{tab:datasetHouston}, respectively.

\begin{table}[htb]
	\caption{Dataset description for the source image, Houston-2013 (H13), and the target image, Houston-2018 (H18).}
	\label{tab:datasetHouston}
	\centering
		\begin{tabular}{c|c|c|c}
			\hline
			No. & Class & Source: H13 & Target: H18\\
			\hline
			1 & Grass healthy & 345 & 2099\\
			2 & Grass stressed & 375 & 7693\\
			3 & Trees & 365 & 3128\\
			4 & Water & 285 & 30\\
			5 & Residential buildings & 313 & 8271\\
			6 & Non-residential buildings & 408 & 49093\\
			7 & Road & 439 & 9558\\
			\hline
			& Total & 2530 & 79872\\
			\hline
			& 1\%&25&0\\
			& 3\%&75&0\\
			& 5\%&126&0\\
			\hline 
		\end{tabular}%
\end{table} 

\begin{table}[htb]
	\caption{Dataset description for the source image, Shanghai, and  the target image, Hanghzou}
	\label{tab:datasetDataCube}
	\centering
		\begin{tabular}{c|c|c|c}
			\hline
			No. & Class & Source: Hanghzou & Target: Shanghai\\
			\hline
			1 & Water  & 18043 & 123123\\
			2 & Ground/Building & 77450 & 161689\\
			3 & Plants & 40207 & 83188 \\
			\hline
			& Total &135700 &368000\\
			\hline
			& 1\%&1357&0\\
			& 3\%&4071&0\\
			& 5\%&6785&0\\
			\hline 
		\end{tabular}%
\end{table}

\begin{figure}
	\centering
	\subfloat[Hangzhou]{\includegraphics[angle=90,width=0.45\linewidth]{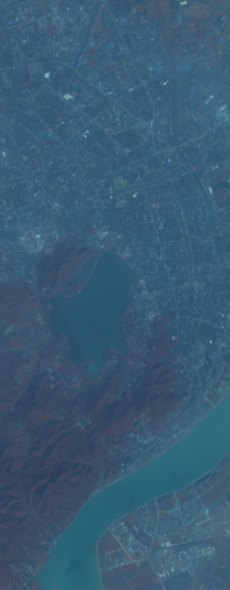}}\hfill
	\subfloat[GT of Hangzhou]{\includegraphics[angle=90,width=0.45\linewidth]{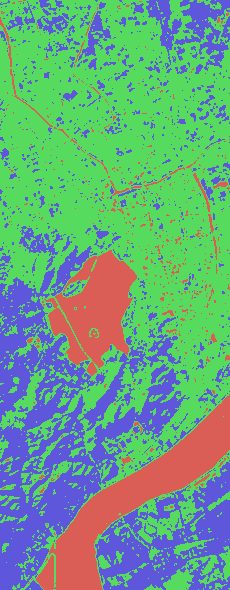}}\\
	\subfloat[Shanghai]{\includegraphics[angle=90,width=1\linewidth]{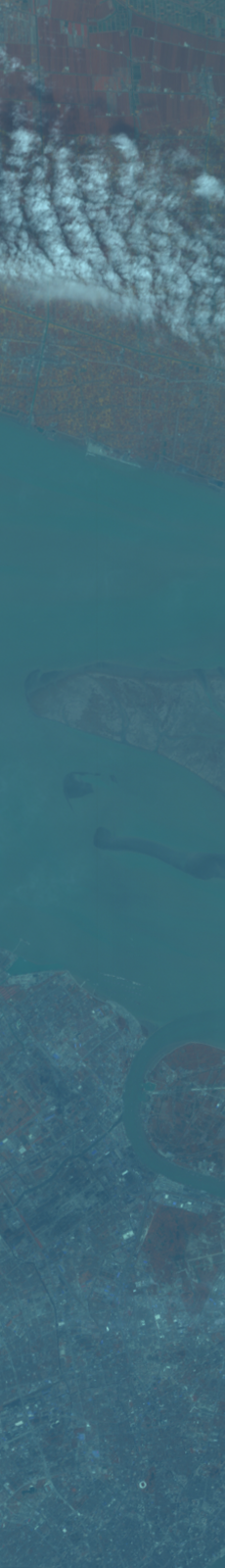}}\hfill
	\subfloat[GT of Shanghai]{\includegraphics[angle=90,width=1\linewidth]{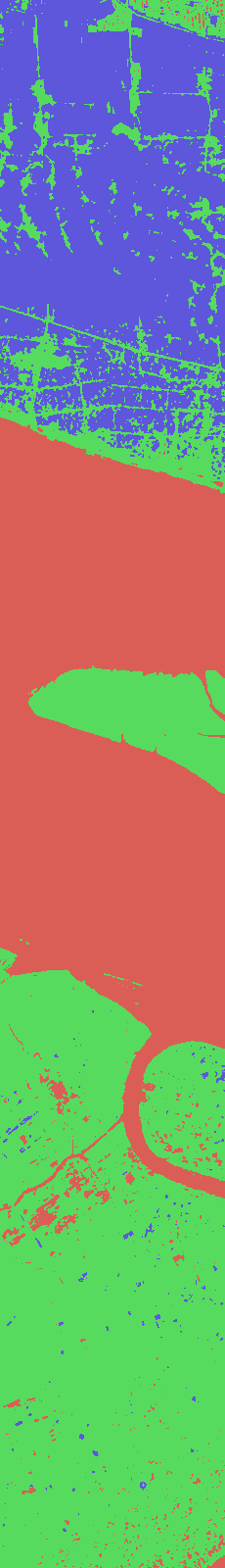}}
	\caption{(a) The source image Hangzhou. (b) The ground truth of Hangzhou. (c) The target image Shanghai. (d) The ground truth of Shanghai.}
	\label{fig:datacube_input}
\end{figure}

\subsubsection{Hanghzou--Shanghai} This pair of datasets, consisting of the source image, Hangzhou, and the target image, Shanghai, was acquired by the REO-1 Hyperion hyperspectral sensor~\cite{ye2017dictionary}. The number of spectral bands obtained by the sensor is 220, among which 198 spectral bands are kept after removing bad bands. The Hangzhou dataset, with a dimension of $590 \times 230$, was acquired on November 2, 2002, over the Hangzhou city. It covers the water areas of the West Lake and Qiantang River, buildings and plants. The Shanghai dataset, with a dimension of $1600 \times 230$, was acquired on April 1, 2002, over the Shanghai city, which is 170 km far away from Hangzhou. It covers buildings, plants and the water areas of the Yangtze River and Huangpu River. The detail of the datasets can be found in Table \ref{tab:datasetDataCube} and the images are shown in Fig.~\ref{fig:datacube_input}. Note that compared to the other two datasets, this dataset is very noisy, hence it can be used to evaluate the robustness of different classifiers.

\subsection{Experimental Setup}
The classification results of the proposed approach are compared with eight state-of-the-art approaches including 1D-CNN~\cite{SongYang2019LCCf} based on 1D convolution neural network (CNN);
Spec-Spat~\cite{li2017spectral}, 3D-CNN~\cite{hamida20183}, and SSRN~\cite{zhong2018spectral} based on 3D CNN;  HSI-CNN~\cite{luo2018hsi} and HybridSN~\cite{roy2019hybridsn} based on both 2D and 3D CNN; and the transfer learning framework HT-CNN~\cite{he2019heterogeneous} based on VGGNet~\cite{simonyan2014very} and the cross-temporal method proposed by Liu~\etal~\cite{liu2019unsupervised}. Note that since the code for Liu~\etal is not available, Note that since the code for Liu~\etal is not available, we directly list their result in overall classification accuracy on the shared Houston dataset at the end of Sec.~\ref{sec:classification}. 

All these methods are reported with the best performance in recent literature with available codes. To test the robustness of the methods, 1\%, 3\%, and 5\% of the samples in the source dataset are randomly selected as training samples. And the trained classifiers are further applied to predict the labels of both the source and target datasets. For quantitative comparison, the classification performance is evaluated with three widely-used metrics, \ie, overall accuracy (OA), average accuracy (AA), and kappa coefficient ($\kappa$)~\cite{paoletti2019deep}. OA is estimated by dividing the number of corrected predicted labels with the number of total labels for the entire image. AA is the average classification accuracy for all the classes. And $\kappa$ is a statistical coefficient, which measures the degree of accuracy and reliability for classification. The higher the metrics, the better the classification performance.  

We conduct three sets of experiments. In Sec.~\ref{sec:classification}, we thoroughly study the classification accuracy of the proposed scheme and compare to the seven state-of-the-art approaches. In Sec.~\ref{sec:evaluate_tf}, we provide visual inspection of same class sample distributions in both the raw image space and the abundance space to qualitatively show the effectiveness of the proposed scheme. In Sec.~\ref{sec:ablation}, we perform ablation study on the different components forming the proposed scheme.

\subsection{Classification Results}
\label{sec:classification}
We conduct two groups of experiments. The first group performs training and testing on the same dataset to set up baseline. The second group performs training on the source dataset and testing on the target dataset.

\subsubsection{Training and Testing on the Same Dataset}
\begin{figure*}
\centering
\subfloat[Average OA]{\includegraphics[width=.33\linewidth]{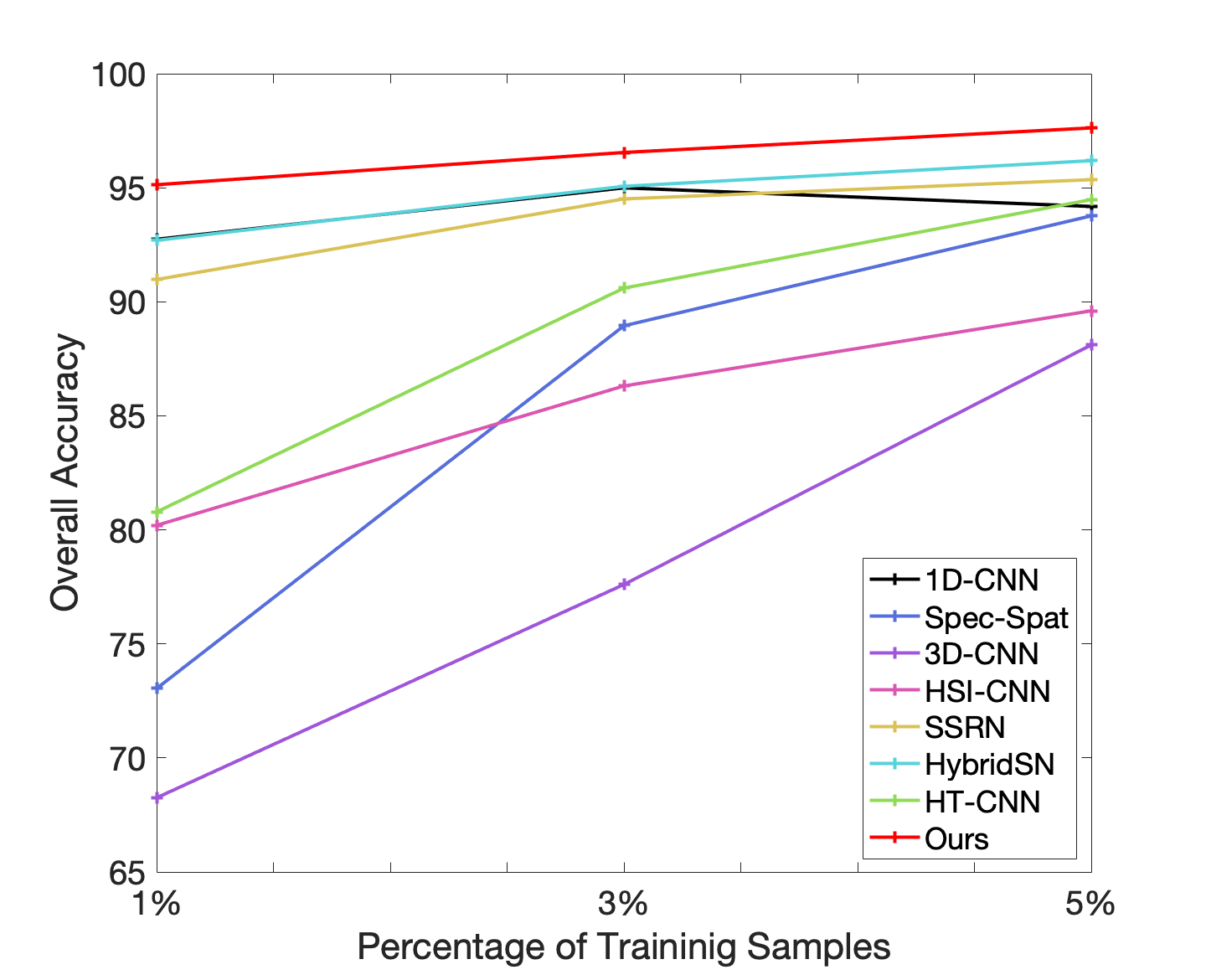}}
\subfloat[Average AA]{\includegraphics[width=.33\linewidth]{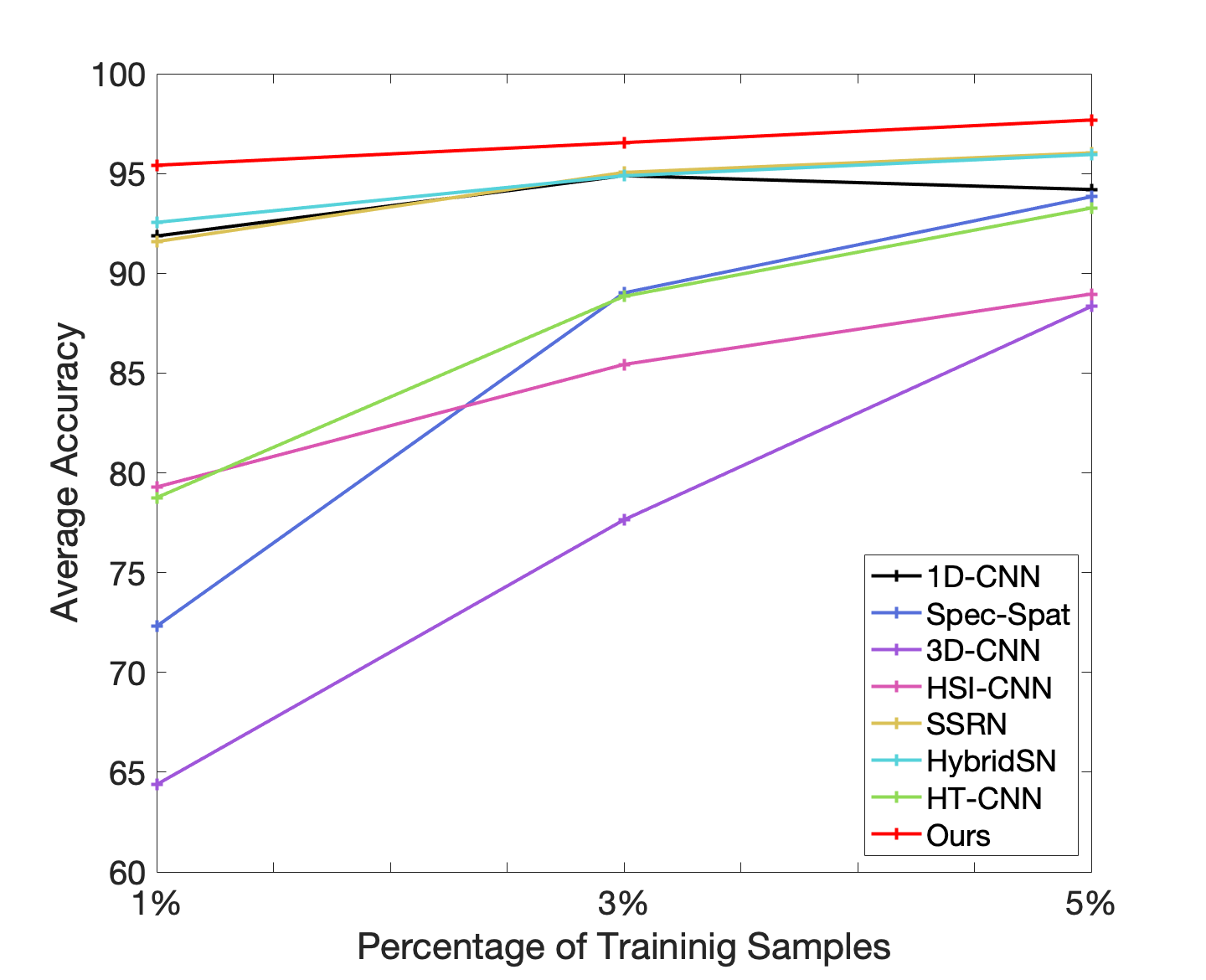}}
\subfloat[Average Kappa]{\includegraphics[width=.33\linewidth]{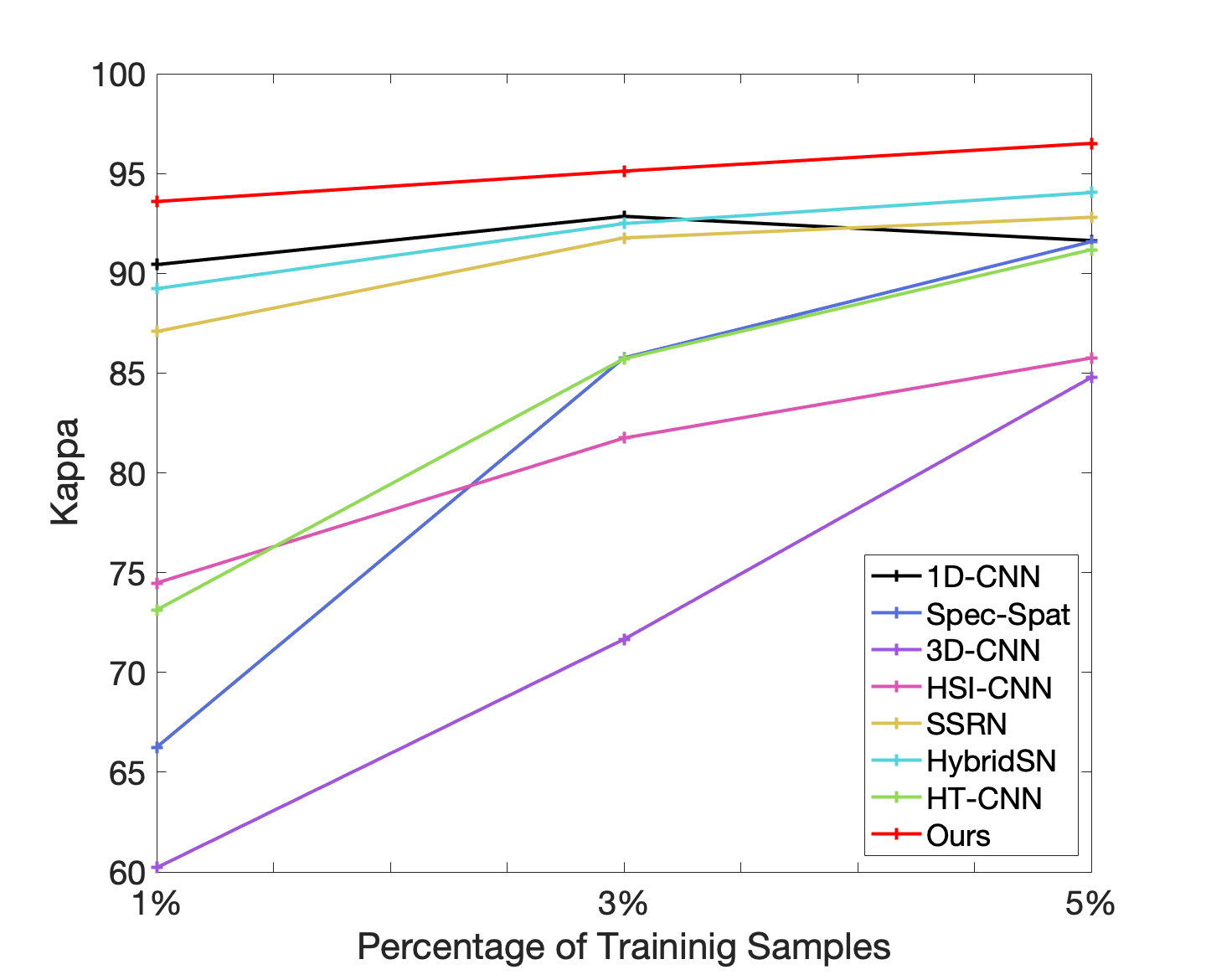}}
\caption{The average classification metrics over the three source datasets.}
\label{fig:source_res}
\end{figure*}

We first perform experiments on the general case where both the training and testing are conducted on the source dataset. The methods are trained with 1\%, 3\% and 5\% of the labeled samples from each source dataset. The classification results of different methods on the three datasets are reported in Tables~\ref{tab:paviaU},~\ref{tab:houston13} and~\ref{tab:datacube2}, respectively. Since the spectral characteristics are diverse in different datasets, the performance of the same method may work well for one dataset but fail on the other dataset. Thus, to demonstrate the trends of the methods in different scenarios, we average each metric over the three datasets and show the results in Fig.~\ref{fig:source_res}. 

We can observe that when the number of training samples increases, most methods could achieve a better classification accuracy. The Spec-Spat~\cite{li2017spectral}, HSI-CNN~\cite{luo2018hsi} and especially the 3D-CNN~\cite{hamida20183} methods are more sensitive to the number of training samples. They could achieve a relatively good accuracy when there are sufficient number of training samples, ~\eg, on the first dataset, Pavia University,  and the third dataset, Hangzhou, as shown in Tables~\ref{tab:paviaU} and~\ref{tab:datacube2}, respectively. However, their classification performance drops significantly on the second dataset, Houston-2013, which has limited number of training samples. %, even the training and testing are performed on the same dataset. 
We can observe that the overall accuracy of these methods are below 60\% when we train the network with  1\% of the labeled samples in the Houston dataset. The 1D-CNN~\cite{SongYang2019LCCf} performs well with limited number of training samples. However, since 1D-CNN depends only on the spectral information of the pixels, its performance can be unstable due to noise. For example, in the third dataset which is more noisy than the other two, when the number of training samples increases, the classification performance actually drops. The SSRN~\cite{zhong2018spectral} and HybridSN~\cite{roy2019hybridsn} work well for the first two datasets as shown in Tables~\ref{tab:paviaU} and~\ref{tab:houston13}. However, they are not robust enough to noisy data. As shown from the results of the noisy data in Table~\ref{tab:datacube2}, the performance of these two methods drops even the third dataset has more number of training samples than that of the first two datasets. 

The transfer learning based method HT-CNN~\cite{he2019heterogeneous} achieves better accuracy than that of Spec-Spat, HSI-CNN and 3D-CNN. That is because the network is designed based on VGGNet which carries the image priors. However, the limited number of training samples limited the classification performance of the HT-CNN due to its large number of network weights. The proposed PCTL-SAS achieves better or comparable accuracy to the other methods. As shown in Fig.~\ref{fig:source_res}, the average OA, AA and Kappa of the proposed method outperform the other methods to a large margin because of its strong representative and discriminative power. Note that, since the networks are trained to extract representations from both the source and the target domains, it sacrifices the accuracy on the source image a little bit. If we only need to classify the images in the source domain, we could remove the reconstruction branch for the target domain. In this way, the classification accuracy would be further increased. 
%TODO: run datacube2 of my method to increase the accuracy. 
%1D-CNN~\cite{SongYang2019LCCf}, Spec-Spat~\cite{li2017spectral}, 3D-CNN~\cite{hamida20183}, and SSRN~\cite{zhong2018spectral}, HSI-CNN~\cite{luo2018hsi}and HybridSN~\cite{roy2019hybridsn}. HT-CNN~\cite{he2019heterogeneous}

\begin{table*}[htbp]
  \centering
  \caption{Performance comparison when the models are trained and tested on the same dataset, Pavia University.}
    \begin{tabular}{c|c|cccccccc}
    \hline
    & Method & 1D-CNN & Spec-Spat & 3D-CNN & HSI-CNN & SSRN  & HybridSN & HT-CNN & Ours \\
    \hline
    5\%   & OA    & 98.43$\pm$0.42 & 98.98$\pm$0.26 & 97.3$\pm$1.57 & 97.64$\pm$0.23 & 99.93$\pm$0.05 & 99.68$\pm$0.15 & 99.61$\pm$0.19 & 99.83$\pm$0.06 \\
          & AA    & 97.92$\pm$0.88 & 98.86$\pm$0.25 & 97.01$\pm$1.54 & 96.94$\pm$0.36 & 99.95$\pm$0.03 & 99.65$\pm$0.18 & 99.60$\pm$0.18 & 99.85$\pm$0.06 \\
          & kappa & 97.78$\pm$0.59 & 98.56$\pm$0.37 & 96.19$\pm$2.22 & 96.67$\pm$0.32 & 99.90$\pm$0.07 & 99.55$\pm$0.22 & 99.44$\pm$0.26 & 99.76$\pm$0.09 \\
    \hline
    3\%   & OA    & 97.94$\pm$0.13 & 98.06$\pm$0.25 & 87.59$\pm$10.28 & 95.07$\pm$3.1 & 99.71$\pm$0.32 & 99.55$\pm$0.15 & 99.33$\pm$0.35 & 99.61$\pm$0.10\\
          & AA    & 96.66$\pm$0.65 & 97.76$\pm$0.81 & 88.17$\pm$8.70 & 94.13$\pm$2.25 & 99.72$\pm$0.27 & 99.51$\pm$0.11 & 99.30$\pm$0.38 & 99.66$\pm$0.06\\
          & kappa & 97.09$\pm$0.18 & 97.26$\pm$0.35 & 83.01$\pm$13.85 & 93.11$\pm$4.22 & 99.59$\pm$0.45 & 99.37$\pm$0.21 & 99.05$\pm$0.49 & 99.44$\pm$0.15\\
    \hline
    1\%   & OA    & 97.07$\pm$0.75 & 95.61$\pm$0.87 & 88.5$\pm$2.13 & 93.2$\pm$0.96 & 99.77$\pm$0.1 & 99.38$\pm$0.15 & 97.94$\pm$0.50 & 99.57$\pm$0.08 \\
          & AA    & 95.22$\pm$1.71 & 94.86$\pm$0.91 & 78.69$\pm$4.40 & 91.07$\pm$0.94 & 99.77$\pm$0.12 & 99.24$\pm$0.17 & 97.59$\pm$0.69 & 99.55$\pm$0.09 \\
          & kappa & 95.06$\pm$1.07 & 93.80$\pm$1.24 & 83.50$\pm$3.11 & 90.39$\pm$1.36 & 99.68$\pm$0.15 & 99.13$\pm$0.21 & 97.08$\pm$0.70 & 99.39$\pm$0.12 \\
    \hline
    \end{tabular}%
  \label{tab:paviaU}%
\end{table*}%

\begin{table*}[htbp]
  \centering
  \caption{Performance comparison when the models are trained and tested on the same dataset, Houston 2013.}
    \begin{tabular}{c|c|cccccccc}
    \hline
          & Method & 1D-CNN & Spec-Spat & 3D-CNN & HSI-CNN & SSRN  & HybridSN & HT-CNN & Ours \\
    \hline
    5\%   & OA    & 91.98$\pm$3.4 & 88.24$\pm$3.84 & 73.28$\pm$4.58 & 81.41$\pm$2.18 & 94.62$\pm$0.72 & 96.26$\pm$1.25 & 94.81$\pm$1.22 & 96.58$\pm$0.29 \\
          & AA    & 90.87$\pm$4.14 & 88.67$\pm$3.44 & 74.29$\pm$4.69 & 82.21$\pm$2.21 & 95.36$\pm$0.62 & 96.41$\pm$1.54 & 92.58$\pm$1.64 & 96.66$\pm$0.31 \\
          & kappa & 90.61$\pm$3.99 & 86.28$\pm$4.47 & 68.81$\pm$5.33 & 78.33$\pm$2.55 & 93.70$\pm$0.84 & 95.63$\pm$1.47 & 93.86$\pm$1.43 & 96.01$\pm$0.34 \\
   	\hline
    3\%   & OA    & 93.2$\pm$1.22 & 75.27$\pm$24.37 & 52.31$\pm$19.57 & 73.83$\pm$2.92 & 92.71$\pm$1.41 & 94.08$\pm$1.3 & 88.02$\pm$2.74 & 94.19$\pm$0.89 \\
          & AA    & 92.76$\pm$1.27 & 76.04$\pm$24.17 & 52.30$\pm$20.34 & 74.15$\pm$3.06 & 93.58$\pm$1.52 & 94.65$\pm$1.12 & 83.85$\pm$3.23 & 94.77$\pm$0.63 \\
          & kappa & 92.05$\pm$1.43 & 71.13$\pm$28$\pm$50 & 44.10$\pm$23.12 & 69.41$\pm$3.41 & 91.48$\pm$1.65 & 93.08$\pm$1.52 & 85.79$\pm$3.25 & 93.21$\pm$1.04 \\
    \hline
    1\%   & OA    & 84.06$\pm$8.49 & 31.74$\pm$23.02 & 25.6$\pm$8.39 & 57.37$\pm$6.01 & 85.04$\pm$2.94 & 89.48$\pm$3.92 & 62.52$\pm$5.89 & 89.18$\pm$2.03 \\
          & AA    & 82.57$\pm$8.93 & 30.43$\pm$24.13 & 24.45$\pm$6.11 & 57.81$\pm$6.32 & 86.07$\pm$3.63 & 90.00$\pm$3.98 & 59.6$\pm$5.75 & 89.78$\pm$1.48 \\
          & kappa & 81.32$\pm$9.93 & 18.94$\pm$27.73 & 13.14$\pm$7.63 & 50.14$\pm$7.04 & 82.49$\pm$3.42 & 87.68$\pm$4.60 & 55.58$\pm$6.93 & 87.32$\pm$2.39 \\
    \hline
    \end{tabular}%
  \label{tab:houston13}%
\end{table*}%

\begin{table*}[htbp]
  \centering
  \caption{Performance comparison when the models are trained and tested on the same dataset, Hangzhou.}
    \begin{tabular}{c|c|cccccccc}
    \hline
          & Method & 1D-CNN & Spec-Spat & 3D-CNN & HSI-CNN & SSRN  & HybridSN & HT-CNN & Ours \\
    \hline
    5\%   & OA    & 92.13$\pm$0.64 & 94.08$\pm$0.35 & 93.74$\pm$0.31 & 89.76$\pm$0.52 & 91.51$\pm$0.32 & 92.62$\pm$0.26 & 89.02$\pm$0.77 & 96.46$\pm$0.36 \\
          & AA    & 93.80$\pm$0.48 & 93.97$\pm$0.10 & 93.71$\pm$0.49 & 87.73$\pm$0.54 & 92.79$\pm$0.83 & 91.79$\pm$0.20 & 87.63$\pm$1.46 & 96.53$\pm$0.67 \\
          & kappa & 86.53$\pm$1.07 & 89.91$\pm$0.55 & 89.34$\pm$0.51 & 82.26$\pm$0.91 & 84.82$\pm$0.64 & 86.95$\pm$0.43 & 80.25$\pm$1.52 & 93.76$\pm$0.64 \\
    \hline
    3\%   & OA    & 93.85$\pm$0.75 & 93.51$\pm$0.25 & 92.91$\pm$0.18 & 90.03$\pm$0.83 & 91.11$\pm$0.44 & 91.56$\pm$0.3 & 84.45$\pm$5.33 & 95.84$\pm$0.53 \\
          & AA    & 95.25$\pm$0.58 & 93.25$\pm$0.24 & 92.46$\pm$0.48 & 88.01$\pm$1.09 & 91.86$\pm$1.02 & 90.49$\pm$0.55 & 83.38$\pm$4.54 & 95.21$\pm$1.66 \\
          & kappa & 89.42$\pm$1.26 & 88.90$\pm$0.43 & 87.86$\pm$0.34 & 82.71$\pm$1.67 & 84.25$\pm$0.71 & 85.02$\pm$0.57 & 72.31$\pm$9.02 & 92.71$\pm$0.90 \\
    \hline
    1\%   & OA    & 97.09$\pm$0.8 & 91.78$\pm$1.47 & 90.66$\pm$1.12 & 90.02$\pm$0.69 & 88.12$\pm$1.22 & 89.22$\pm$0.38 & 81.90$\pm$0.65 & 96.65$\pm$0.31 \\
          & AA    & 97.81$\pm$0.80 & 91.64$\pm$0.35 & 90.00$\pm$0.74 & 88.98$\pm$0.25 & 88.92$\pm$1.74 & 88.41$\pm$0.59 & 79.08$\pm$0.96 & 96.91$\pm$ 0.30 \\
          & kappa & 94.92$\pm$1.39 & 86.00$\pm$2.26 & 83.99$\pm$1.63 & 82.91$\pm$0.98 & 79.07$\pm$1.69 & 80.88$\pm$0.76 & 66.73$\pm$1.28 & 94.08$\pm$0.57 \\
    \hline
    \end{tabular}%
  \label{tab:datacube2}%
\end{table*}%

\subsubsection{Training on the Source Dataset and Testing on the Target Dataset}
\begin{figure*}
\centering
\subfloat[Average OA]{\includegraphics[width=.33\linewidth]{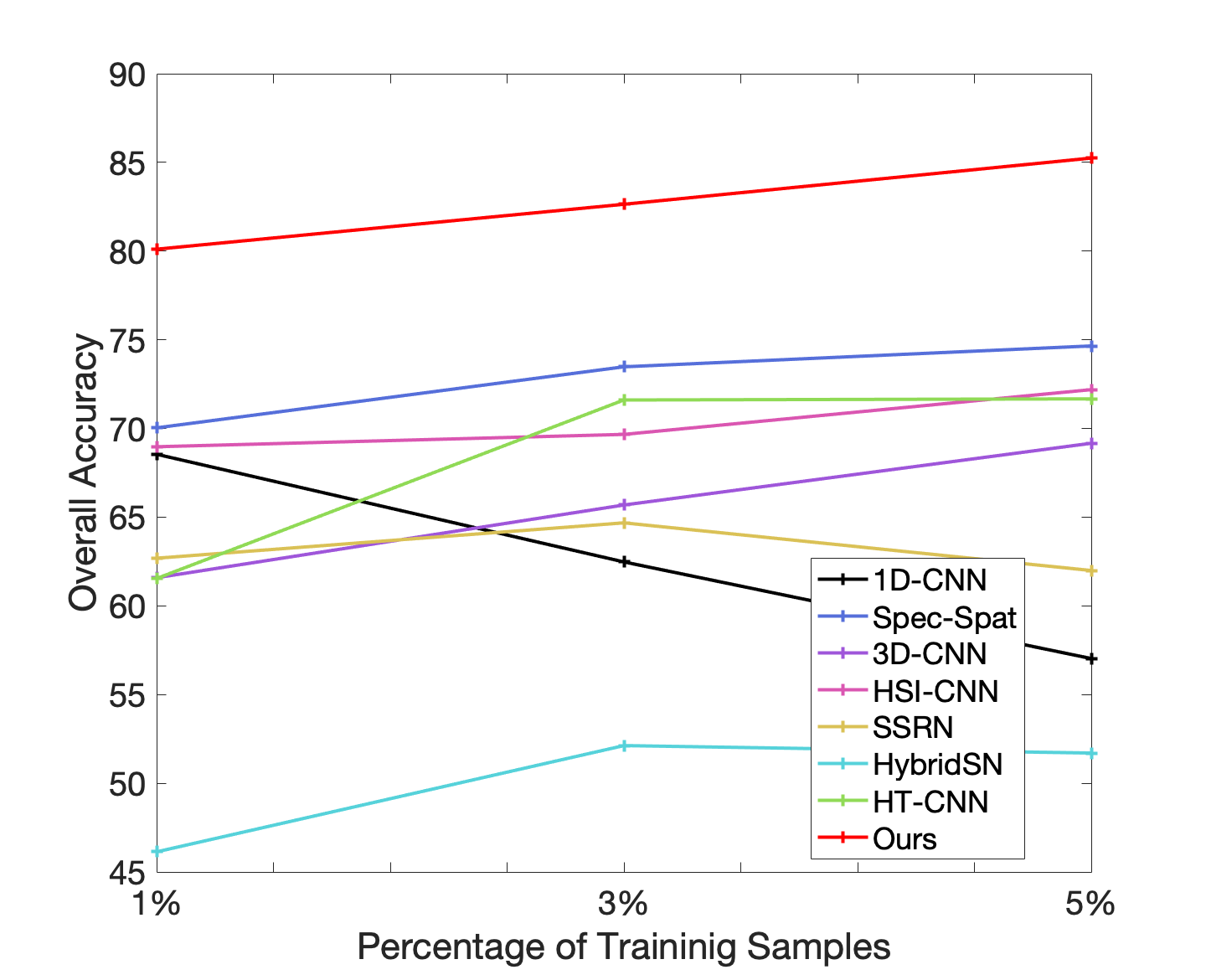}}
\subfloat[Average AA]{\includegraphics[width=.33\linewidth]{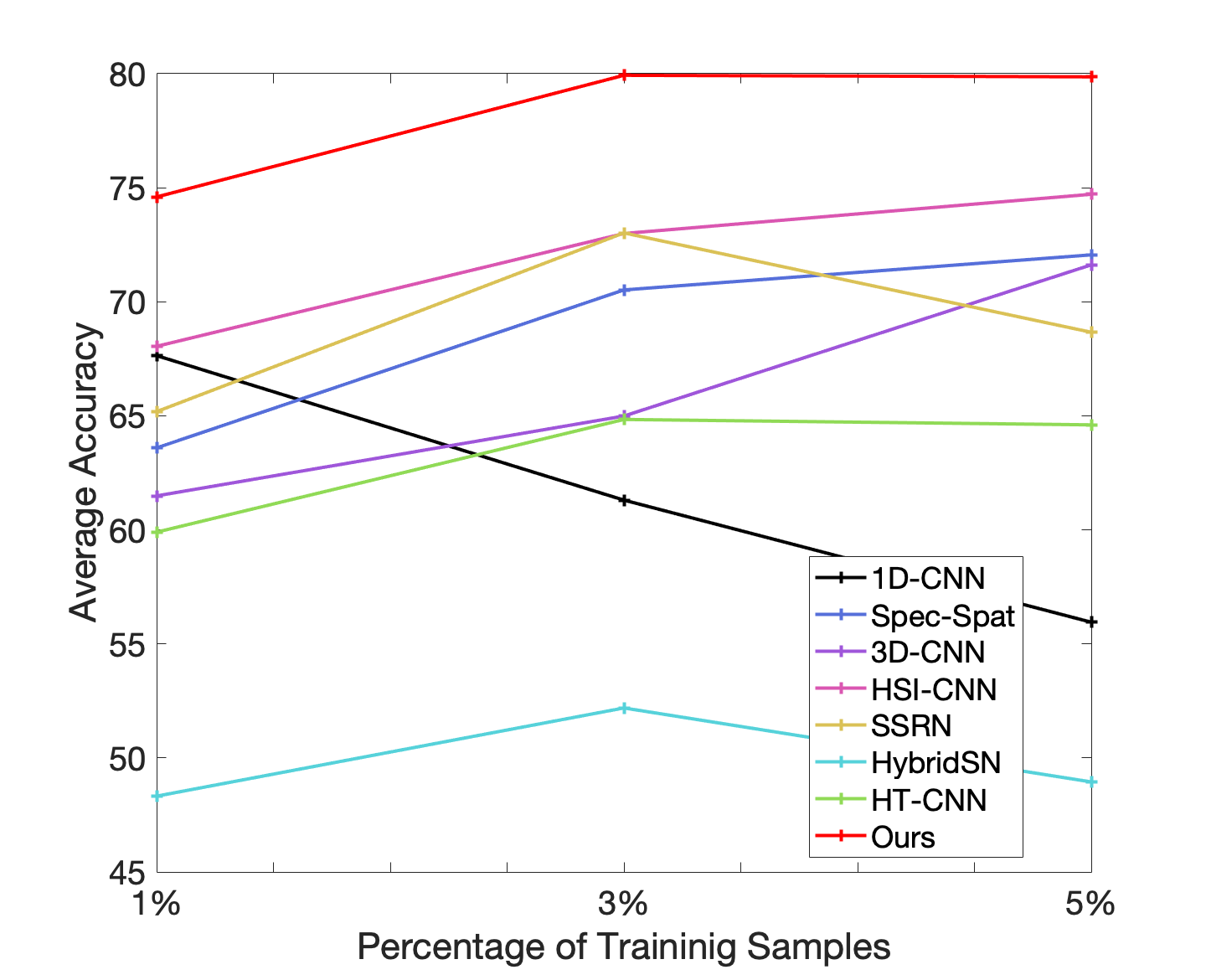}}
\subfloat[Average Kappa]{\includegraphics[width=.33\linewidth]{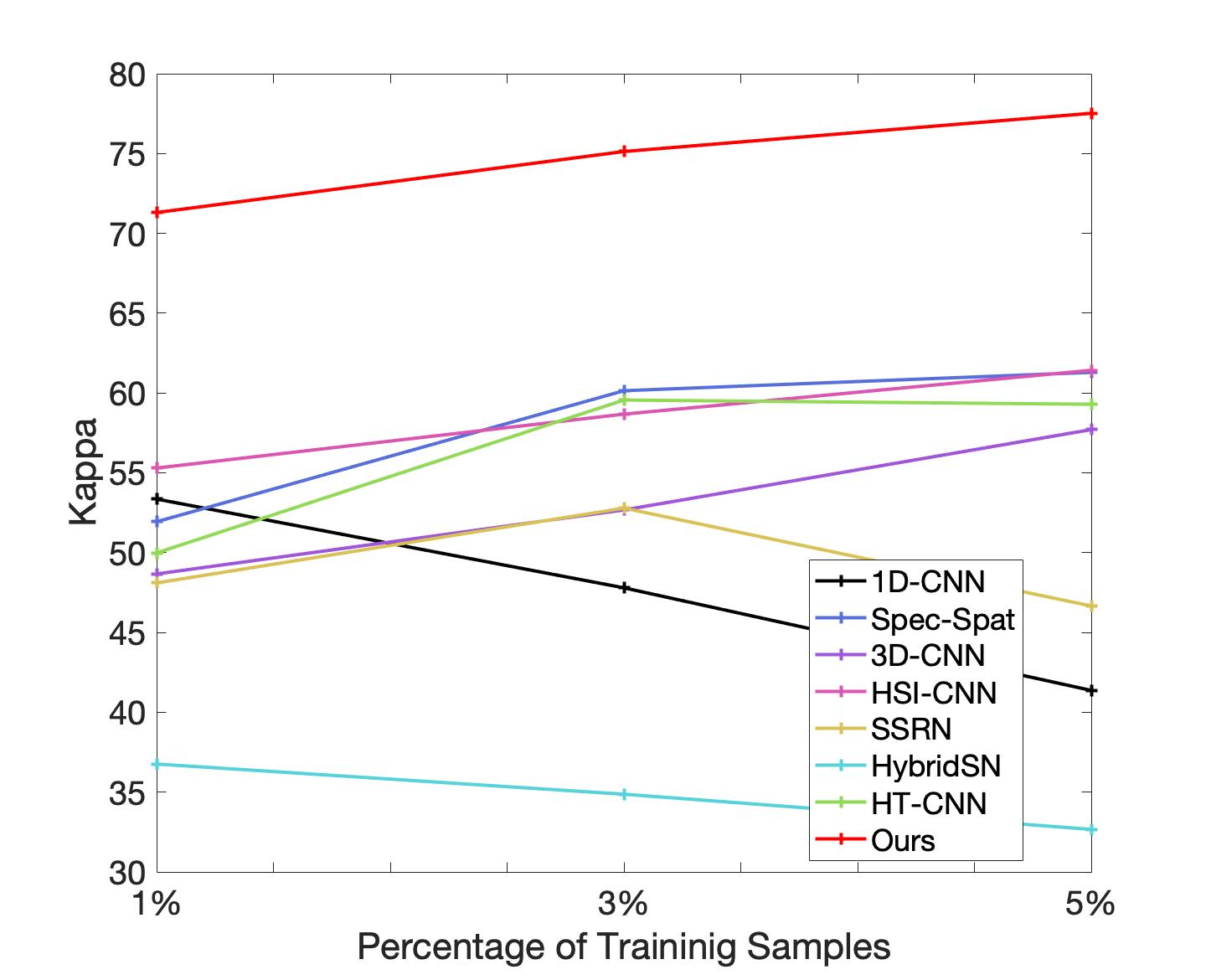}}
\caption{The average classification metric over the three target datasets.}
\label{fig:target_res}
\end{figure*}

\begin{figure*}
	\centering
	\subfloat[]{\includegraphics[width=0.18\linewidth]{fig/pavia/pavia_gt}}\hfill
	\subfloat[]{\includegraphics[width=0.18\linewidth]{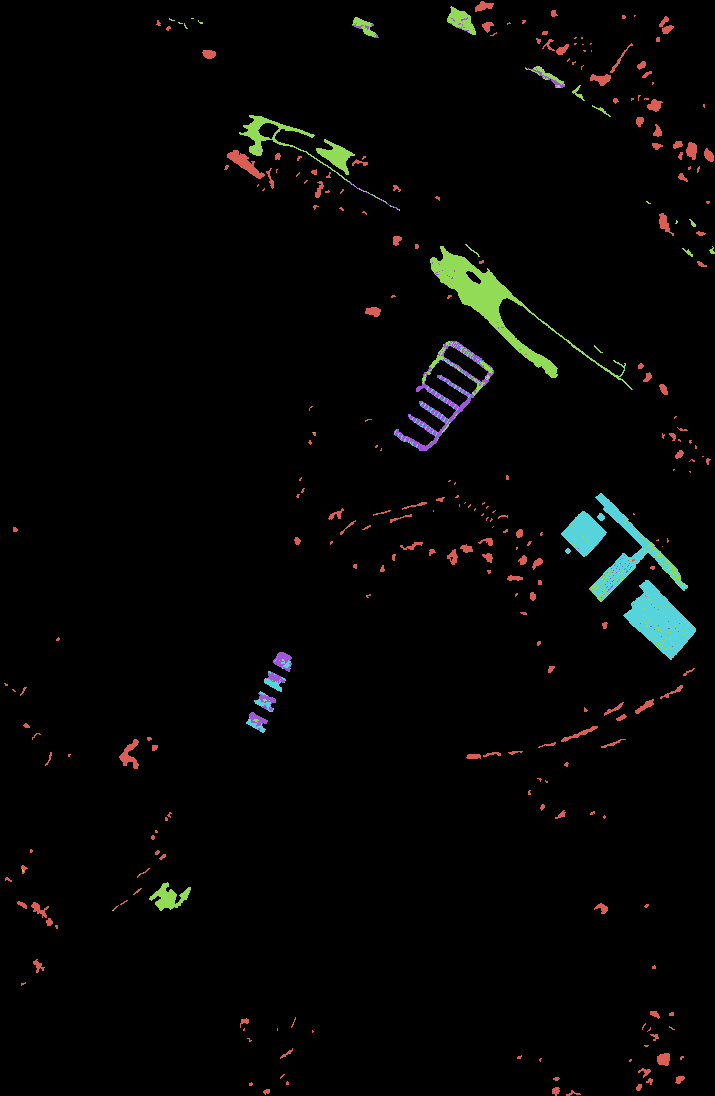}}\hfill
	\subfloat[]{\includegraphics[width=0.18\linewidth]{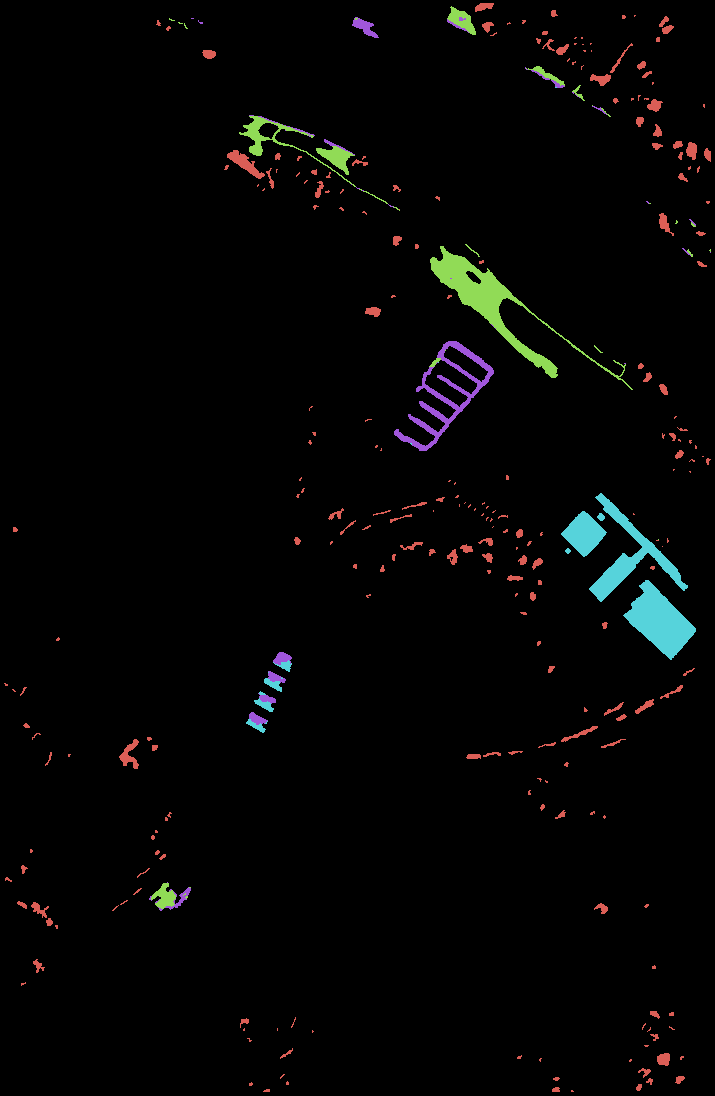}}\hfill
	\subfloat[]{\includegraphics[width=0.18\linewidth]{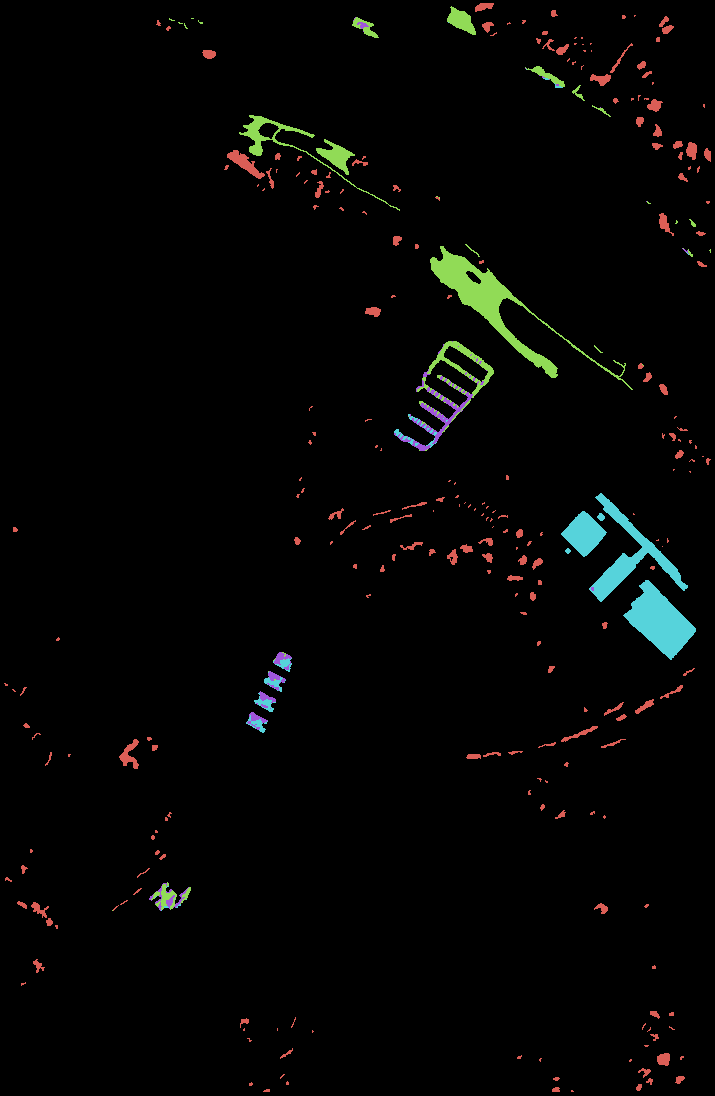}}\hfill
	\subfloat[]{\includegraphics[width=0.18\linewidth]{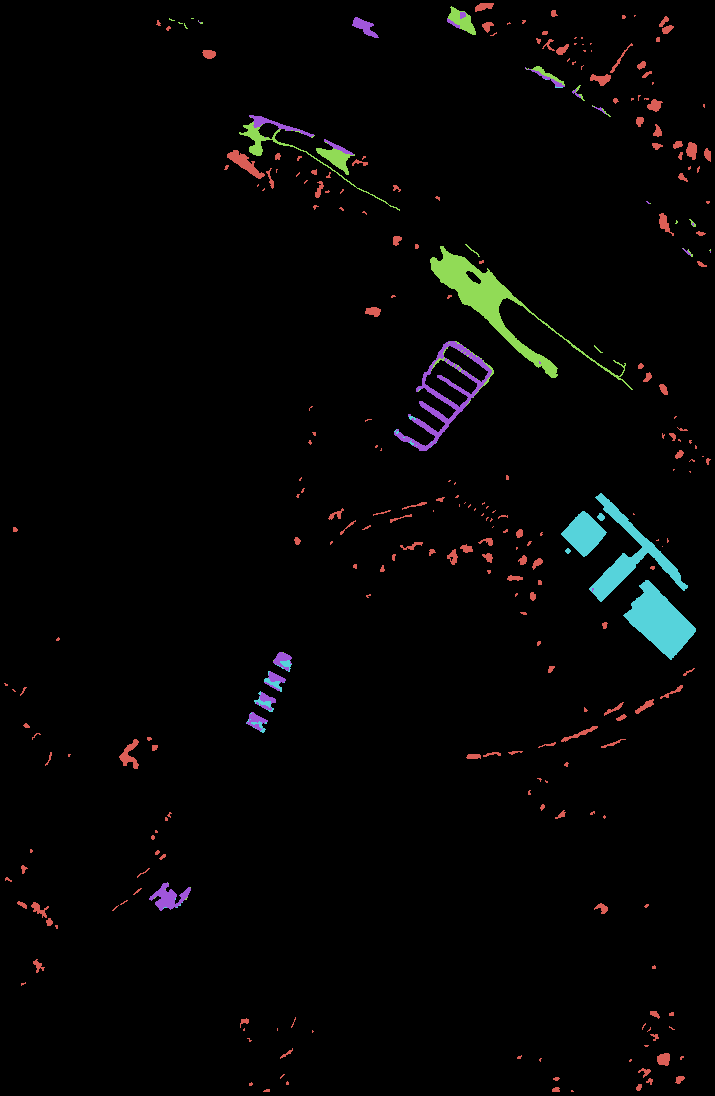}}\\
	\subfloat[]{\includegraphics[width=0.18\linewidth]{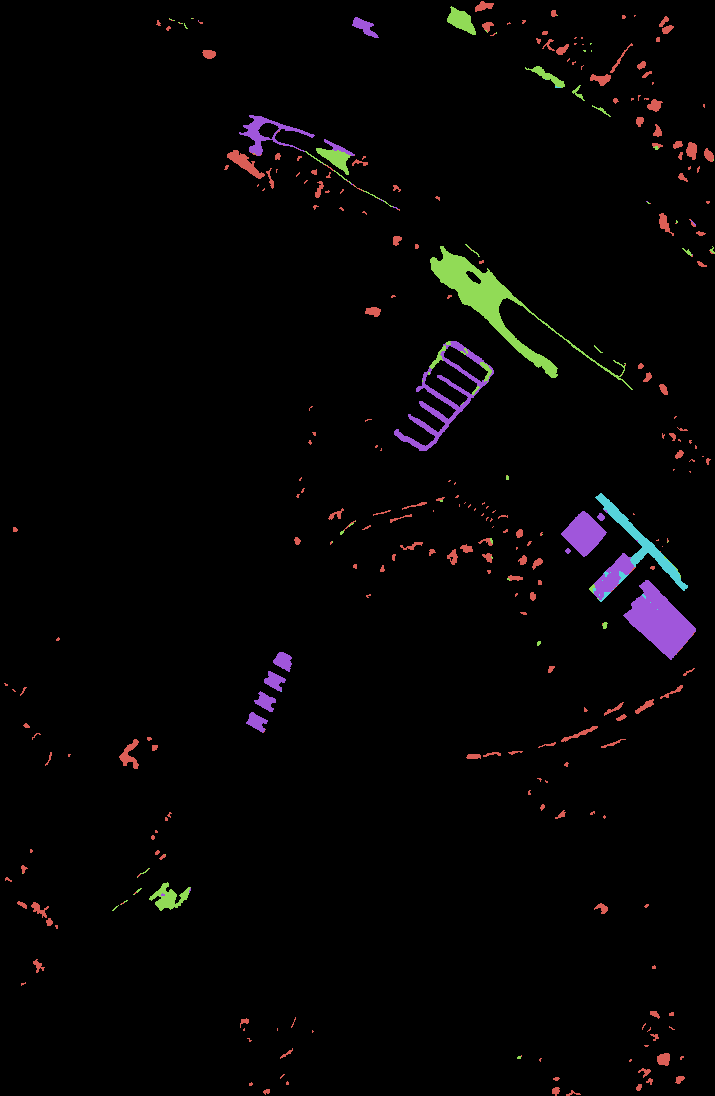}}\hfill
	\subfloat[]{\includegraphics[width=0.18\linewidth]{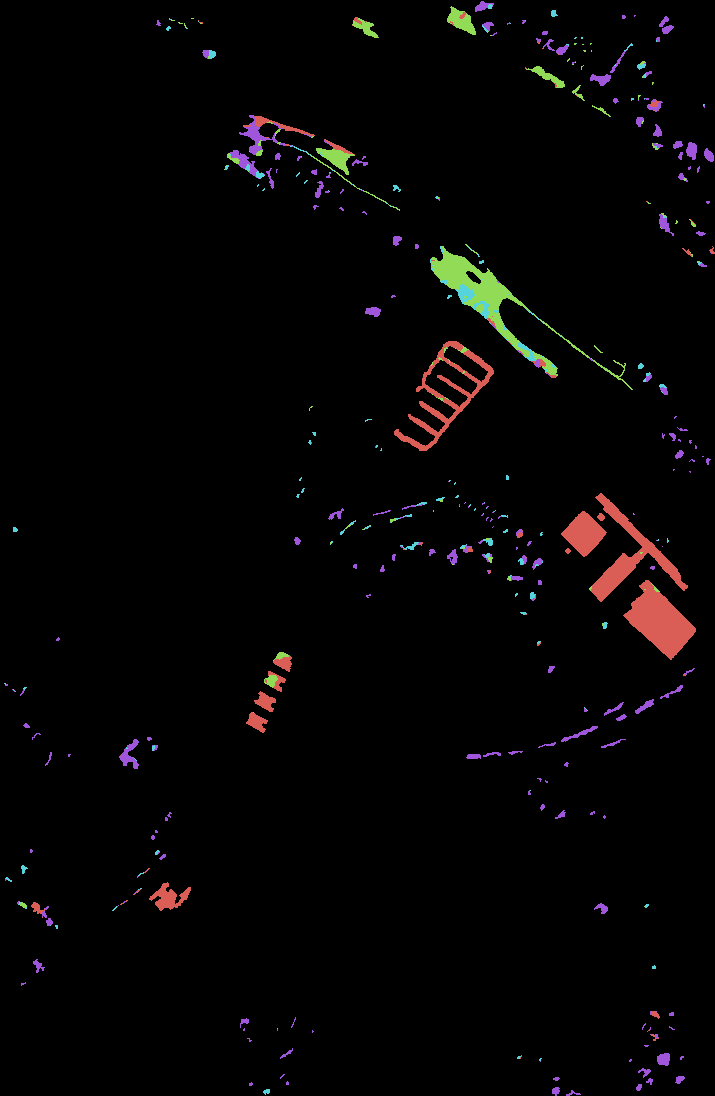}}\hfill
	\subfloat[]{\includegraphics[width=0.18\linewidth]{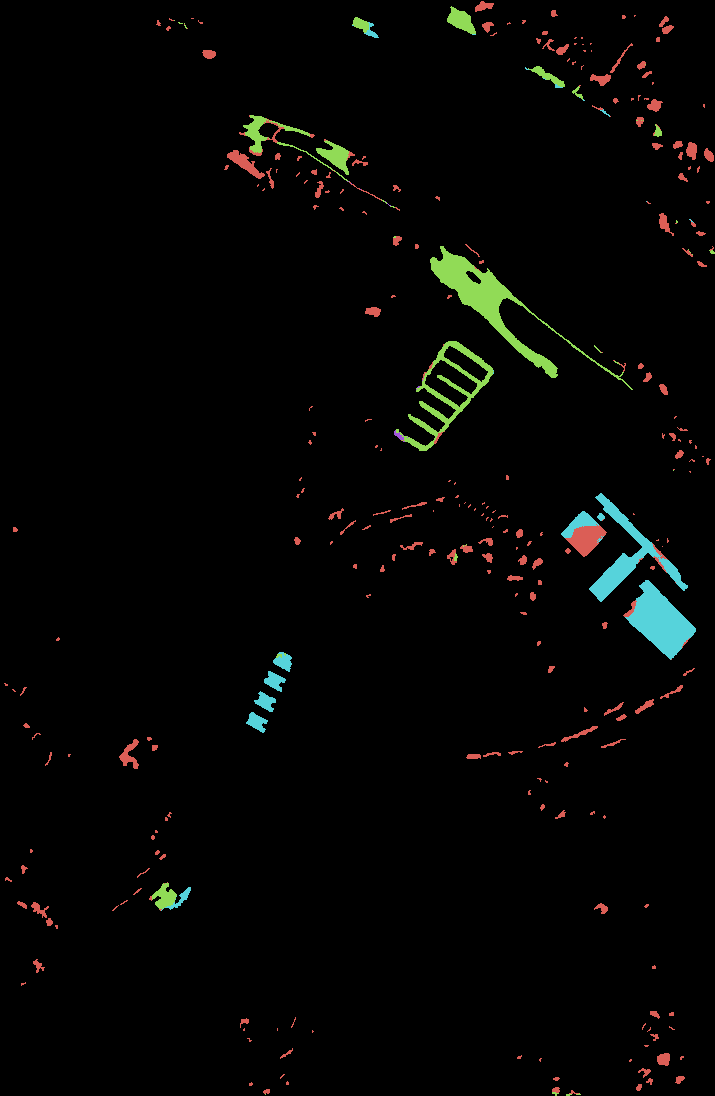}}\hfill
	\subfloat[]{\includegraphics[width=0.18\linewidth]{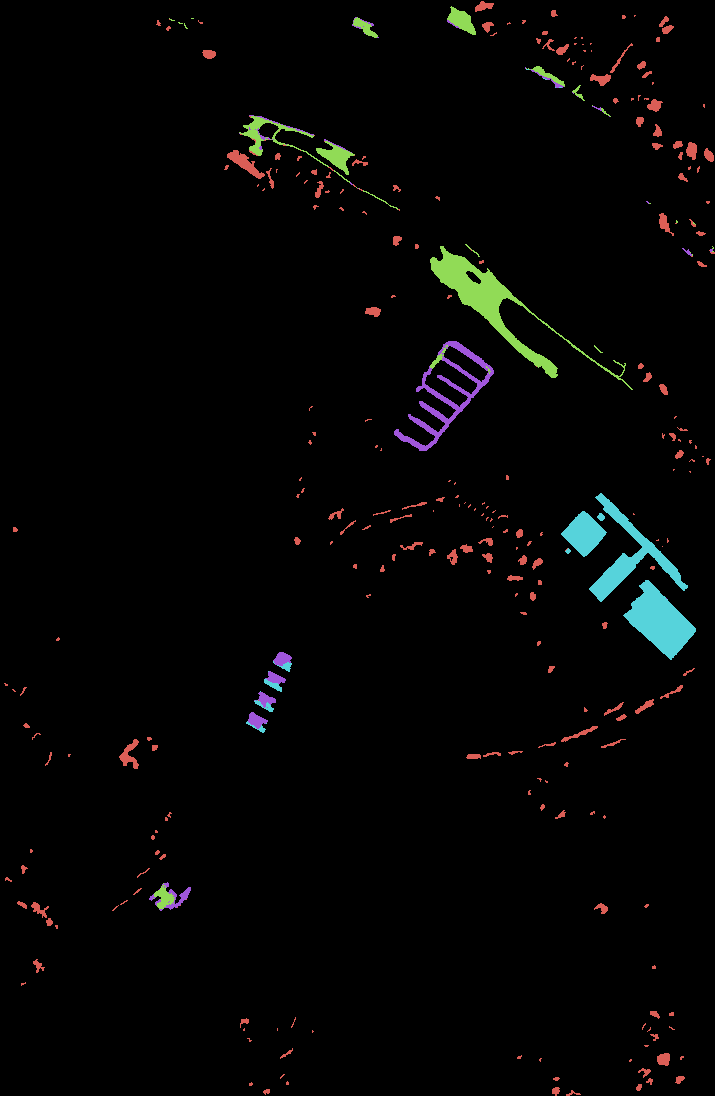}}
	\caption{(a) The ground truth classification map of the target image, Pavia Center. The predicted  map from (b) 1D-CNN, (c) Spec-Spat, (d) 3D-CNN, (e) HSI-CNN, (f) SSRN, (g) HybirdSN, (h) HT-CNN, (i) Ours.}
	\label{fig:pavia_output}
\end{figure*}

\begin{figure*}
	\centering
	\subfloat[]{\includegraphics[angle=90,width=0.11\linewidth]{fig/houston/houston_gt}}\hfill
	\subfloat[]{\includegraphics[angle=90,width=0.11\linewidth]{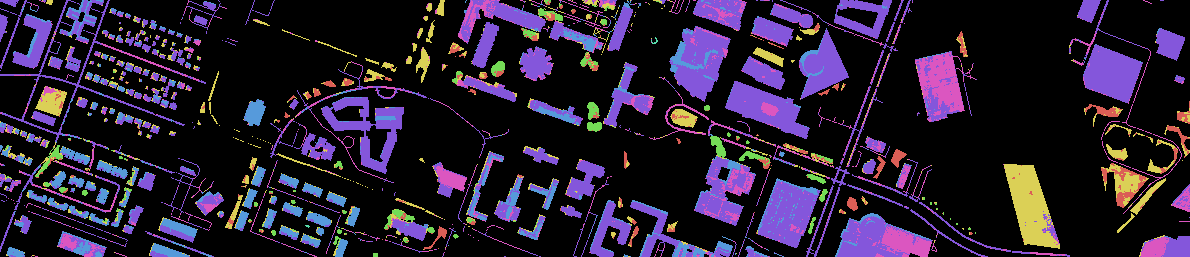}}\hfill
	\subfloat[]{\includegraphics[angle=90,width=0.11\linewidth]{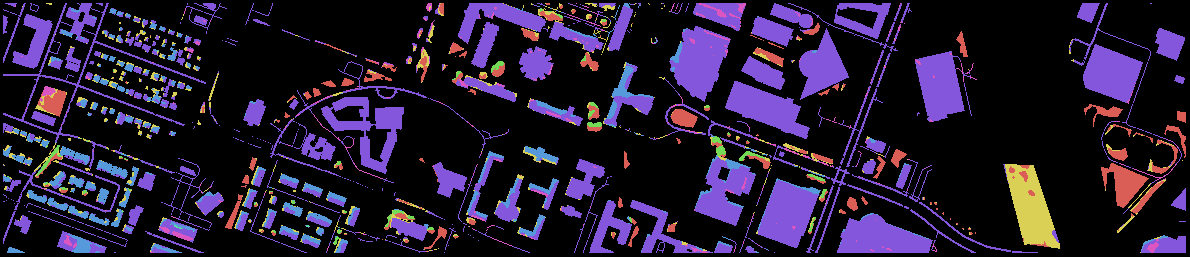}}\hfill
	\subfloat[]{\includegraphics[angle=90,width=0.11\linewidth]{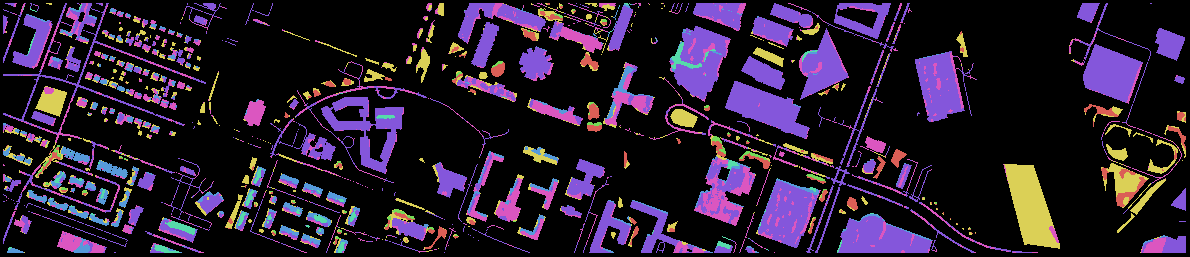}}\hfill
	\subfloat[]{\includegraphics[angle=90,width=0.11\linewidth]{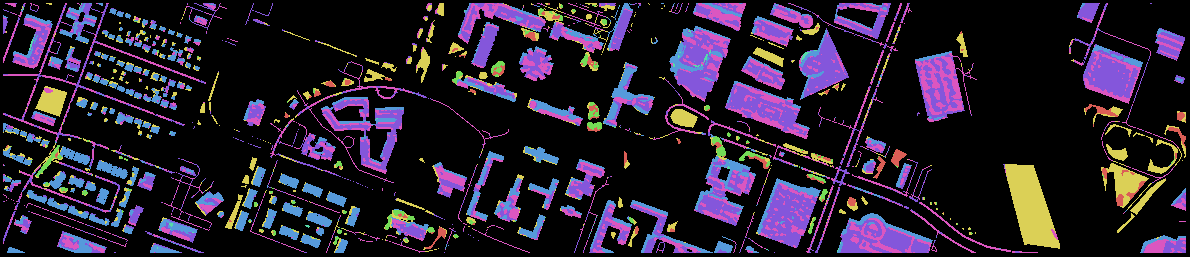}}\hfill
	\subfloat[]{\includegraphics[angle=90,width=0.11\linewidth]{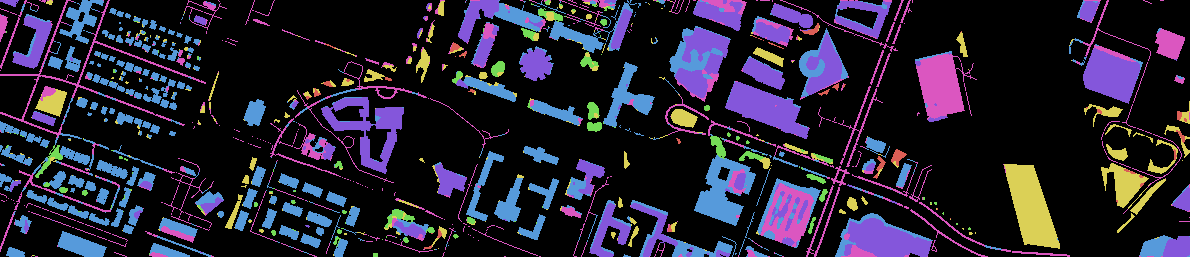}}\hfill
	\subfloat[]{\includegraphics[angle=90,width=0.11\linewidth]{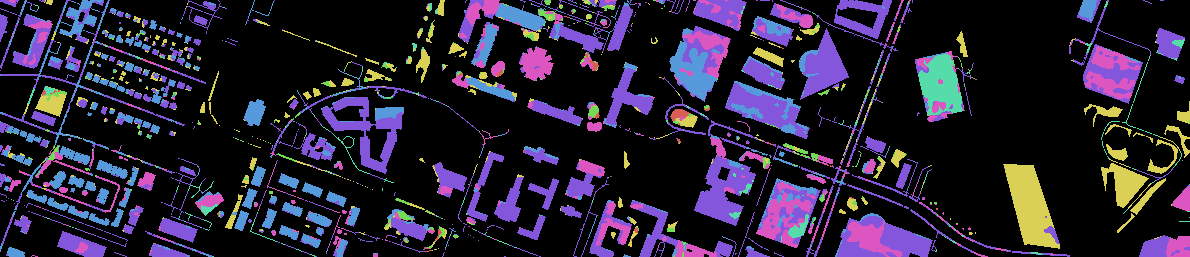}}\hfill
	\subfloat[]{\includegraphics[angle=90,width=0.11\linewidth]{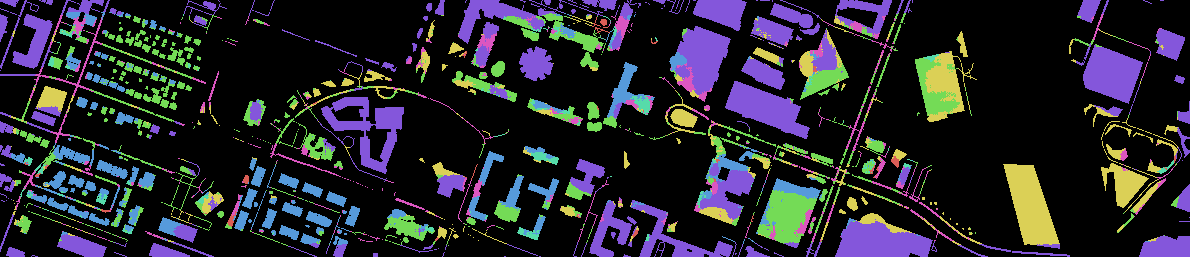}}\hfill
	\subfloat[]{\includegraphics[angle=90,width=0.11\linewidth]{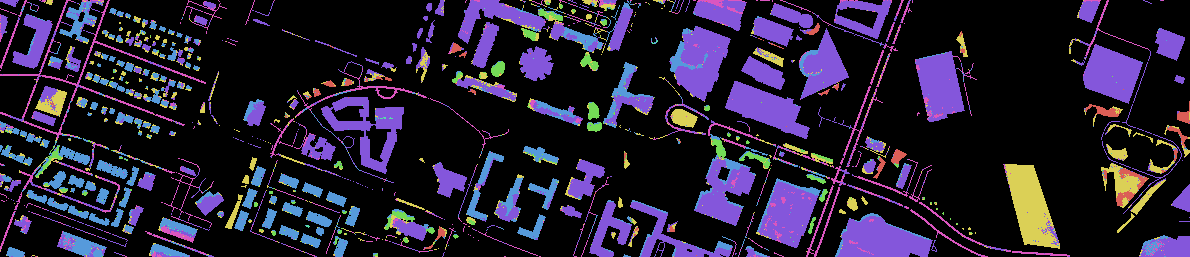}}
	\caption{(a) The ground truth of the target image, houston 2018. The predicted classification map from (b) 1D-CNN, (c) Spec-Spat, (d) 3D-CNN, (e) HSI-CNN, (f) SSRN, (g) HybirdSN, (h) HT-CNN, (i) Ours.}
	\label{fig:houston_output}
\end{figure*}

\begin{figure*}
	\centering
	\subfloat[]{\includegraphics[width=0.11\linewidth]{fig/datacube/datacube_gt}}\hfill
	\subfloat[]{\includegraphics[width=0.11\linewidth]{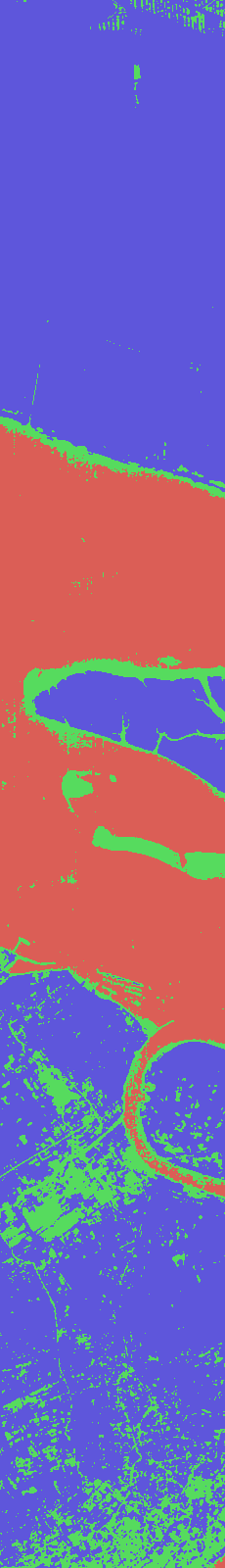}}\hfill
	\subfloat[]{\includegraphics[width=0.11\linewidth]{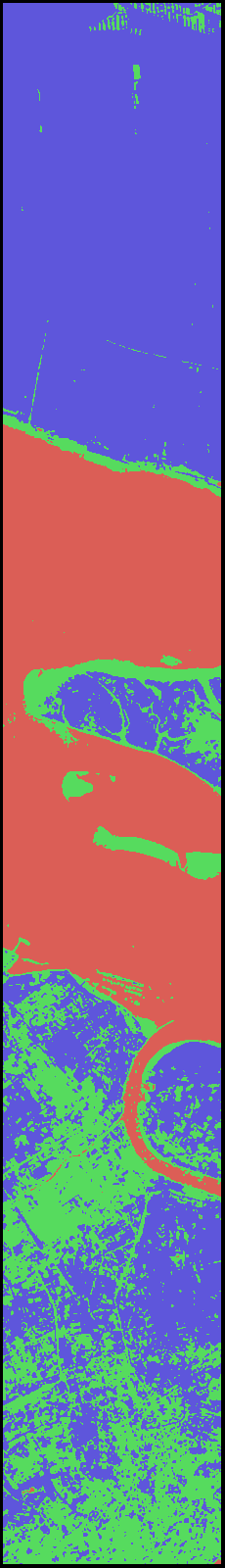}}\hfill
	\subfloat[]{\includegraphics[width=0.11\linewidth]{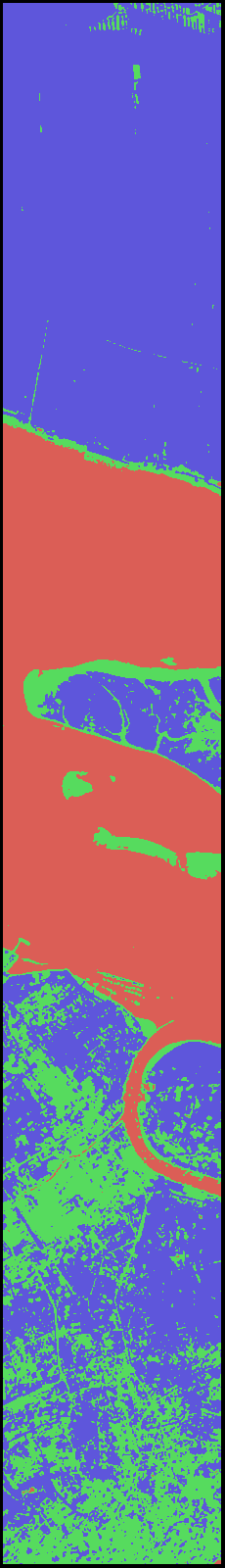}}\hfill
	\subfloat[]{\includegraphics[width=0.11\linewidth]{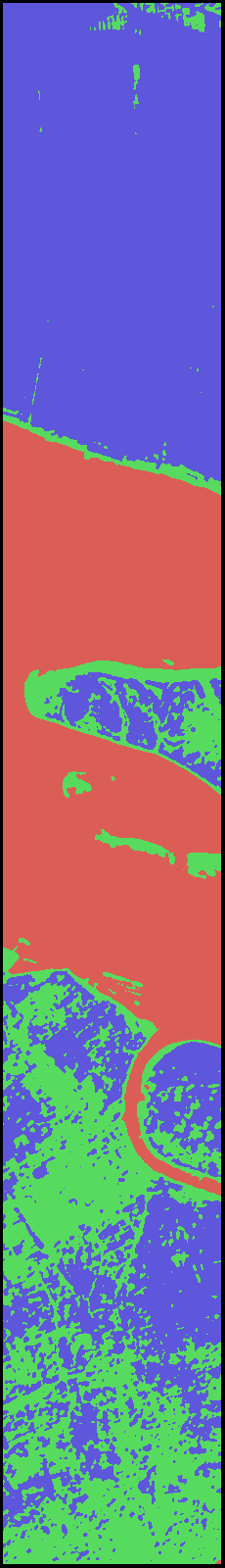}}\hfill
	\subfloat[]{\includegraphics[width=0.11\linewidth]{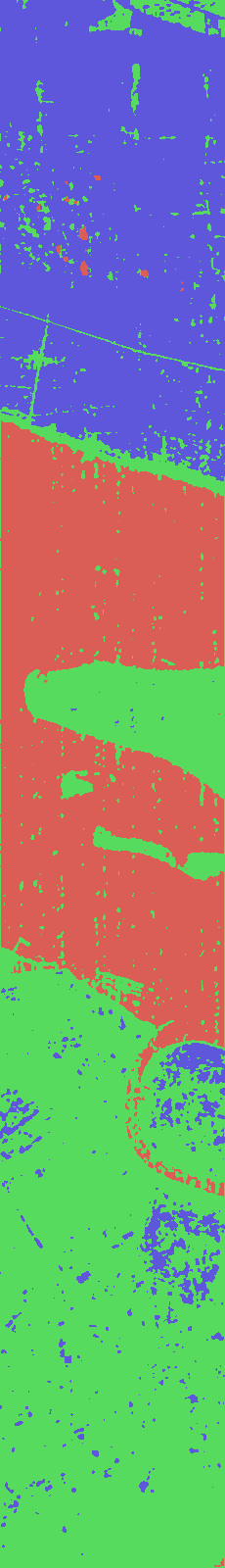}}\hfill
	\subfloat[]{\includegraphics[width=0.11\linewidth]{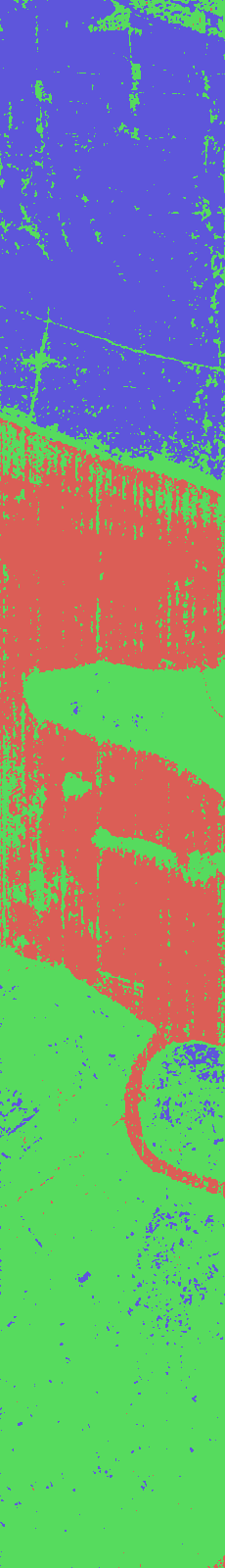}}\hfill
	\subfloat[]{\includegraphics[width=0.11\linewidth]{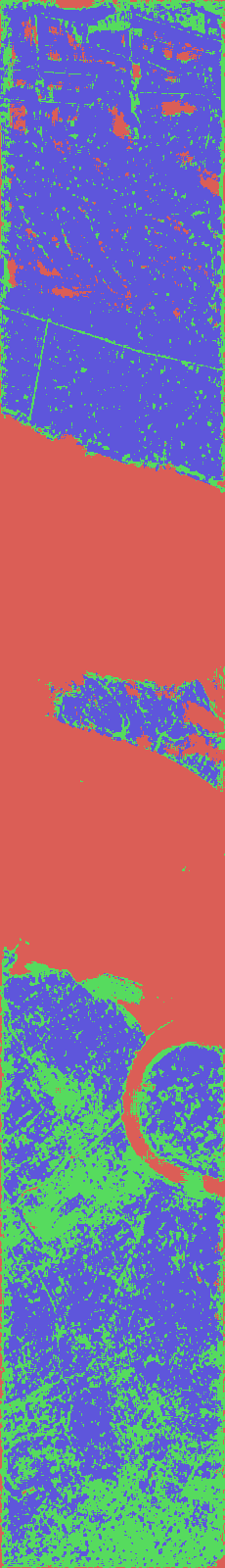}}\hfill
	\subfloat[]{\includegraphics[width=0.11\linewidth]{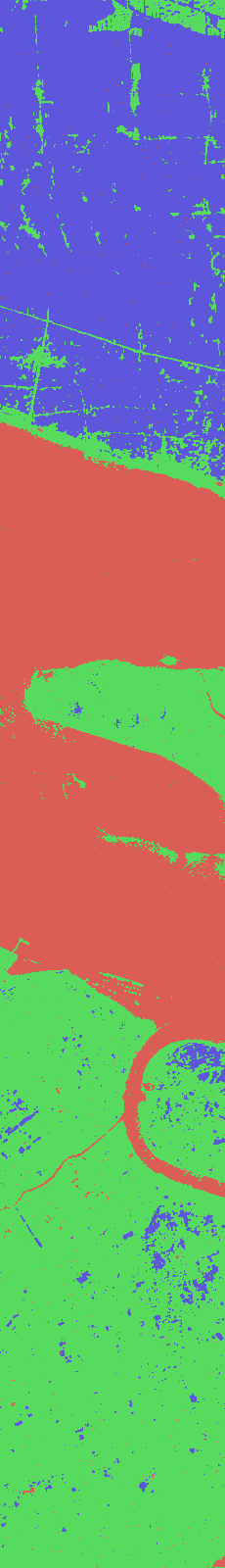}}
	\caption{(a) The ground truth of the target image, Shanghai. The predicted classification map from (b) 1D-CNN, (c) Spec-Spat, (d) 3D-CNN, (e) HSI-CNN, (f) SSRN, (g) HybirdSN, (h) HT-CNN, (i) Ours.}
	\label{fig:datacube_output}
\end{figure*}

In the second group of experiments, the pre-trained models based on 1\%, 3\% and 5\% of the source labeled samples are adopted to classify the pixels of the target datasets without additional training. The classification results of different methods on the three target datasets are shown in Tables~\ref{tab:paviaC},~\ref{tab:houston18} and~\ref{tab:datacube1}, respectively. Similarly, the average metrics over the three datasets are illustrated in Fig.~\ref{fig:target_res}. The best classification map of each method on the three target datasets are shown in Figs.~\ref{fig:pavia_output},~\ref{fig:houston_output} and~\ref{fig:datacube_output}, respectively. 

We can observe that the accuracy of all the methods decreases in this challenging task due to the pixel variability caused by different acquisition conditions. The 1D-CNN method works well on the first target dataset, but its performance drops on the second target dataset and fails on the third target dataset as shown in Fig.~\ref{fig:datacube_output}. Although  SSRN~\cite{zhong2018spectral} and HybridSN~\cite{roy2019hybridsn} work better than most of the methods on the first source dataset, they could not achieve comparable results to that of the Spec-Spat, HSI-CNN and 3D-CNN methods on the first target dataset as shown in Fig.~\ref{fig:pavia_output}. This is because there exists spectral variability between the same classes in different domains, and the methods could not catch the intrinsic representations among different domains. Same can be observed %Similarly, the Spec-Spat, HSI-CNN and 3D-CNN methods could not extract intrinsic representations 
on the second and third datasets,  %thus their performance drops on the second and third target datasets 
as shown in Figs.~\ref{fig:houston_output} and~\ref{fig:datacube_output}. 

It is worth mentioning that since HybridSN projects the dataset from the image domain to the low-dimensional domain with PCA before classification, it removes some noise and achieves better accuracy than most of the methods on the third target dataset. However, the PCA projection is not suitable for the first and second datasets, thus it fails on these two datasets as shown in Figs.~\ref{fig:pavia_output} and~\ref{fig:houston_output}. The HT-CNN methods could achieve better results than some of the other methods due to its transformed image priors, but it could not maximize its efficiency without involving training samples on the target domain. The proposed method projects data from the image domain to the shared abundance space with the affine-transfer decoder and mutual discriminative network, thus it achieves promising results and outperforms the other methods by a large margin on the target datasets as shown in Fig.~\ref{fig:target_res}. 

Note that we also compare the proposed method with the cross-temporal method proposed by Liu~\etal~\cite{liu2019unsupervised}. Since the code for Liu~\etal is not available, we directly list their result on the shared Houston dataset for fair comparison. With the model trained with 100\% training samples on the source dataset, Houston-2013, Liu~\etal~\cite{liu2019unsupervised} achieves the overall accuracy of 67\% on the target dataset, Houston-2018, while the proposed method could achieve the average overall accuracy of 71.8\% on the target dataset with only 5\% of the training samples on the source dataset.

\begin{table*}[htbp]
  \centering
  \caption{Performance comparison when the models are trained on the source dataset, PaviaU, and tested on the target dataset, PaviaC.}
    \begin{tabular}{c|c|cccccccc}
     \hline
          & Method & 1D-CNN & Spec-Spat & 3D-CNN & HSI-CNN & SSRN  & HybridSN & HT-CNN & Ours \\
     \hline
    5\%   & OA    & 83.52$\pm$6.37 & 93.45$\pm$0.33 & 92.96$\pm$1.55 & 93.76$\pm$0.87 & 61.91$\pm$4.25 & 19.08$\pm$3.61 & 79.59$\pm$2.05 & 93.92$\pm$0.58 \\
          & AA    & 71.66$\pm$7.39 & 91.56$\pm$0.66 & 90.12$\pm$4.34 & 92.20$\pm$1.06 & 66.41$\pm$6.76 & 17.77$\pm$3.37 & 68.59$\pm$0.94 & 91.78$\pm$0.74 \\
          & kappa & 77.44$\pm$8.38 & 91.02$\pm$0.45 & 90.31$\pm$2.21 & 90.12$\pm$1.17 & 49.20$\pm$5.90 & -8.70$\pm$6.59 & 71.22$\pm$2.79 & 91.55$\pm$0.75 \\
    \hline
    3\%   & OA    & 88.6$\pm$0.13 & 93.03$\pm$0.47 & 90.05$\pm$2.27 & 90.79$\pm$0.84 & 66.17$\pm$8.09 & 24.43$\pm$1.65 & 85.19$\pm$0.77 & 93.82$\pm$1.21 \\
          & AA    & 77.70$\pm$2.13 & 90.99$\pm$2.22 & 83.74$\pm$8.17 & 88.81$\pm$3.64 & 72.83$\pm$5.71 & 23.20$\pm$2.39 & 75.12$\pm$0.99 & 90.83$\pm$1.48 \\
          & kappa & 84.22$\pm$2.26 & 90.44$\pm$0.68 & 86.16$\pm$3.39 & 87.40$\pm$1.25 & 54.89$\pm$11.15 & -2.64$\pm$3.38 & 79.11$\pm$1.11 & 91.54$\pm$1.63 \\
    \hline
    1\%   & OA    & 90.45$\pm$0.72 & 91.04$\pm$0.78 & 87.95$\pm$3.21 & 88.88$\pm$0.77 & 74.9$\pm$4.01 & 17.11$\pm$4.98 & 81.71$\pm$3.96 & 91.15$\pm$1.16 \\
          & AA    & 80.00$\pm$2.12 & 90.00$\pm$0.97 & 88.97$\pm$2.69 & 86.93$\pm$1.28 & 70.12$\pm$6.01 & 16.47$\pm$4.85 & 73.18$\pm$4.88 & 87.19$\pm$1.78 \\
          & kappa & 86.78$\pm$1.01 & 87.79$\pm$1.05 & 83.82$\pm$4.16 & 84.86$\pm$1.01 & 66.05$\pm$4.62 & 10.80$\pm$6.33 & 74.35$\pm$5.60 & 87.87$\pm$1.61 \\
    \hline
    \end{tabular}%
  \label{tab:paviaC}%
\end{table*}%

\begin{table*}[htbp]
  \centering
  \caption{Performance comparison when the models are trained on the source dataset, Houston 2013, and tested on the target dataset, Houston 2018.}
    \begin{tabular}{c|c|cccccccc}
    \hline
          & Method & 1D-CNN & Spec-Spat & 3D-CNN & HSI-CNN & SSRN  & HybridSN & HT-CNN & Ours \\
    \hline
    5\%   & OA    & 39.96$\pm$4.76 & 63.06$\pm$7.97 & 49.37$\pm$12,12 & 53.53$\pm$5.69 & 57.44$\pm$2.33 & 49.41$\pm$6.47 & 57.71$\pm$4.61 & 71.80$\pm$3.31 \\
          & AA    & 41.89$\pm$6.83 & 51.02$\pm$5.18 & 53.06$\pm$1.25 & 56.81$\pm$0.98 & 61.41$\pm$7.27 & 41.05$\pm$4.62 & 46.93$\pm$2.37 & 57.80$\pm$4.95 \\
          & kappa & 19.81$\pm$5.00 & 38.97$\pm$4.25 & 31.98$\pm$7.37 & 37.78$\pm$5.74 & 44.07$\pm$2.32 & 27.33$\pm$4.87 & 40.33$\pm$3.40 & 56.36$\pm$4.69 \\
   	\hline
    3\%   & OA    & 48.75$\pm$4.4 & 59.67$\pm$5.42 & 43.73$\pm$15.21 & 50.64$\pm$7.03 & 48.4$\pm$7.09 & 43.13$\pm$7.49 & 58.64$\pm$10.20 & 64.85$\pm$2.37 \\
          & AA    & 50.13$\pm$6.04 & 46.68$\pm$11.98 & 40.93$\pm$11.62 & 56.32$\pm$2.80 & 62.55$\pm$9.42 & 43.46$\pm$3.70 & 45.55$\pm$3.86 & 60.00$\pm$2.60 \\
          & kappa & 29.06$\pm$4.70 & 35.66$\pm$9.46 & 23.36$\pm$11.25 & 34.39$\pm$5.14 & 35.50$\pm$5.15 & 24.50$\pm$4.78 & 42.10$\pm$8.56 & 50.15$\pm$3.45 \\
   	\hline
    1\%   & OA    & 67.13$\pm$5.61 & 51.32$\pm$11.78 & 29.88$\pm$28.18 & 55.4$\pm$15.02 & 46.61$\pm$8.14 & 33.19$\pm$6.49 & 34.38$\pm$14.7 &  60.56$\pm$4.63 \\
          & AA    & 65.13$\pm$7.68 & 26.81$\pm$18.55 & 21.88$\pm$6.14 & 47.10$\pm$5.99 & 56.29$\pm$8.31 & 39.78$\pm$4.19 & 32.08$\pm$11.41 & 48.10$\pm$4.62 \\
          & kappa & 46.90$\pm$4.32 & 13.56$\pm$16.32 & 8.54$\pm$8.23 & 33.02$\pm$10.83 & 31.56$\pm$5.41 & 17.85$\pm$4.91 & 20.65$\pm$8.38 & 43.33$\pm$3.36 \\
    \hline
    \end{tabular}%
  \label{tab:houston18}%
\end{table*}%

\begin{table*}[htbp]
  \centering
  \caption{Performance comparison when the models are trained on the source dataset, Hangzhou, and tested on the target dataset, Shanghai.}
    \begin{tabular}{c|c|cccccccc}
    \hline
          & Method & 1D-CNN & Spec-Spat & 3D-CNN & HSI-CNN & SSRN  & HybridSN & HT-CNN & Ours \\
    \hline
    5\%   & OA    & 47.61$\pm$6.35 & 67.45$\pm$1.48 & 65.17$\pm$2.23 & 69.28$\pm$1.85 & 66.63$\pm$7.56 & 86.62$\pm$3.46 & 77.71$\pm$5.87 & 90.01$\pm$2.43 \\
          & AA    & 54.30$\pm$0.98 & 73.59$\pm$1.22 & 71.68$\pm$2.04 & 75.13$\pm$1.14 & 78.17$\pm$3.26 & 88.02$\pm$3.35 & 78.29$\pm$7.07 & 90.01$\pm$1.82 \\
          & kappa & 26.86$\pm$9.35 & 53.89$\pm$1.97 & 50.83$\pm$3.14 & 56.37$\pm$2.39 & 46.71$\pm$12.64 & 79.37$\pm$5.42 & 66.33$\pm$8.85 & 84.65$\pm$3.77 \\
    \hline
    3\%   & OA    & 50.08$\pm$5.66 & 67.73$\pm$1.62 & 63.28$\pm$4.09 & 67.57$\pm$4.05 & 79.47$\pm$5.65 & 88.81$\pm$2.01 & 70.99$\pm$13.45 & 89.25$\pm$2.29 \\
          & AA    & 56.08$\pm$1.55 & 73.89$\pm$1.28 & 70.30$\pm$3.84 & 73.85$\pm$3.10 & 83.68$\pm$2.45 & 89.91$\pm$2.02 & 73.85$\pm$13.27 & 88.96$\pm$1.90 \\
          & kappa & 30.11$\pm$8.09 & 54.33$\pm$2.11 & 48.50$\pm$5.83 & 54.24$\pm$5.21 & 67.98$\pm$9.16 & 82.75$\pm$3.16 & 57.47$\pm$19.22 & 83.69$\pm$3.41 \\
    \hline
    1\%   & OA    & 48.06$\pm$10.15 & 67.75$\pm$3.78 & 66.93$\pm$3.82 & 62.63$\pm$3.22 & 66.55$\pm$7.06 & 88.1$\pm$1.71 & 68.54$\pm$ 0.65 & 88.61$\pm$2.52 \\
          & AA    & 57.79$\pm$0.74 & 73.97$\pm$2.76 & 73.62$\pm$2.94 & 70.11$\pm$2.53 & 69.14$\pm$15.76 & 88.72$\pm$1.93 & 74.44$\pm$0.75 &  88.50$\pm$2.12 \\
          & kappa & 26.39$\pm$15.58 & 54.46$\pm$4.71 & 53.64$\pm$4.80 & 48.02$\pm$4.06 & 46.70$\pm$12.05 & 81.63$\pm$2.69 & 54.95$\pm$0.97 & 82.71$\pm$ 3.70 \\
    \hline
    \end{tabular}%
  \label{tab:datacube1}%
\end{table*}%

\subsection{Visualizing the Shared Abundance Space}
\label{sec:evaluate_tf}
\begin{figure}[htp]
	\centering
	\subfloat[Raw spectral in the image domain ]{\includegraphics[width=0.8\linewidth]{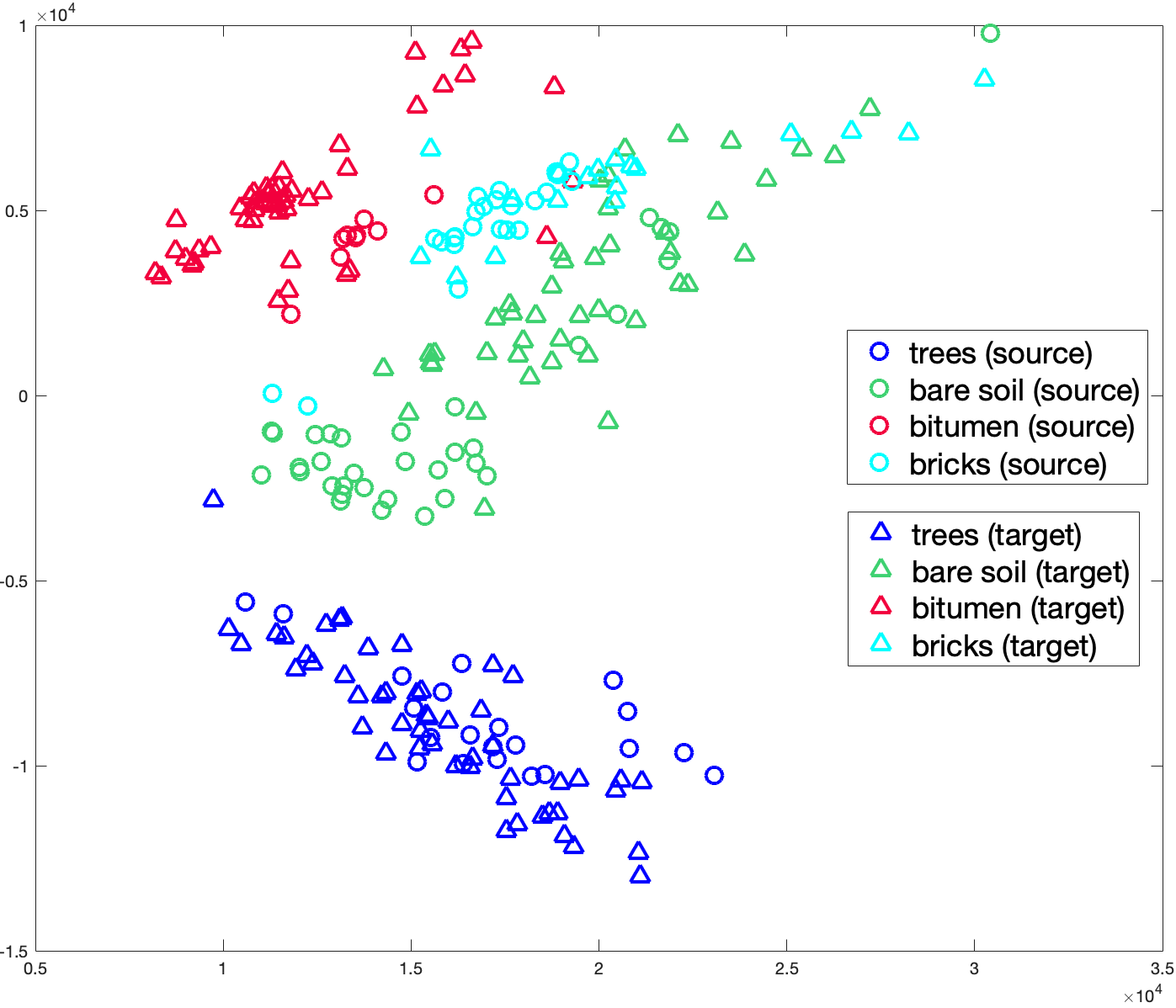} \label{fig:visual_pavia:a}}\\
	\subfloat[Abundance vectors in the abundance space]{\includegraphics[width=0.8\linewidth]{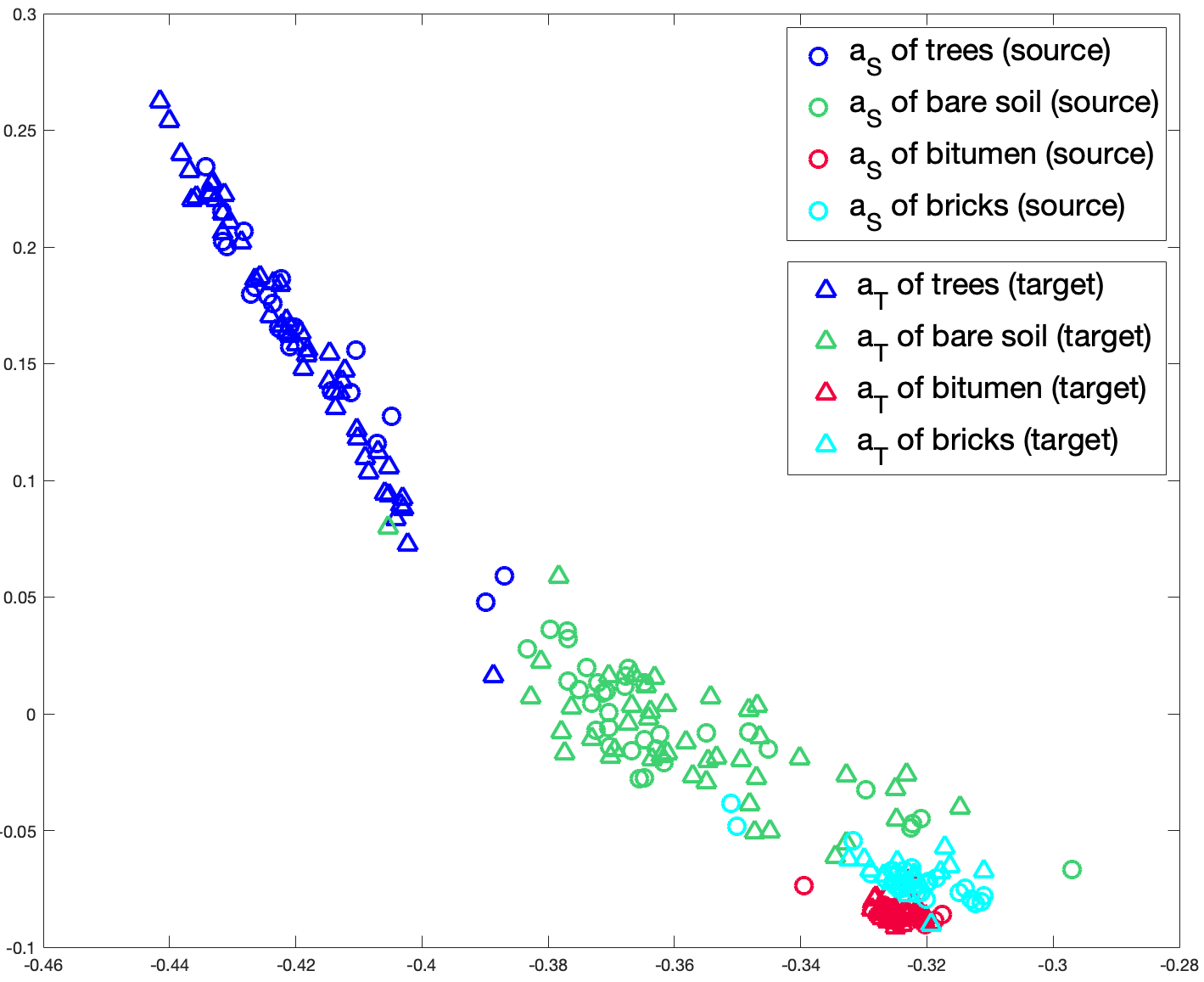}\label{fig:visual_pavia:b}}
	\caption{(a) The spectra and (b) the abundance vector of the classes on the source domain, PaviaU, and the target domain, PaviaC. Different colors represent different classes. Circles indicate the class samples on the source domain, and triangles indicate the class samples on the target domain.}
	\label{fig:visual_pavia}
\end{figure}

\begin{figure}[htp]
	\centering
	\subfloat[Raw spectra in the image domain]{\includegraphics[width=0.9\linewidth]{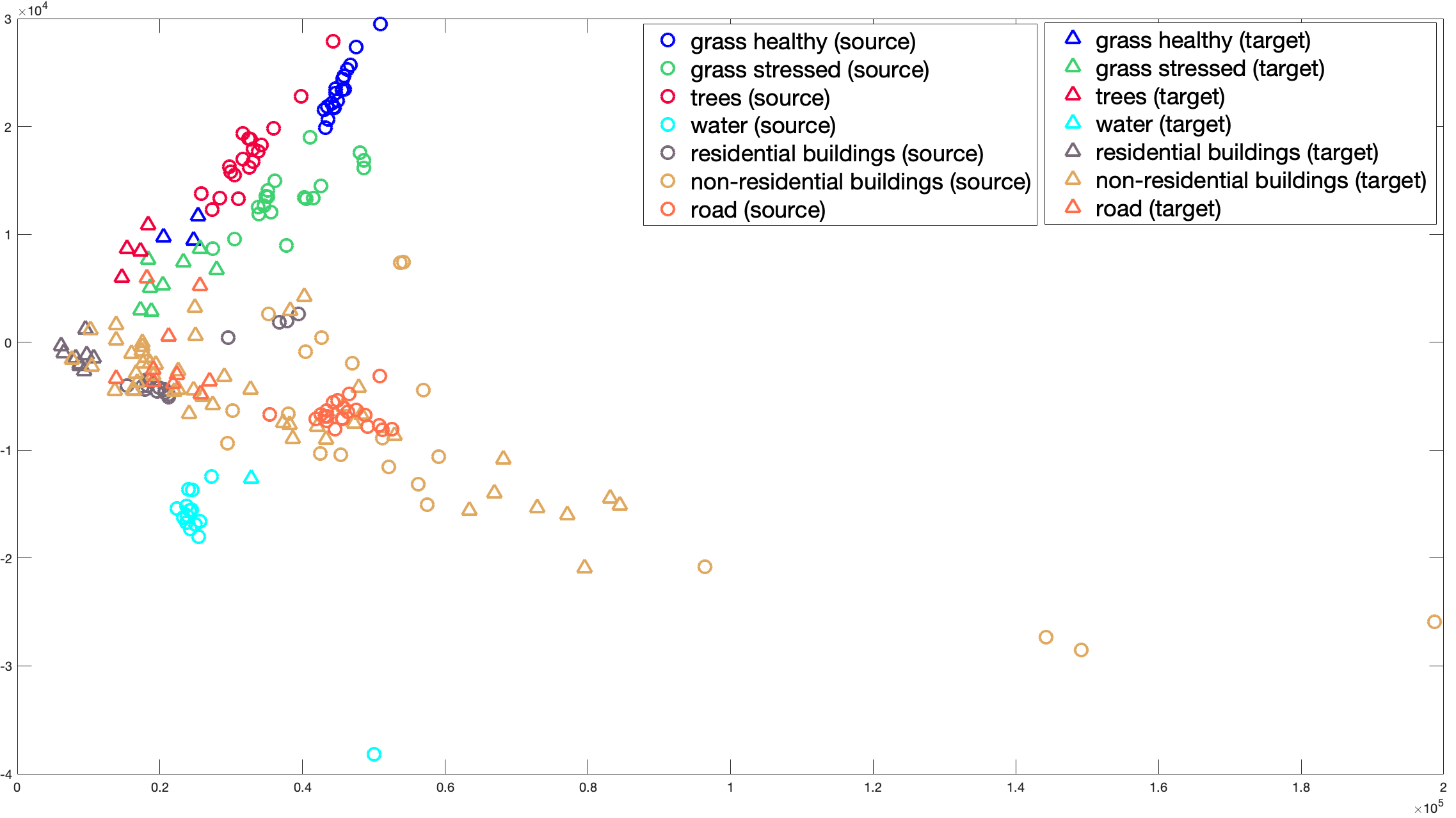}\label{fig:visual_houston:a}}\\
	\subfloat[Abundance vectors in the abundance space]{\includegraphics[width=0.9\linewidth]{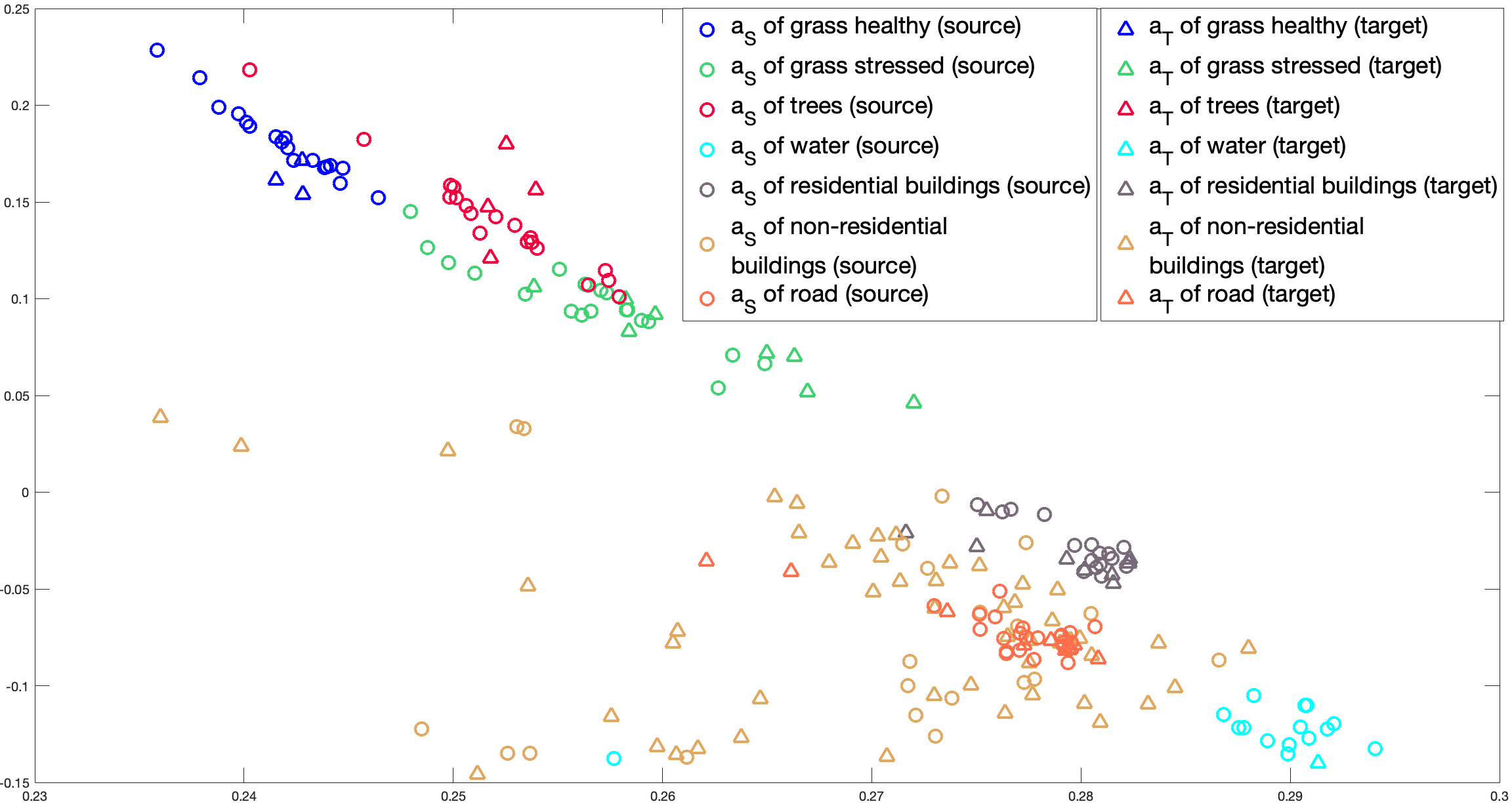}\label{fig:visual_houston:b}}
	\caption{(a) The spectra and (b) the abundance vector of the classes on the source domain, Houston-2013, and the target domain, Houston-2018. Different colors represent different classes. Circles indicate the class samples on the source domain, and triangles indicate the class samples on the target domain.}
	\label{fig:visual_houston}
\end{figure}

\begin{figure}[htp]
	\centering
	\subfloat[Raw spectra in the image domain]{\includegraphics[width=0.8\linewidth]{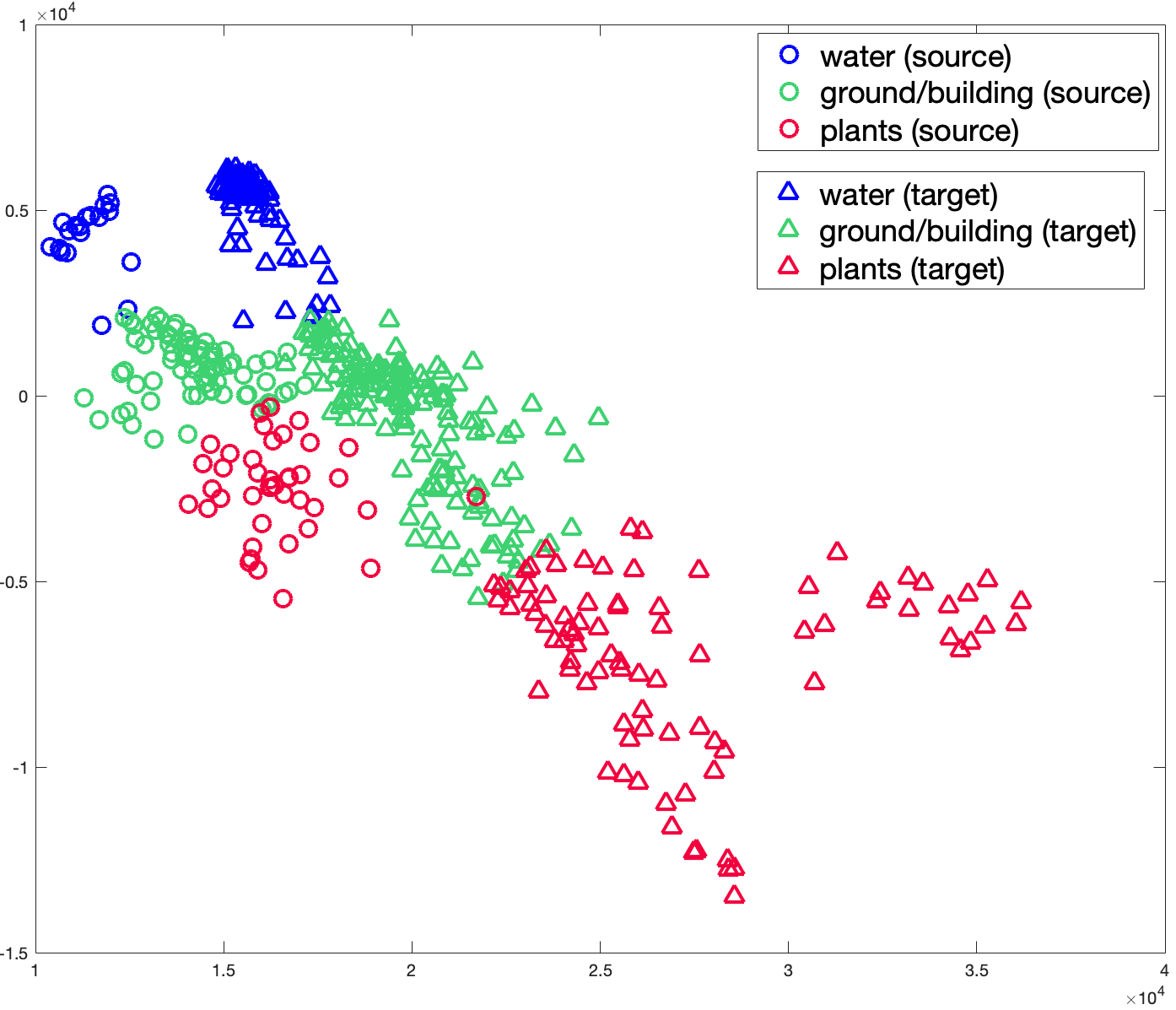}\label{fig:visual_datacube:a}}\\
	\subfloat[Abundance vectors in the abundance space]{\includegraphics[width=0.8\linewidth]{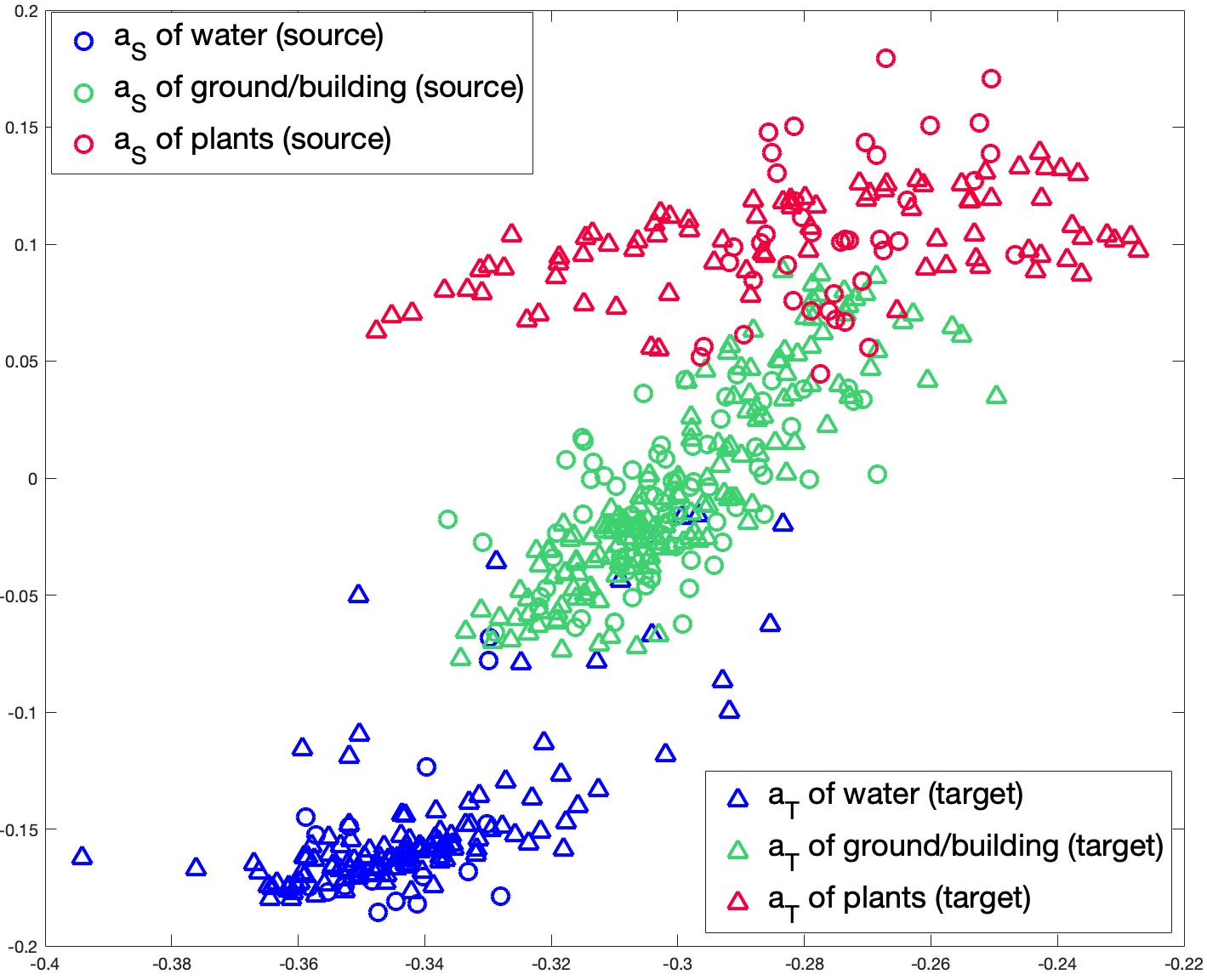}\label{fig:visual_datacube:b}}
	\caption{(a) The spectra and (b) the abundance vector of the classes on the source domain, Hangzhou, and the target domain, Shanghai. Different colors represent different classes. Circles indicate the class samples on the source domain, and triangles indicate the class samples on the target domain.}
	\label{fig:visual_datacube}
\end{figure}

As discussed in Secs.~\ref{sec:formulate} and ~\ref{sec:proposed}, with the proposed transfer learning scheme, we are able to project the data from both the source and target domains into a shared abundance space, such that the classifier trained on the source domain can work on the target domain without data labeling or network retraining. This is mainly because, with proper physical constraints, the domain discrepancy has been minimized to a large extent in the abundance space. To further demonstrate the effectiveness of the proposed transfer learning scheme, we project the HSI and its features on the abundance space into a two-dimensional space using SVD method for visualization purpose. The results of the three dataset pairs are illustrated in Figs.~\ref{fig:visual_pavia},~\ref{fig:visual_houston}, and~\ref{fig:visual_datacube}, respectively. Note that, different colors represent different classes. And the class samples on the source domain are drawn with circles, while the class samples on the target domain are drawn with triangles.

%TODO should I captalize the name of datasets?
We observe that, there are some overlap between the spectra of the same class on the source and target domains on the PaviaU-PaviaC datasets, as shown in Fig.~\ref{fig:visual_pavia:a}. This explains why some state-of-the-art methods trained on the source domain are able to perform well on the target domain. We further observe that on the second and third dataset pairs (images are taken at different time or locations), there exists large margin between the same class on the source and target domains, as shown in Figs.~\ref{fig:visual_houston:a} and~\ref{fig:visual_datacube:a}. Such domain discrepancy lead to the failure of the state-of-the-art classifiers. On the other hand, regardless of the type of datasets used, the proposed methods can always minimize the domain discrepancy between the source and target datasets by projecting the data samples onto the shared abundance space with physical constrains, as shown in Figs.~\ref{fig:visual_pavia:b},~\ref{fig:visual_houston:b}, and~\ref{fig:visual_datacube:b}, where more samples of the same classes overlap although they are from different domains.

\subsection{Ablation Study}
\label{sec:ablation}
As shown in Fig.~\ref{fig:flow}, the architecture of the proposed PCTL-SAS consists of two branches, ~\ie, (1) a reconstruction branch with shared Dirichlet-encoder, an affine-transfer decoder, and a mutual discriminative network designed based on the physical constraints, 
and (2) a classification branch with densely-connected 3D-CNN classifier. The reconstruction branch projects  both the source and target images onto the shared abundance space, and the classifier is concatenated to the shared abundance space, such that even though the network is trained on the source domain, it  could still characterize the target domain without additional data labelling or retraining. In this section, we provide comprehensive ablation studies of the above components to discuss the effectiveness of the proposed method. The first dataset is used to conduct the ablation studies and the average overall accuracy is adopted to evaluate the classification performance. 

\subsubsection{Effects of the Classifier}
\begin{figure}
	\centering
%	\subfloat[]{\includegraphics[width=0.68\linewidth]{fig/analysis/ablation_classifier_source}}\\
%	\subfloat[]
	{\includegraphics[width=0.68\linewidth]{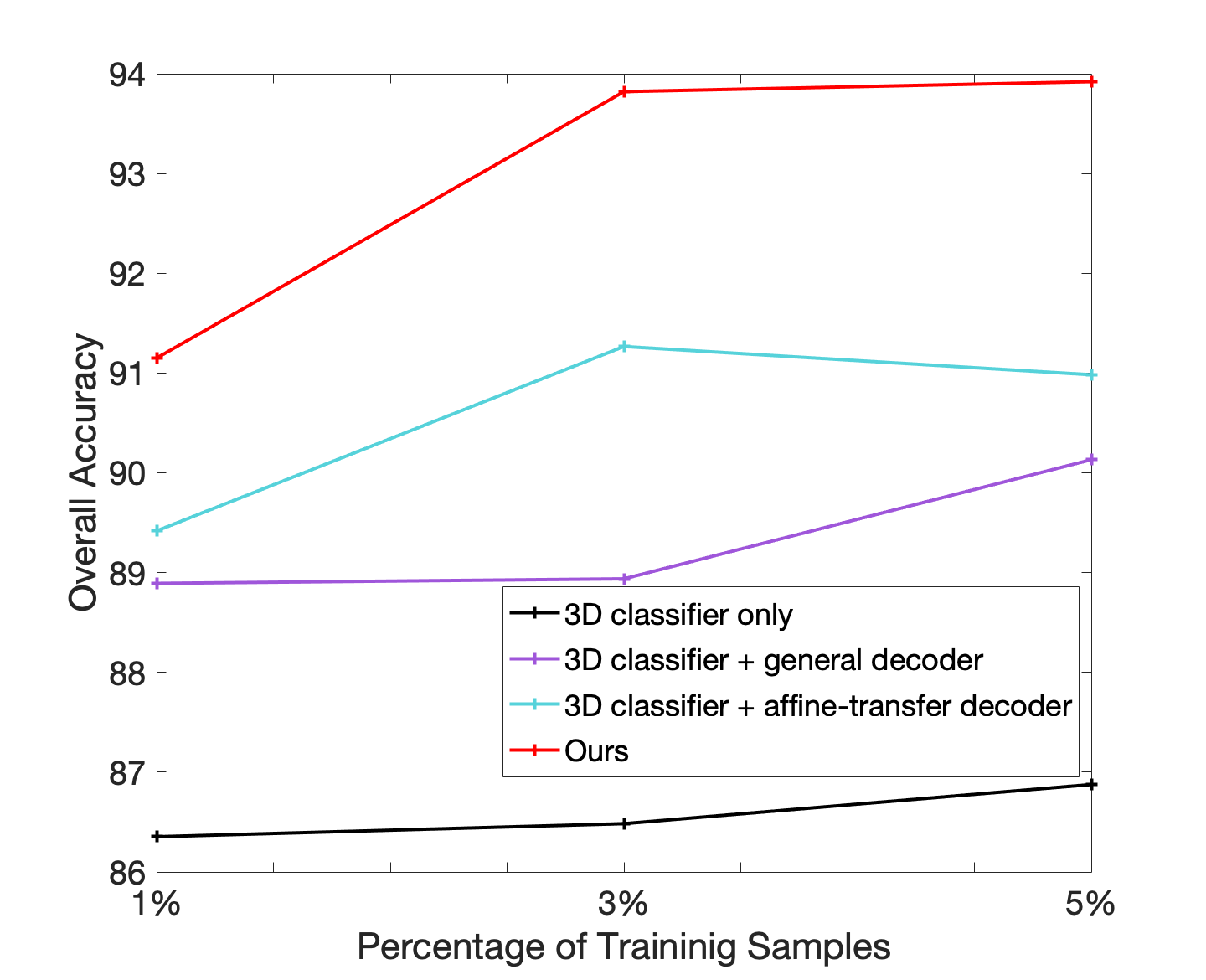}}	
	\caption{The average overall accuracy of the trained model when tested on the target dataset.}
	\label{fig:ablation_classifier}
\end{figure}

We first evaluate the densely-connected 3D-CNN classifier without the reconstruction branch. The results are reported in Fig.~\ref{fig:ablation_classifier}. We can observe that if only the 3D classifier is adopted,  its accuracy on the target dataset drops significantly. That is because the classifier itself could not extract the shared intrinsic representations between images in different domains. Therefore, the extraction of the shared representations would contribute significantly to the classification performance in a different domain. But if the training and testing data are in the same domain, a simple classifier should be sufficient to categorize pixels correctly. 

%is trained on the source image directly, the overall accuracy on the source dataset can be further increased as compared to the one with the reconstruction branch. However, its accuracy on the target dataset drops significantly. That is because the classifier itself could not extract the shared intrinsic representations between images in different domains. Therefore, the extraction of the shared representations would contribute significantly to the classification performance in a different domain. But if the training and testing data are in the same domain, a simple classifier should be sufficient to categorize pixels correctly. 

\subsubsection{Effects of the Affine-Transfer Decoder}
The affine-transfer decoder, designed based on the potential physical relationship between the same-class samples in different domains, is an important component of the reconstruction branch which allows the network to project the data from both the source and target domains into a shared abundance space. In the second experiment, we evaluate the effect of the affine-transfer decoder and the results are demonstrated in Fig.~\ref{fig:ablation_classifier} as well. Assume that there is only one general decoder to reconstruct both the source image and target images. The classification performance on the target image increases at least 3\% percent because the network learns the representations from both domains. When we adopt the proposed affine-transfer decoder, the accuracy on the target domain is further increased. This is because the network is enforced to simulate the potential physical relationship between different domains, which allows the network to extract shared representations between the source and target images.  

%on the source image drops a little bit as compared to the results from directly applying the classifier on the source image. However, its performance on the target image increases at least 3\% percent because the network learns the representations from both domains. When we adopt the proposed affine-transfer decoder, the accuracy on the target domain is further increased. This is because the network is enforced to simulate the potential physical relationship between different domains, which allows the network to extract shared representations between the source and target images.  

\subsubsection{Effects of the Sparse Constraint and Mutual Discriminative Network}
\begin{figure}
	\centering
	\subfloat[ Fix $\lambda$ and adjust $\alpha$.]{\includegraphics[width=0.7\linewidth]{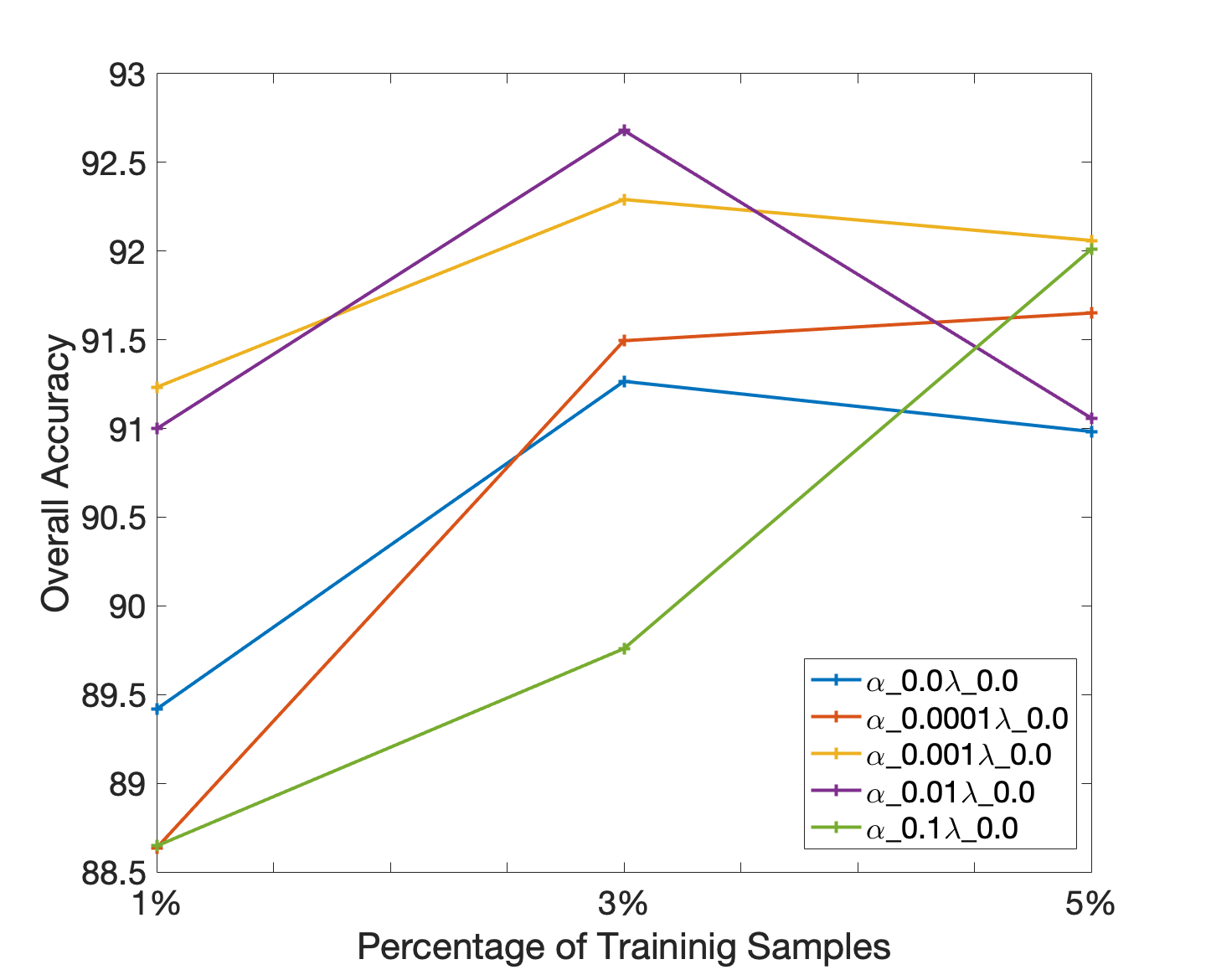}\label{fig:para1}}\\
	\subfloat[Fix $\alpha$  and adjust $\lambda$.]{\includegraphics[width=0.7\linewidth]{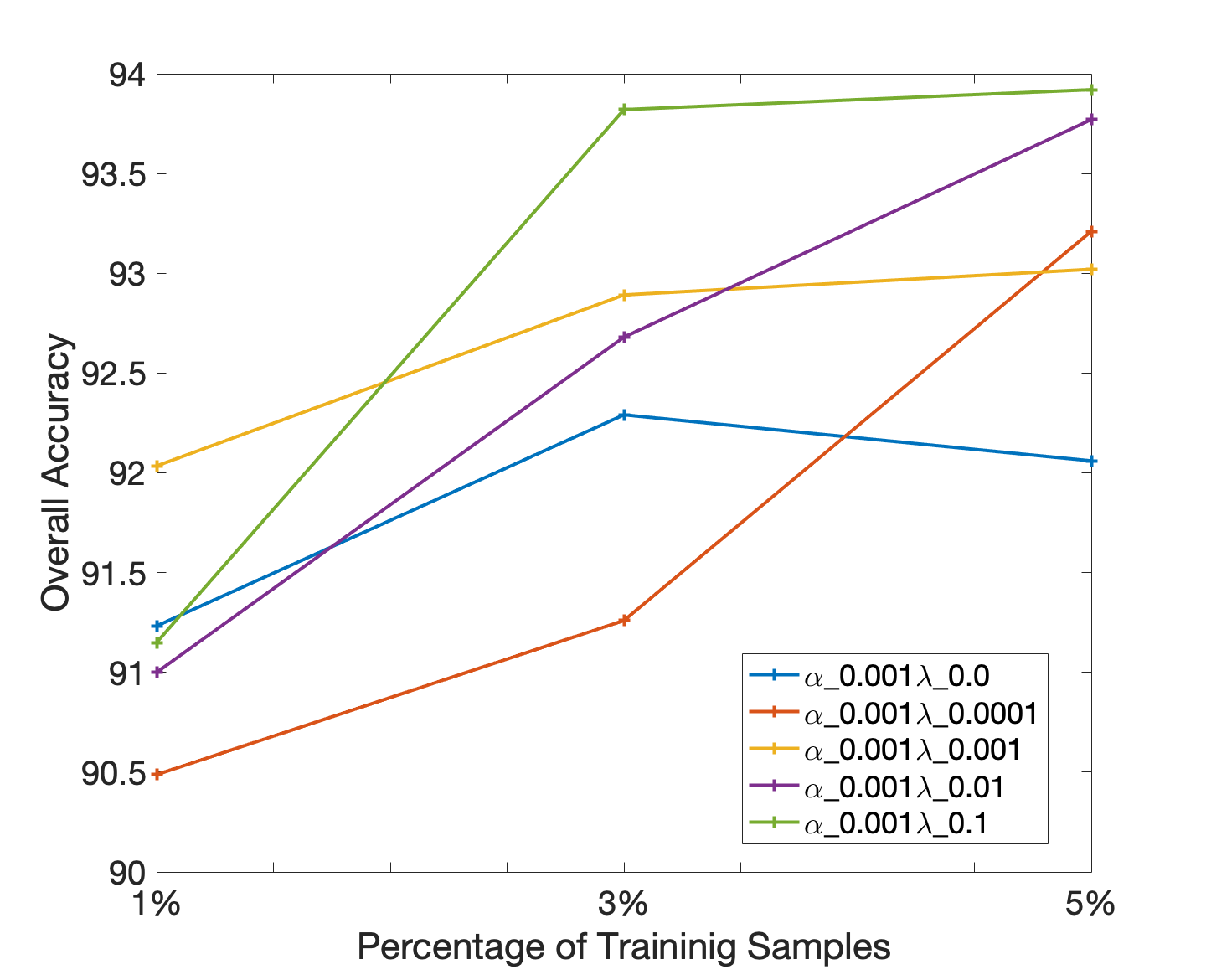}\label{fig:para2}}\\	
%	\subfloat[]{\includegraphics[width=0.7\linewidth]{fig/analysis/ablation_para3}\label{fig:para3}}\\
	\caption{The average overall accuracy on the target domain with parameter, $\lambda$ for the mutual discriminative network and $\alpha$ for the sparse constraint.}
	\label{fig:para}
\end{figure}

We also observe from Fig.~\ref{fig:ablation_classifier} that the proposed method with more physical constraints based on sparse regularization and mutual discriminative network could increase the overall accuracy of the target image to a large margin. In this study, we evaluate how these two constraints influence the transfer learning ability of the model on the target image. We first fix the parameter, $\lambda$, for mutual discriminative network, and adjust the parameter, $\alpha$, for sparse constraint. The result is shown in Fig.~\ref{fig:para1}. We can observe that, when we increase $\alpha$, the average overall accuracy gradually increases in most scenarios, which means the sparse-regularized with entropy function is able to enforce the network to extract more effective representations. But when $\alpha$ is too large, the performance is decreased because it reduces the reconstruction accuracy which, in turn, prevents the network to extract representative features. Thus in our network settings, $\alpha = 0.001$. 

In Figs.~\ref{fig:para2}, we fix $\alpha = 0.001$ and adjust the parameter $\lambda$. We can observe that in most scenarios when  $\lambda$ increases, the average overall accuracy becomes better. The main reason is that the mutual discriminative network enforces the correspondence between the pixels belonging to the same classes in different domains better. Thus, in our experimental settings, $\lambda \in \{0.01,0.1\}$.

\section{Conclusion} 
\label{sec:conclusion}
We presented a new transfer learning scheme for HSI classification such that a trained model from the source domain does not need to be retrained or fine-tuned using extra labeled data from the target domain. The key to this new learning scheme is the construction of a shared abundance space that would largely reduce the discrepancies between the different domains, due to the intrinsic physical constraints it must incorporates when constructing such a space. The learning scheme includes three novel developments. First, a shared Dirichlet-encoder is proposed that is constrained by the sparse entropy function  to extract effective representations on the shared abundance space from the raw data of different domains. Second, an affine-transfer decoder regularized by mutual discriminative network is proposed to extract shared intrinsic spectral bases such that pixels of the same class but from different domains can be matched. Third, a densely-connected 3D-CNN classifier is proposed and concatenated to the shared encoder, such that the classifier could be trained on the source domain and perform on the target domain without extra efforts for data labeling or network retraining. Extensive experiments on different types of data pairs demonstrates the superiority of the proposed approach over state-of-the-art.

\section*{Acknowledgment}
This work was supported by the Gonzalez Family Professorship Fund, and in part by the National Natural Science Foundation of China (61922013 and 41722108). The authors would like to thank all the developers of the evaluated methods who kindly offer their codes for comparisons.

\ifCLASSOPTIONcaptionsoff
  \newpage
\fi



%\bibliographystyle{IEEEtran}
%\begin{thebibliography}{1}
%  0.5em minus 0.4em\relax Harlow, England: Addison-Wesley, 1999.\\
%\end{thebibliography}

%\clearpage
\bibliographystyle{IEEEtran}
\bibliography{refs}

\begin{thebibliography}{10}
\providecommand{\url}[1]{#1}
\csname url@samestyle\endcsname
\providecommand{\newblock}{\relax}
\providecommand{\bibinfo}[2]{#2}
\providecommand{\BIBentrySTDinterwordspacing}{\spaceskip=0pt\relax}
\providecommand{\BIBentryALTinterwordstretchfactor}{4}
\providecommand{\BIBentryALTinterwordspacing}{\spaceskip=\fontdimen2\font plus
\BIBentryALTinterwordstretchfactor\fontdimen3\font minus
  \fontdimen4\font\relax}
\providecommand{\BIBforeignlanguage}[2]{{%
\expandafter\ifx\csname l@#1\endcsname\relax
\typeout{** WARNING: IEEEtran.bst: No hyphenation pattern has been}%
\typeout{** loaded for the language `#1'. Using the pattern for}%
\typeout{** the default language instead.}%
\else
\language=\csname l@#1\endcsname
\fi
#2}}
\providecommand{\BIBdecl}{\relax}
\BIBdecl

\bibitem{brown2010hydrothermal}
A.~J. Brown, S.~J. Hook, A.~M. Baldridge, J.~K. Crowley, N.~T. Bridges, B.~J.
  Thomson, G.~M. Marion, C.~R. de~Souza~Filho, and J.~L. Bishop, ``Hydrothermal
  formation of clay-carbonate alteration assemblages in the nili fossae region
  of mars,'' \emph{Earth and Planetary Science Letters}, vol. 297, no. 1-2, pp.
  174--182, 2010.

\bibitem{he2017recent}
L.~He, J.~Li, C.~Liu, and S.~Li, ``Recent advances on spectral--spatial
  hyperspectral image classification: An overview and new guidelines,''
  \emph{IEEE Trans. on Geoscience and Remote Sensing}, vol.~56, no.~3, pp.
  1579--1597, 2017.

\bibitem{liang2016sampling}
J.~Liang, J.~Zhou, Y.~Qian, L.~Wen, X.~Bai, and Y.~Gao, ``On the sampling
  strategy for evaluation of spectral-spatial methods in hyperspectral image
  classification,'' \emph{IEEE Transactions on Geoscience and Remote Sensing},
  vol.~55, no.~2, pp. 862--880, 2016.

\bibitem{paoletti2019deep}
M.~Paoletti, J.~Haut, J.~Plaza, and A.~Plaza, ``Deep learning classifiers for
  hyperspectral imaging: A review,'' \emph{ISPRS Journal of Photogrammetry and
  Remote Sensing}, vol. 158, pp. 279--317, 2019.

\bibitem{li2019deep}
S.~Li, W.~Song, L.~Fang, Y.~Chen, P.~Ghamisi, and J.~A. Benediktsson, ``Deep
  learning for hyperspectral image classification: An overview,'' \emph{IEEE
  Transactions on Geoscience and Remote Sensing}, vol.~57, no.~9, pp.
  6690--6709, 2019.

\bibitem{du2001linear}
Q.~Du and C.-I. Chang, ``A linear constrained distance-based discriminant
  analysis for hyperspectral image classification,'' \emph{Pattern
  Recognition}, vol.~34, no.~2, pp. 361--373, 2001.

\bibitem{ghamisi2017advanced}
P.~Ghamisi, J.~Plaza, Y.~Chen, J.~Li, and A.~J. Plaza, ``Advanced spectral
  classifiers for hyperspectral images: A review,'' \emph{IEEE Geoscience and
  Remote Sensing Magazine}, vol.~5, no.~1, pp. 8--32, 2017.

\bibitem{hu2015deep}
W.~Hu, Y.~Huang, L.~Wei, F.~Zhang, and H.~Li, ``Deep convolutional neural
  networks for hyperspectral image classification,'' \emph{Journal of Sensors},
  vol. 2015, 2015.

\bibitem{SongYang2019LCCf}
Y.~Song, Z.~Zhang, R.~K. Baghbaderani, F.~Wang, Y.~Qu, C.~Stuttsy, and H.~Qi,
  ``Land cover classification for satellite images through 1d cnn,'' \emph{2019
  10th Workshop on Hyperspectral Imaging and Signal Processing: Evolution in
  Remote Sensing (WHISPERS)}, vol. 2019-, pp. 1--5, 2019.

\bibitem{sharma2016hyperspectral}
V.~Sharma, A.~Diba, T.~Tuytelaars, and L.~Van~Gool, ``Hyperspectral cnn for
  image classification \& band selection, with application to face
  recognition,'' \emph{Technical report KUL/ESAT/PSI/1604, KU Leuven, ESAT,
  Leuven, Belgium}, 2016.

\bibitem{cao2018hyperspectral}
X.~Cao, F.~Zhou, L.~Xu, D.~Meng, Z.~Xu, and J.~Paisley, ``Hyperspectral image
  classification with markov random fields and a convolutional neural
  network,'' \emph{IEEE Transactions on Image Processing}, vol.~27, no.~5, pp.
  2354--2367, 2018.

\bibitem{zhu2018deformable}
J.~Zhu, L.~Fang, and P.~Ghamisi, ``Deformable convolutional neural networks for
  hyperspectral image classification,'' \emph{IEEE Geoscience and Remote
  Sensing Letters}, vol.~15, no.~8, pp. 1254--1258, 2018.

\bibitem{zhong2018spectral}
Z.~{Zhong}, J.~{Li}, Z.~{Luo}, and M.~{Chapman}, ``Spectral–-spatial residual
  network for hyperspectral image classification: A 3-d deep learning
  framework,'' \emph{IEEE Transactions on Geoscience and Remote Sensing},
  vol.~56, no.~2, pp. 847--858, 2018.

\bibitem{hamida20183}
A.~B. Hamida, A.~Benoit, P.~Lambert, and C.~B. Amar, ``3-d deep learning
  approach for remote sensing image classification,'' \emph{IEEE Transactions
  on Geoscience and Remote Sensing}, vol.~56, no.~8, pp. 4420--4434, 2018.

\bibitem{luo2018hsi}
Y.~Luo, J.~Zou, C.~Yao, X.~Zhao, T.~Li, and G.~Bai, ``Hsi-cnn: a novel
  convolution neural network for hyperspectral image,'' pp. 464--469, 2018.

\bibitem{roy2019hybridsn}
S.~K. Roy, G.~Krishna, S.~R. Dubey, and B.~B. Chaudhuri, ``Hybridsn: Exploring
  3d-2d cnn feature hierarchy for hyperspectral image classification,''
  \emph{IEEE Geoscience and Remote Sensing Letters}, 2020.

\bibitem{chen2014deep}
Y.~Chen, Z.~Lin, X.~Zhao, G.~Wang, and Y.~Gu, ``Deep learning-based
  classification of hyperspectral data,'' \emph{IEEE Journal of Selected topics
  in applied earth observations and remote sensing}, vol.~7, no.~6, pp.
  2094--2107, 2014.

\bibitem{chen2015spectral}
Y.~Chen, X.~Zhao, and X.~Jia, ``Spectral--spatial classification of
  hyperspectral data based on deep belief network,'' \emph{IEEE Journal of
  Selected Topics in Applied Earth Observations and Remote Sensing}, vol.~8,
  no.~6, pp. 2381--2392, 2015.

\bibitem{peng2017active}
L.~Peng, Z.~Hui, and K.~B. Eom, ``Active deep learning for classification of
  hyperspectral images,'' \emph{IEEE Journal of Selected Topics in Applied
  Earth Observations \& Remote Sensing}, vol.~10, no.~2, pp. 712--724, 2017.

\bibitem{luo2018shorten}
H.~Luo, ``Shorten spatial-spectral rnn with parallel-gru for hyperspectral
  image classification,'' \emph{arXiv preprint arXiv:1810.12563}, 2018.

\bibitem{mou2017deep}
L.~Mou, P.~Ghamisi, and X.~X. Zhu, ``Deep recurrent neural networks for
  hyperspectral image classification,'' \emph{IEEE Transactions on Geoscience
  and Remote Sensing}, vol.~55, no.~7, pp. 3639--3655, 2017.

\bibitem{deng2018transfer}
C.~Deng, X.~Liu, C.~Li, and D.~Tao, ``Active multi-kernel domain adaptation for
  hyperspectral image classification,'' \emph{Pattern Recognition}, vol.~77,
  pp. 306--315, 2018.

\bibitem{zhang2019hyperspectral}
H.~Zhang, Y.~Li, Y.~Jiang, P.~Wang, Q.~Shen, and C.~Shen, ``Hyperspectral
  classification based on lightweight 3-d-cnn with transfer learning,''
  \emph{IEEE Transactions on Geoscience and Remote Sensing}, vol.~57, no.~8,
  pp. 5813--5828, 2019.

\bibitem{deng2018active}
C.~Deng, Y.~Xue, X.~Liu, C.~Li, and D.~Tao, ``Active transfer learning network:
  A unified deep joint spectral--spatial feature learning model for
  hyperspectral image classification,'' \emph{IEEE Transactions on Geoscience
  and Remote Sensing}, vol.~57, no.~3, pp. 1741--1754, 2018.

\bibitem{lin2018active}
J.~Lin, L.~Zhao, S.~Li, R.~Ward, and Z.~J. Wang, ``Active-learning-incorporated
  deep transfer learning for hyperspectral image classification,'' \emph{IEEE
  Journal of Selected Topics in Applied Earth Observations and Remote Sensing},
  vol.~11, no.~11, pp. 4048--4062, 2018.

\bibitem{li2016revisiting}
Y.~Li, N.~Wang, J.~Shi, J.~Liu, and X.~Hou, ``Revisiting batch normalization
  for practical domain adaptation,'' \emph{arXiv preprint arXiv:1603.04779},
  2016.

\bibitem{melgani2004classification}
F.~Melgani and L.~Bruzzone, ``Classification of hyperspectral remote sensing
  images with support vector machines,'' \emph{IEEE Transactions on geoscience
  and remote sensing}, vol.~42, no.~8, pp. 1778--1790, 2004.

\bibitem{2017Multiscale}
H.~Yu, L.~Gao, W.~Liao, B.~Zhang, and W.~Philips, ``Multiscale superpixel-level
  subspace-based support vector machines for hyperspectral image
  classification,'' \emph{IEEE Geoscience and Remote Sensing Letters}, vol.~14,
  no.~11, pp. 1--5, 2017.

\bibitem{samaniego2008supervised}
L.~Samaniego, A.~B{\'a}rdossy, and K.~Schulz, ``Supervised classification of
  remotely sensed imagery using a modified $ k $-nn technique,'' \emph{IEEE
  Transactions on Geoscience and Remote Sensing}, vol.~46, no.~7, pp.
  2112--2125, 2008.

\bibitem{ediriwickrema1997hierarchical}
J.~Ediriwickrema and S.~Khorram, ``Hierarchical maximum-likelihood
  classification for improved accuracies,'' \emph{IEEE Transactions on
  Geoscience and Remote Sensing}, vol.~35, no.~4, pp. 810--816, 1997.

\bibitem{li2010semisupervised}
J.~Li, J.~M. Bioucas-Dias, and A.~Plaza, ``Semisupervised hyperspectral image
  segmentation using multinomial logistic regression with active learning,''
  \emph{IEEE Transactions on Geoscience and Remote Sensing}, vol.~48, no.~11,
  pp. 4085--4098, 2010.

\bibitem{licciardi2011linear}
G.~Licciardi, P.~R. Marpu, J.~Chanussot, and J.~A. Benediktsson, ``Linear
  versus nonlinear pca for the classification of hyperspectral data based on
  the extended morphological profiles,'' \emph{IEEE Geoscience and Remote
  Sensing Letters}, vol.~9, no.~3, pp. 447--451, 2011.

\bibitem{jenson1979principal}
S.~K. Jenson and F.~A. Waltz, ``Principal components-analysis and canonical
  analysis in remote-sensing,'' vol.~45, pp. 783--784, 1979.

\bibitem{bandos2009classification}
T.~V. Bandos, L.~Bruzzone, and G.~Camps-Valls, ``Classification of
  hyperspectral images with regularized linear discriminant analysis,''
  \emph{IEEE Transactions on Geoscience and Remote Sensing}, vol.~47, no.~3,
  pp. 862--873, 2009.

\bibitem{villa2011hyperspectral}
A.~Villa, J.~A. Benediktsson, J.~Chanussot, and C.~Jutten, ``Hyperspectral
  image classification with independent component discriminant analysis,''
  \emph{IEEE transactions on Geoscience and remote sensing}, vol.~49, no.~12,
  pp. 4865--4876, 2011.

\bibitem{green1988transformation}
A.~A. Green, M.~Berman, P.~Switzer, and M.~D. Craig, ``A transformation for
  ordering multispectral data in terms of image quality with implications for
  noise removal,'' \emph{IEEE Transactions on Geoscience and Remote Sensing},
  vol.~26, no.~1, pp. 65--74, 1988.

\bibitem{2018yu}
H.~{Yu}, L.~{Gao}, W.~{Liao}, P.~{Gamba}, and B.~{Zhang}, ``Global spatial and
  local spectral similarity-based group sparse representation for hyperspectral
  imagery classification,'' pp. 3579--3582, 2018.

\bibitem{yu2019global}
H.~Yu, L.~Gao, W.~Liao, B.~Zhang, L.~Zhuang, M.~Song, and J.~Chanussot,
  ``Global spatial and local spectral similarity-based manifold learning group
  sparse representation for hyperspectral imagery classification,'' \emph{IEEE
  Transactions on Geoscience and Remote Sensing}, vol.~58, no.~5, pp.
  3043--3056, 2019.

\bibitem{hong2020graph}
D.~Hong, L.~Gao, J.~Yao, B.~Zhang, A.~Plaza, and J.~Chanussot, ``Graph
  convolutional networks for hyperspectral image classification,'' \emph{arXiv
  preprint arXiv:2008.02457}, 2020.

\bibitem{li2017spectral}
Y.~Li, H.~Zhang, and Q.~Shen, ``Spectral--spatial classification of
  hyperspectral imagery with 3d convolutional neural network,'' \emph{Remote
  Sensing}, vol.~9, no.~1, p.~67, 2017.

\bibitem{yue2015spectral}
J.~Yue, W.~Zhao, S.~Mao, and H.~Liu, ``Spectral--spatial classification of
  hyperspectral images using deep convolutional neural networks,'' \emph{Remote
  Sensing Letters}, vol.~6, no.~6, pp. 468--477, 2015.

\bibitem{liu2019unsupervised}
T.~Liu, X.~Zhang, and Y.~Gu, ``Unsupervised cross-temporal classification of
  hyperspectral images with multiple geodesic flow kernel learning,''
  \emph{IEEE Transactions on Geoscience and Remote Sensing}, vol.~57, no.~12,
  pp. 9688--9701, 2019.

\bibitem{he2019heterogeneous}
X.~{He}, Y.~{Chen}, and P.~{Ghamisi}, ``Heterogeneous transfer learning for
  hyperspectral image classification based on convolutional neural network,''
  \emph{IEEE Transactions on Geoscience and Remote Sensing}, vol.~58, no.~5,
  pp. 3246--3263, 2020.

\bibitem{simonyan2014very}
K.~Simonyan and A.~Zisserman, ``Very deep convolutional networks for
  large-scale image recognition,'' \emph{arXiv preprint arXiv:1409.1556}, 2014.

\bibitem{Borsoi2020HSIReview}
R.~Borsoi, T.~Imbiriba, J.~Bermudez, C.~Richard, J.~Chanussot, L.~Drumetz,
  J.~Tourneret, A.~Zare, and C.~Jutten, ``Spectral variability in hyperspectral
  data unmixing: A comprehensive review,'' \emph{CoRR}, vol. abs/2001.07307,
  Under Review.

\bibitem{nalisnick2016deep}
E.~Nalisnick and P.~Smyth, ``Deep generative models with stick-breaking
  priors,'' \emph{ICML}, 2017.

\bibitem{qu2018unsupervised}
Y.~Qu, H.~Qi, and C.~Kwan, ``Unsupervised sparse dirichlet-net for
  hyperspectral image super-resolution,'' \emph{Proceedings of the IEEE
  Conference on Computer Vision and Pattern Recognition}, pp. 2511--2520, 2018.

\bibitem{huang2017sparse}
S.~Huang and T.~D. Tran, ``Sparse signal recovery via generalized entropy
  functions minimization,'' \emph{arXiv preprint arXiv:1703.10556}, 2017.

\bibitem{belghazi2018mine}
I.~Belghazi, S.~Rajeswar, A.~Baratin, R.~D. Hjelm, and A.~Courville, ``Mine:
  mutual information neural estimation,'' \emph{arXiv preprint
  arXiv:1801.04062}, 2018.

\bibitem{Hjelm2018Learning}
R.~D. Hjelm, A.~Fedorov, S.~Lavoie-Marchildon, K.~Grewal, P.~Bachman,
  A.~Trischler, and Y.~Bengio, ``Learning deep representations by mutual
  information estimation and maximization,'' 2018.

\bibitem{ioffe2015batch}
S.~Ioffe and C.~Szegedy, ``Batch normalization: Accelerating deep network
  training by reducing internal covariate shift,'' \emph{arXiv preprint
  arXiv:1502.03167}, 2015.

\bibitem{nair2010rectified}
V.~Nair and G.~E. Hinton, ``Rectified linear units improve restricted boltzmann
  machines,'' in \emph{ICML}, 2010.

\bibitem{srivastava2014dropout}
N.~Srivastava, G.~Hinton, A.~Krizhevsky, I.~Sutskever, and R.~Salakhutdinov,
  ``Dropout: a simple way to prevent neural networks from overfitting,''
  \emph{The journal of machine learning research}, vol.~15, no.~1, pp.
  1929--1958, 2014.

\bibitem{paviadata}
B.~A. M~Graña, MA~Veganzons, ``Hyperspectral remote sensing scenes,''
  \url{http://www.ehu.eus/ccwintco/index.php?title=Hyperspectral_Remote_Sensing_Scenes}.

\bibitem{debes2014hyperspectral}
C.~Debes, A.~Merentitis, R.~Heremans, J.~Hahn, N.~Frangiadakis, T.~van
  Kasteren, W.~Liao, R.~Bellens, A.~Pi{\v{z}}urica, S.~Gautama, W.~Philips,
  S.~Prasad, Q.~Du, and F.~Pacifici, ``Hyperspectral and lidar data fusion:
  Outcome of the 2013 grss data fusion contest,'' \emph{IEEE Journal of
  Selected Topics in Applied Earth Observations and Remoe Sensing}, vol.~7,
  no.~6, pp. 2405--2418, 2014.

\bibitem{le20182018}
B.~Le~Saux, N.~Yokoya, R.~Hansch, and S.~Prasad, ``2018 ieee grss data fusion
  contest: Multimodal land use classification [technical committees],''
  \emph{IEEE Geoscience and Remote Sensing Magazine}, vol.~6, no.~1, pp.
  52--54, 2018.

\bibitem{ye2017dictionary}
M.~Ye, Y.~Qian, J.~Zhou, and Y.~Y. Tang, ``Dictionary learning-based
  feature-level domain adaptation for cross-scene hyperspectral image
  classification,'' \emph{IEEE Transactions on Geoscience and Remote Sensing},
  vol.~55, no.~3, pp. 1544--1562, 2017.

\end{thebibliography}
\end{document}